\documentclass[]{amap}
\usepackage{multirow}
\usepackage[toc,page,header]{appendix}
\usepackage{minitoc}
\usepackage{solarized-light}
\usepackage{booktabs}
\usepackage{tikz}
\usetikzlibrary{arrows.meta, positioning}
\usepackage{multirow}
\usepackage{graphicx}
\usepackage{array}
\usepackage{siunitx,array}
\usepackage[table]{xcolor}
\usepackage{makecell,array,booktabs}
\usepackage{makecell}
\usepackage{framed}

\usepackage{bbding}
\usepackage{hyperref}
\usepackage{amsmath}
\usepackage{amsfonts}
\usepackage{placeins}
\usepackage[colorinlistoftodos]{todonotes}
\usepackage{longtable}
\usepackage{hhline}
\usepackage{fancyvrb}
\usepackage{graphicx}
\usepackage[table]{xcolor}
\usepackage{booktabs}
\usepackage{float}
\usepackage{fvextra}
\usepackage{CJKutf8}
\usepackage{multicol}
\usepackage{float}
\usepackage{placeins}
\usepackage{cleveref}
\usepackage{tablefootnote}
\usepackage{threeparttable}
\usepackage{tabularx}
\usepackage{mdframed}
\usepackage{subcaption}
\usepackage[usestackEOL]{stackengine}
\usepackage[numbers]{natbib}
\newcommand{\commentout}[1]{}
\renewcommand{\paragraph}[1]{\noindent\textbf{#1.}\hspace*{1em}}
\usepackage{enumitem}
\usepackage{hyperref}
\setlist[itemize]{leftmargin=15pt}
\usepackage{siunitx,array}
\usepackage{adjustbox}
\usepackage{arydshln}
\usepackage{pifont}
\usepackage{bbding}
\definecolor{ampblue}{rgb}{0.82, 0.88, 0.94}
\usepackage[table]{xcolor} % 用于颜色
\usepackage{array} 
\usepackage[dvipsnames]{xcolor}
\usepackage{booktabs}
\usepackage{multirow}
\usepackage{makecell}
\usepackage{threeparttable}
\usepackage{multirow}
\usepackage{adjustbox}
\usepackage{float}
\usepackage{caption}

\usepackage[utf8]{inputenc}
\usepackage{booktabs}
\usepackage{array}
\usepackage{pgfplots}
\usepgfplotslibrary{groupplots}
\pgfplotsset{compat=1.18}
\usepackage{xcolor}
\usepackage{listings}
\lstdefinestyle{promptstyle}{
    basicstyle=\ttfamily\scriptsize,
    breaklines=true,
    breakatwhitespace=false,
    columns=fullflexible,
    keepspaces=true,
    showstringspaces=false,
    frame=single,
    rulecolor=\color{black!20},
    backgroundcolor=\color{black!2},
    xleftmargin=1em,
    xrightmargin=1em,
    aboveskip=4pt,
    belowskip=6pt
}

\newcommand{\promptheading}[1]{%
    \vspace{0.6em}
    \noindent\textbf{#1}
    \par\vspace{0.25em}
}

\providecommand{\memgroup}[2]{%
\midrule
\multicolumn{#1}{c}{\textit{#2}}\\
\midrule
}

\RequirePackage{xspace}
\makeatletter
\DeclareRobustCommand\onedot{\futurelet\@let@token\@onedot}
\def\@onedot{\ifx\@let@token.\else.\null\fi\xspace}

\makeatother

\usepackage{xcolor}

\definecolor{abot1}{HTML}{0185FE}
\definecolor{abot2}{HTML}{0185FE}
\definecolor{abot3}{HTML}{0185FE}
\definecolor{abot4}{HTML}{0185FE}
\definecolor{abot5}{HTML}{FB8C00}
\definecolor{abot6}{HTML}{FB8C00}
\definecolor{abot7}{HTML}{FB8C00}

\title{ABot-AgentOS: A General Robotic Agent OS with Lifelong Multi-modal Memory}

\author{AMAP CV Lab}
\vspace{-11pt}

\contribution{See \hyperref[sec:contributions]{Contributions} section for a full author list.}

\abstract{

Recent VLM and VLA systems have improved robotic perception and action prediction, yet long-horizon embodied agents still require a general runtime layer for reasoning, memory, tool use, verification, and cross-embodiment execution. We present \textbf{ABot-AgentOS}, a general robotic Agent Operating System that sits above low-level controllers and provides a deliberative agent layer for scene-conditioned planning, context-isolated skill execution, multi-stage verification, multi-modal memory, and edge-cloud collaboration. To evaluate such systems, we introduce \textbf{EmbodiedWorldBench}, an executable benchmark with 16 indoor, outdoor, and hybrid scenes, four difficulty levels, and over 200 tasks involving navigation, object search, NPC dialogue, dynamic events, and trace-grounded scoring.

ABot-AgentOS further introduces \textbf{Universal Multi-modal Graph Memory}, a persistent source-grounded substrate that converts dialogue, visual observations, spatial context, temporal relations, and task traces into typed nodes and edges. A failure-driven self-evolution loop converts diagnosed memory failures into gated runtime evo-assets that are promoted only to later evaluation splits, preventing current-split ground-truth leakage while enabling continual improvement. On an initial EmbodiedWorldBench subset, ABot-AgentOS improves over a single-controller baseline in both task success and goal completion. Across memory benchmarks, ABot-AgentOS Static achieves 87.5 on LoCoMo, 59.9 on OpenEQA EM-EQA, 88.6 on Mem-Gallery, and 76.5 Acc@All on NExT-QA; self-evolution further improves LoCoMo to 88.7, OpenEQA to 60.4, and Mem-Gallery to 89.0. These results suggest that a general Agent OS layer can improve long-horizon embodied execution while providing persistent, auditable memory for continual interaction.

\bigskip

\textbf{Date:} July 10, 2026

\textbf{Project Page:} \url{https://amap-cvlab.github.io/ABot-AgentOS}

}

\begin{document}
\maketitle
\vspace{-4pt}

% \begin{figure}[!h]
%     \centering
%     \vspace{-10pt}
% \includegraphics[width=0.6\linewidth]{figure/bot5.png}
%     % \caption{\textbf{xxx}.}
%     \label{fig:model}
% \end{figure} 

% \begin{figure}[!h]
%     \centering
%     \vspace{-10pt}
% \includegraphics[width=1.0\linewidth]{figure/teaser1.png}
%     \caption{\textbf{\VLAFM{}}.}
%     \label{fig:model}
% \end{figure}

% \begin{figure}[!h]
%     \centering
%     \vspace{-10pt}
% \includegraphics[width=1.0\linewidth]{figure/teaser2.png}
%     \caption{\textbf{\VLAFM{}}.}
%     \label{fig:model}
% \end{figure}

% \begin{figure}[!h]
%     \centering
%     \vspace{-10pt}
% \includegraphics[width=1.0\linewidth]{figure/白色背景.png}
%     \caption{\textbf{\VLAFM{}}.}
%     \label{fig:model}
% \end{figure}

\newpage
\tableofcontents
\newpage

%\newpage

\section{Introduction}
\label{sec:intro}

Embodied intelligence is rapidly ushering artificial intelligence from digital environments into the physical world. Recent breakthroughs in Vision-Language Models (VLMs) and vision-language-action systems have endowed robots with unprecedented capabilities for natural language understanding, visual scene comprehension, navigation, and action prediction ~\cite{gong2026poinavbenchmarkingenhancingfinalmeters,chen2026explorelikehumansautonomous,chu2026abotn0technicalreportvla,chen2026astranavworldworldmodelforesight,xiang2025navr2dualrelationreasoninggeneralizable,Liu_2026_CVPR,Chen_2026_CVPR,xue2026omninavunifiedframeworkprospective}. Nevertheless, a critical chasm remains between understanding and acting: how can high-level semantic reasoning be translated into reliable multi-step physical execution? How can capabilities generalize across diverse robot morphologies without extensive retraining? And how can robots build long-term, persistent memory for continuous interaction? These are foundational challenges that must be addressed to transition embodied agents from lab prototypes to practical, scalable systems.

Three lines of work point toward a solution, but remain only partially connected. First, drawing on the broad motivation of Dual-System Theory~\cite{team2025gemini}, robotic foundation-model systems have begun to combine fast perception-action policies with slower deliberative reasoning. Galaxea G0 ~\cite{jiang2025galaxea}, Hi Robot ~\cite{shi2025hi}, and the RoboBrain lineage ~\cite{ji2025robobrain,team2025robobrain} demonstrate how VLA models and hierarchical robot intelligence can connect abstract conceptual grounding with concrete motor execution. Brain-inspired systems such as RoboMemory ~\cite{lei2025robomemory} and Agentic Robot ~\cite{yang2025agentic} further explore memory architectures and cognitive decomposition for embodied agents. These works significantly advance robotic perception-to-action mapping, but often remain tied to a particular model stack, embodiment, or control interface.

Second, general agent research has shown the value of explicit reasoning, tool use, and executable workflows. ReAct~\cite{yao2022react}, Toolformer ~\cite{schick2023toolformer}, and SWE-agent ~\cite{yang2024sweagent} show that language models can interleave reasoning with actions and use external tools; newer agent systems such as OpenAI Operator ~\cite{openai2025operator}, ChatGPT agent ~\cite{openai2025chatgptagent}, and Claude computer use~\cite{anthropic2024computeruse} extend this idea to interactive digital environments. Benchmarks such as OSWorld~\cite{xie2024osworld} further reveal that even strong multi-modal agents struggle with open-ended long-horizon execution. In robotics, systems such as Gemini Robotics~\cite{deepmind2025geminirobotics}, SayCan~\cite{ahn2022saycan}, Inner Monologue~\cite{huang2022innermonologue}, Voyager~\cite{wang2023voyager}, and AutoGen~\cite{wu2023autogen} illustrate the importance of grounded skill use, feedback, reusable skills, and agent specialization. However, physical execution introduces additional constraints absent from most software agents: partial observability, actuation uncertainty, ambiguous completion signals, and the need to verify whether a planned action actually changed the world.

Third, long-term memory for agents has progressed from memory streams and reflection in early generative agents~\cite{generativeagents} to persistent user memory, hierarchical context management, and scalable memory infrastructure in MemoryBank, MemGPT, and Mem0~\cite{memorybank,packer2023memgpt,mem0}. These systems show that agents should not rely only on the current prompt or parametric model memory. Yet embodied agents require a stronger substrate: memory must bind dialogue, egocentric visual evidence, object states, person or animal identities, places, temporal relations, spatial relations, provenance, and robot task traces in a form that can be retrieved and audited during future physical interaction.

Taken together, existing systems leave three gaps. First, there is a reasoning-execution gap: many foundation-model-based robotic systems map model outputs directly to actions or rely on monolithic pipelines, lacking an intermediate agent layer for task decomposition, tool invocation, skill delegation, verification, recovery, and multi-step logical reasoning. Second, there is an embodiment-generalization gap: robotic agent systems are often tightly coupled to specific hardware, control APIs, or environment assumptions, making cross-body transfer costly. Third, there is a persistent embodied-memory gap: short-term buffers, text-only caches, or task-specific memory modules cannot reliably preserve multi-modal, relational, source-grounded experience across long-term interaction. These gaps also expose an evaluation need: long-horizon embodied agents require benchmarks that go beyond isolated navigation, manipulation, or VQA and instead test executable multi-scene scenarios with dynamic events, interaction, and trace-grounded scoring.

To bridge these gaps, we propose ABot-AgentOS, a General Robotic Agent Operating System that provides a reusable agent layer  above low-level controllers for embodied reasoning, memory, tool use, verification, and cross-embodiment execution. Built upon powerful VLMs and robot skill interfaces, ABot-AgentOS constructs a full-fledged Agent OS that connects cognitive reasoning with physical execution. Our core contributions are threefold:

\begin{enumerate}
    \item \textbf{A General Robotic Agent OS Architecture Bridging VLMs and Physical Execution.}We design a modular Agent OS situated above the VLM foundation model and low-level robot controllers. By decoupling high-level cognition from physical actuation through structured agent components---including a main LLM, context management, a skill runner, tool interfaces, and multi-stage verification---we support conversational dialogue, locomotion control, and fine-grained manipulation within a unified runtime. Rather than baking capabilities into a single model, ABot-AgentOS adopts a plugin-based skill integration paradigm. By plugging in embodiment-specific mobility and manipulation interfaces, the system can adapt to heterogeneous forms ranging from humanoid robots to quadrupedal dogs.
     \item \textbf{A Comprehensive Benchmark for Agent Capabilities.}We introduce EmbodiedWorldBench, a systematic evaluation framework covering a wide spectrum of long-horizon embodied tasks. The benchmark probes agent competencies including hierarchical task planning, multi-step reasoning, tool-augmented problem solving, error recovery, NPC interaction, dynamic instruction response, and robustness under partial observability. By moving beyond static manipulation or navigation-only settings to executable indoor, outdoor, and hybrid scenarios, it provides a standardized yardstick for assessing next-generation embodied agents.
      \item \textbf{A Universal Multi-modal Memory System with Trace-Driven Lifelong Self-Evolution.} ABot-AgentOS introduces a typed, source-grounded multi-modal memory graph that stores compact records of entities, events, places, sessions, visual evidence, temporal context, spatial relations, and provenance. Instead of retaining raw transcripts, full videos, or isolated text snippets, the system writes embodied experience into structured nodes and edges, retrieves local evidence subgraphs through a fixed hybrid graph retriever, and records inspectable retrieval traces. These traces enable a failure-driven lifelong self-evolution loop: after each evaluation or deployment split, ABot-AgentOS diagnoses memory-writing, evidence-selection, frame-selection, temporal-grounding, entity-matching, and answer-composition failures, then compiles safe fixes into gated runtime evo-assets that can be promoted only to later splits.

\end{enumerate}

In addition to these public system, benchmark, and memory components, we also describe a deployment-oriented small-model training pipeline. This pipeline transfers long-horizon tool-use behavior from stronger teacher agents to a smaller embodied agent model through text-based embodied environments, teacher trajectory distillation, supervised fine-tuning, reinforcement learning, and a self-evolving LLM-as-a-Judge reward engine. Because this component targets business deployment, we focus on the method and do not release private data or production results.

Through these innovations, ABot-AgentOS delivers a cohesive, deployable Agent OS for embodied intelligence, providing the deliberative reasoning, memory, tool-use, and verification layer needed above low-level robot controllers. It provides a practical foundation for building robotic agents that can perceive, remember, reason, act, evaluate their own progress, and improve over long-term interaction with the physical world.
\section{Agent Framework}  
\label{sec:framework}  

\subsection{Architecture Overview}

\begin{figure*}[ht] 
\vspace{-8mm}
\centering 
\includegraphics[width=0.75\linewidth]{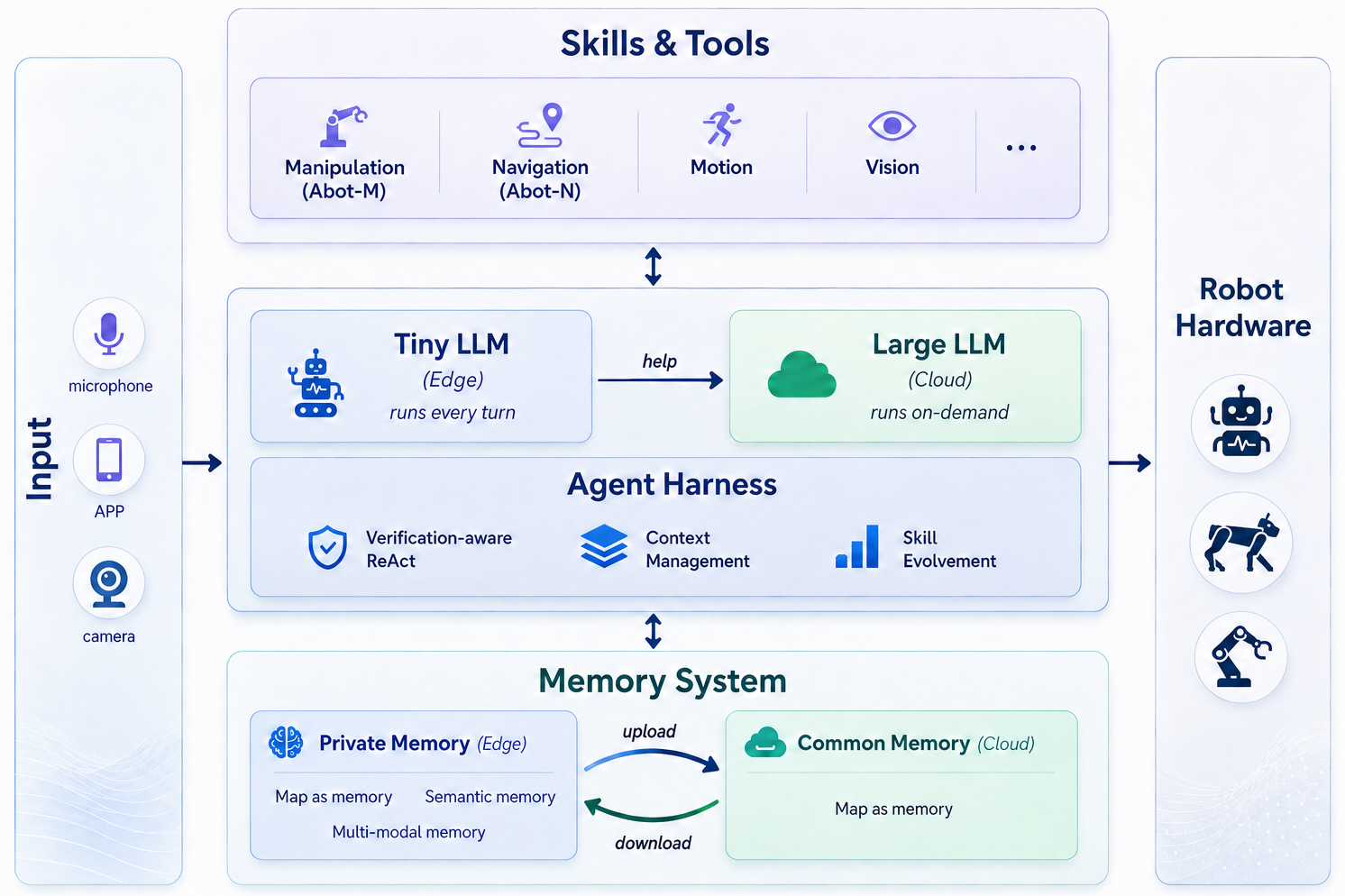}
\caption{System architecture of the proposed robot agent. Inputs from multiple sources (microphone, APP, camera) to a dual-LLM core, where a Tiny LLM on the edge handles every turn and escalates to a cloud-based Large LLM on demand. The Agent Harness manages verification-aware ReAct loop, context, and skill evolvement, enabling an extensible skill library (Manipulation, Navigation, Motion, Vision, etc.). A hierarchical Memory System synchronizes private edge memory with shared cloud memory to support cross-robot knowledge transfer. Outputs are dispatched to the robot hardware for execution. } %最终文档中希望显示的图片标题
\label{Fig.main2} %用于文内引用的标签
\end{figure*}

ABot-AgentOS is a general-purpose robotic Agent Operating System
designed for embodied intelligence, providing a deliberative agent layer
above robot hardware and low-level controllers. Rather than replacing existing control stacks, ABot-AgentOS provides a unified cognitive layer that connects multi-modal perception, memory, reasoning, planning, and skill execution across heterogeneous robotic platforms, including humanoids, quadruped robot dogs, mobile manipulators, and robotic arms.

As shown in Figure~\ref{Fig.main2} , ABot-AgentOS receives multi-modal inputs from microphones, mobile applications, cameras, and other sensors, and maps them into a runtime loop for perception, reasoning, and action. The runtime adopts an edge-cloud collaborative LLM architecture. A lightweight Tiny LLM runs on the edge at every interaction turn, enabling low-latency perception, instruction understanding, state tracking, and routine decision-making. When the task requires more complex reasoning, long-horizon planning, or ambiguity resolution, the system can request help from a cloud-based Large LLM, which is invoked on demand to provide stronger semantic understanding and planning capability.

The Agent Harness layer contains three key components: Verification-aware ReAct, Context Management, and Skill Evolvement. It organizes prompts, tools, APIs, and execution protocols into a reliable agent loop. Verification-aware ReAct adds a verifier module on top of ReAct to improve the robustness of the agent system. Context Management selects, compresses, and retrieves relevant information from observations, dialogue history, robot states, and memory. Skill Evolvement enables the agent to refine, reuse, and expand its skill set through continuous interaction with the physical world.

% Above the runtime, ABot-AgentOS abstracts robot capabilities into a unified Skills and Tools layer, including manipulation skills such as Abot-M ~\cite{yang2026abot}, navigation skills such as Abot-N ~\cite{chu2026abot},
Above the runtime, ABot-AgentOS abstracts robot capabilities into a unified Skills and Tools layer, including manipulation skills~\cite{yang2026abot}, navigation skills~\cite{chu2026abot,yang2026asyncshieldplugandplayedgeadapter,yang2025cenavflowguidedreinforcementrefinement},
motion control, vision, and other extensible tools. This abstraction allows the same agent system to operate across different robot embodiments while reusing high-level reasoning and task-planning logic.

A central component of ABot-AgentOS is its multi-modal memory system, which consists of edge-side private memory and cloud-side common memory. Private memory stores robot-specific information such as maps, semantic knowledge, multi-modal interaction history, user preferences, and local environmental experiences, enabling personalization and privacy-preserving operation. Common memory aggregates shareable knowledge, such as reusable map representations and general task experiences, and supports upload/download mechanisms for knowledge transfer across robots. In this way, individual robots can learn from their own embodied experiences while benefiting from collective knowledge accumulated by other agents.

Overall, ABot-AgentOS integrates edge intelligence, cloud reasoning, multi-modal memory, and modular skill execution into a unified operating system for general robotic agents. By decoupling high-level cognition from specific hardware embodiments, it provides a scalable foundation for building embodied agents that can perceive, remember, reason, act, and continuously evolve in real-world environments.

\subsection{Agent Harness}

\begin{figure*}[ht] 
\centering 
\includegraphics[width=0.98\linewidth]{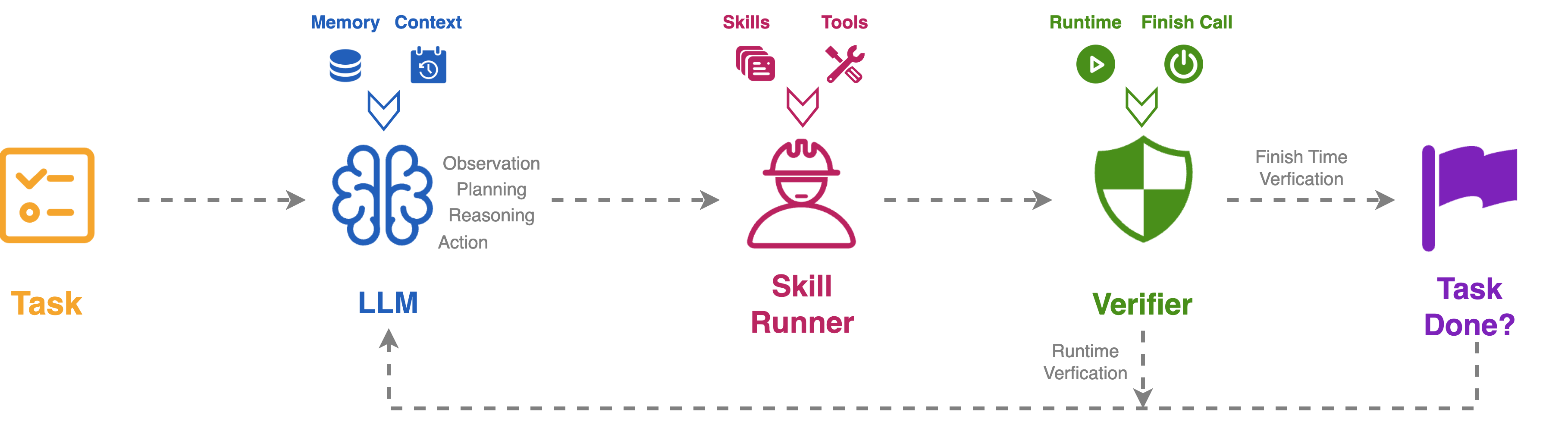}
\caption{Overview of the Agent Harness. The main LLM performs scene-conditioned planning with memory and context, delegates procedural subtasks to the Skill Runner, and receives corrective feedback from the Verifier to form a reasoning-execution-verification loop.}
\label{fig:agentworkflow} 
\end{figure*}

% \subsubsection{Overview}

We propose a hierarchical LLM agent framework for long-horizon embodied tasks. The central motivation is that an embodied agent must not only interpret a natural-language instruction, but also understand the scene in which the instruction is grounded, the current execution stage, and the conditions under which the task can be considered complete. Unlike code agents and ordinary tool-use agents, such as SWE-agent for software engineering through agent-computer interfaces~\cite{yang2024sweagent}, Toolformer for API-based tool use~\cite{schick2023toolformer}, and ReAct for interleaved reasoning and acting~\cite{yao2022react}, embodied agents often lack explicit completion signals for intermediate steps. An agent may invoke a navigation command without actually leaving its current location, or rotate and collide repeatedly while still believing, at the language level, that the task is progressing. Our framework therefore does not simply make the LLM call more tools. Instead, it organizes embodied execution as a closed loop of reasoning, execution, and verification.

The framework separates three roles that are often entangled in a single
controller: the main LLM, the Skill Runner, and the Verifier.
As shown in Figure~\ref{fig:agentworkflow}, the task is first processed by a main LLM. Supported by memory and context, the main LLM observes the task, interprets it with respect to the current scene, and forms a high-level plan. It then decides whether to directly invoke a tool or delegate a multi-step subtask to the Skill Runner. The Skill Runner is a skill-level subagent that manages procedural execution in an isolated local context, maintains intermediate state, handles repeated observations and actions, and returns a compressed outcome to the main LLM. The Verifier supervises execution both during runtime and at finish time, checking whether the behavior is still making progress, whether a skill has become stagnant, and whether the task is truly ready to terminate. This loop grounds LLM reasoning in environment facts and reduces procedural drift and premature termination in long-horizon embodied tasks.

\subsubsection{Scene-Conditioned Task Planning}

The main LLM operates as the semantic planner in the Agent Harness. Its
role is to interpret the user instruction with respect to the current
scene, map memory, robot state, recent history, and available skills,
rather than to issue every low-level movement. Before acting, it
determines whether the task requires navigation, search, human
interaction, reporting, manipulation, or additional observation, and it
forms a revisable high-level plan with explicit completion conditions.

This planning step is scene-conditioned rather than purely linguistic.
The same instruction may require different execution strategies
depending on the agent's current location, available visual evidence,
known map structure, reachable regions, nearby objects, and interaction
history. The main LLM therefore reasons about what information is
already sufficient, what must be observed or queried, which skills are
feasible in the current state, and what observable conditions would
constitute completion.

The plan is updated as new observations, tool results, skill summaries,
and verifier feedback enter the context. The main LLM decides which goal
should be pursued, when direct tool use is sufficient, when a subtask
should be delegated to the Skill Runner, and when the task may be ready
to finish. By keeping local execution details out of the main reasoning
thread, the agent preserves a coherent global task state while allowing
lower-level skills to handle procedural detail.

\subsubsection{Skill Runner for Procedural Execution}
The Skill Runner handles subtasks that require sustained local
execution. A skill is not treated as a one-shot tool call or a simple
action macro; it is executed by a skill-level subagent with an isolated
local context that includes the subgoal, recent observations, skill
state, failed attempts, and recovery strategy. The main LLM receives
compressed skill progress and outcomes rather than the full sequence of
intermediate actions.

This context isolation is important for long-horizon embodied tasks,
where local execution may involve repeated movement, observation,
relocalization, view adjustment, and recovery. If every local collision,
short-range movement, failed attempt, or visual adjustment were appended
to the main LLM context, the global objective would be obscured by
procedural detail. The Skill Runner absorbs this local complexity while
preserving the information needed by the main LLM for higher-level
stage management.

During execution, the Skill Runner maintains procedural continuity by
checking whether progress is being made, deciding when additional
observation or local recovery is needed, and determining when control
should return to the main LLM. At termination, it returns a compact
semantic summary indicating whether the subgoal was achieved, why it
failed if it did, what scene information was discovered, what recovery
was attempted, and how the main LLM should use the outcome.

\subsubsection{Multi-Stage Verification}
The Verifier addresses the mismatch between language-level belief and
environment-grounded completion. Unlike many digital tasks, where
success can be checked through explicit external signals such as tests,
API responses, or page states, embodied tasks often require consistency
between the agent's declared progress, the execution trajectory, and the
observed scene state. The Verifier therefore supervises execution by
checking whether recent behavior, skill outcomes, and final answers are
supported by environment evidence rather than by the agent's belief
alone.

Verification is applied at three stages. Runtime verification monitors
whether the recent trajectory and skill state indicate effective
progress or reveal stagnation, repeated collisions, local loops, or
behavior inconsistent with the current plan. Skill-level verification
checks whether a delegated subtask has satisfied its semantic objective,
rather than accepting success solely because a tool or subagent returned
normally. Finish-time verification is performed when the main LLM
attempts to terminate the task; it evaluates the original instruction,
current plan, execution history, skill summaries, observations, and any
new requirements introduced during interaction before allowing the task
to finish.

The Verifier is therefore not merely a final evaluator, but a
supervisory signal inside the Agent Harness. By applying verification
before, during, and after delegated execution, ABot-AgentOS can return
missing conditions to the reasoning layer and reduce stagnation,
misjudgment, and premature termination in open-ended embodied tasks.

\subsubsection{Edge-Cloud Collaborative Routing}

In practical deployment, not every task should be handled by a cloud-scale model. Routine requests can often be completed by an on-device small model and local tools, while long-horizon embodied tasks may require stronger scene understanding, planning, skill execution, and verification. ABot-AgentOS therefore uses an edge-cloud collaborative routing layer: the on-device model first observes the task and context, then either handles the request locally or escalates it to a cloud model for more complex reasoning.

This design follows the general motivation of model routing and hierarchical inference, where systems select among models with different capabilities and costs~\cite{chen2023frugalgpt,ong2024routellm,anthropic2026advisortool}. In our setting, the routing policy is learned from training samples and execution feedback rather than fixed rules. It captures which requests can be reliably solved with local tools, which need cloud-level planning, and which require additional observation before routing. This keeps latency and deployment cost under control while preserving cloud-scale capability for tasks that need deeper long-horizon reasoning.

\begin{figure}[!t]
    \centering
    \includegraphics[width=1\linewidth]{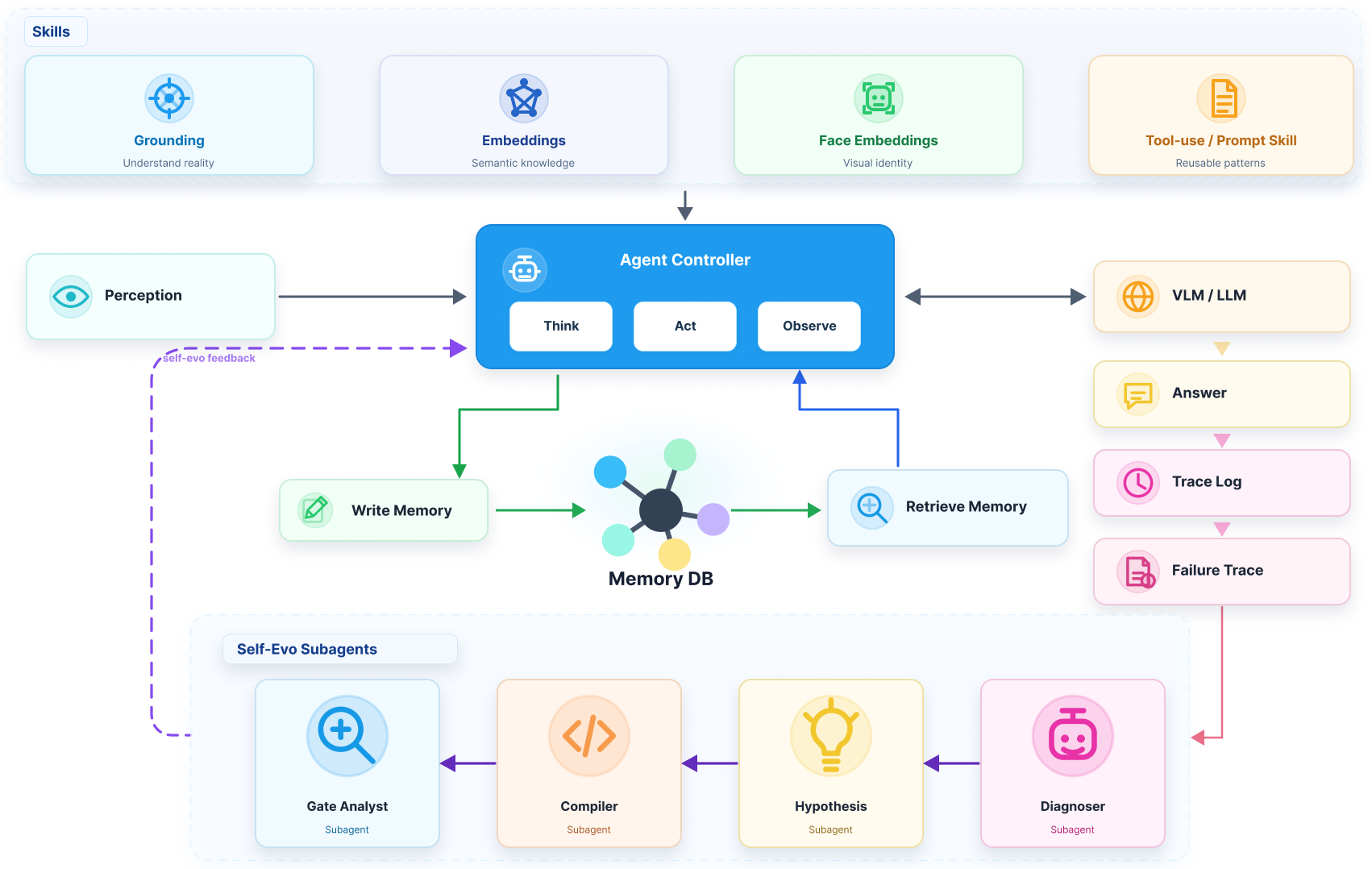}
    \caption{
    Overview of the multi-modal memory architecture.
    During online execution, ABot-AgentOS writes observations and interactions into a source-grounded memory graph, retrieves task-relevant evidence, and records retrieval and answer traces.
    Offline, failure traces are diagnosed and converted into gated runtime evo-assets for later deployments.
    }
    \label{fig:memory_architecture}
\end{figure}

\subsection{Memory System}
\label{sec:memory_system}

Long-horizon embodied interaction requires a memory substrate that is external to the LLM context, persistent across sessions, and grounded in the robot's own experience.
In ABot-AgentOS, the memory system provides this substrate.
It records observations, human--robot dialogue turns, model-extracted semantic facts, visual evidence, identities, object states, spatial relations, and task traces as source-grounded memory records, and makes the accumulated experience available to later reasoning steps through traceable retrieval.
Rather than serving as a raw archive of complete videos, full dialogue transcripts, or unstructured image payloads, the memory system converts multi-modal experience into compact graph records with typed entities, events, relations, temporal context, and provenance.
This design allows the agent to recall people, animals, objects, places, events, temporal facts, spatial relations, and source evidence without overloading the online context window.

The memory system is positioned beneath the online runtime and complements the context management module.
Context management decides what information should be placed in the current prompt, including the current instruction, role strategy, execution plan, recent observations, and reflection summaries.
Memory decides what should persist beyond the current episode and how relevant past evidence should be retrieved back into context.
This separation prevents long-term memory from degenerating into an unstructured prompt dump while still allowing the agent controller to use persistent experience when the current task requires identity recall, temporal grounding, object-location recall, or source-backed visual evidence.

A practical robotic memory must satisfy three requirements. First, it
must be multi-modal, because useful experience is distributed across
language, egocentric vision, object states, spatial layouts, identities,
and task traces. Second, it must be relational, because embodied recall
often depends on who participated in an event, where an object was last
seen, what changed over time, or which frame supports a claim. Third, it
must be auditable and improvable: each answer should expose its evidence
so failures can be attributed to memory writing, evidence selection,
temporal grounding, visual matching, or answer composition.

\subsubsection{Graph Memory Representation}
\label{sec:memory_representation}

We represent long-term experience as a typed graph
\begin{equation}
    \mathcal{G} = (\mathcal{V}, \mathcal{E}),
\end{equation}
where each node $v \in \mathcal{V}$ denotes an entity or evidence unit, and each edge $e \in \mathcal{E}$ denotes a temporal, semantic, spatial, identity, interaction, or provenance relation.
Each node stores a compact JSON-style information field containing schema version, dataset or source reference, time reference, evidence summary, confidence, adapter-specific fields, and provenance.
Provenance records the source id, adapter version, extractor model, and related extraction metadata, while temporal information is represented through fields such as \texttt{time\_ref}, \texttt{created\_at}, or adapter-specific fields.
Together these fields allow a memory item to be traced back to the observation, video segment, frame, image, or session from which it was created.

The schema includes nodes for source containers, evidence units, entities, places, sessions, and semantic events.
Edges encode the relations needed for embodied recall, including temporal order, containment, observation, participation, location, identity continuity, spatial relations, and interaction relations.
The schema is intentionally platform-agnostic: it stores semantic experience and evidence rather than hardware-specific control details, so the same memory interface can be reused across robot embodiments.

This representation differs from conventional retrieval-augmented generation over text chunks.
Text retrieval mainly returns semantically similar snippets, whereas the memory graph first identifies relevant seed nodes and then expands to a local evidence subgraph for reasoning over identity, time, location, participation, provenance, and spatial relations.
As a result, the agent can answer questions about object locations, identity continuity, or relative time using structured evidence rather than parametric memory alone.

\subsubsection{Multi-modal Memory Updating}
\label{sec:memory_writing}

Memory updating can occur during interaction and at reflection checkpoints.
In an online ABot-AgentOS deployment, the controller receives observations, tool results, dialogue turns, and execution traces, while a memory-writing service selects information from these streams and converts it into compact graph records linked to session, place, time, source reference, and provenance.
We instantiate this writer as a set of graph-construction adapters for different data sources and applications.
This design does not modify the robot action space; instead, it augments the agent's reasoning state with persistent, source-grounded evidence that can be retrieved in later tasks.

\begin{figure}[!t]
    \centering
    \includegraphics[width=1\linewidth]{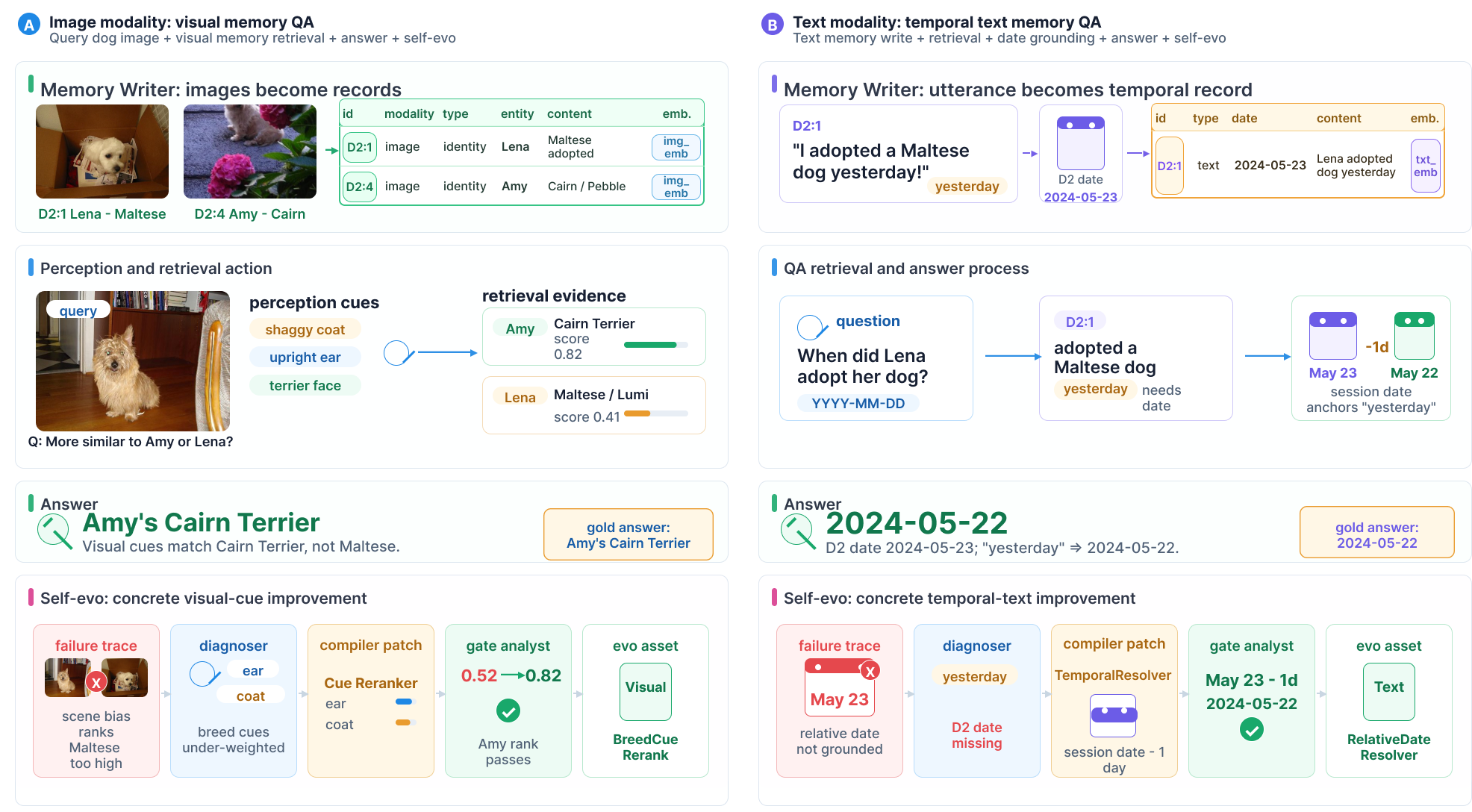}
    \caption{
    Concrete memory failure-to-evolution examples.
    Left: visual memory QA retrieves image-grounded identity evidence but can expose missing breed-specific cues.
    Right: temporal text memory QA uses session metadata to resolve relative dates but can reveal temporal-normalization errors.
    In both cases, the failure trace is converted into targeted memory-writing, evidence-selection, frame-selection, or answering improvements.
    }
    \label{fig:mem_example}
\end{figure}

To support abstraction across modalities, different inputs are normalized into the same source-grounded graph schema.
Video and egocentric streams produce semantic records for visible entities, object states, places, actions, frames, and events; dialogue and multi-modal sessions produce session-level and event-level records, with image attachments stored as evidence nodes connected to the corresponding semantic event.
Across these inputs, the writer favors semantic compression over raw full-turn or full-frame storage, following the principle that long-term memory should increase information density without sacrificing traceability.
Recent work such as SimpleMem further shows that structured semantic compression and query-aware retrieval can improve memory efficiency under limited context budgets~\cite{simplemem}.
In ABot-AgentOS, this principle is instantiated by converting raw observations into typed, source-grounded nodes and edges that can later be queried by content, time, entity mention, modality, provenance, and relation.
For instance, the utterance ``I adopted a Maltese dog yesterday'' is represented as a time-grounded semantic event that links the utterance, resolved temporal context, identity hypothesis, confidence estimate, and source evidence, rather than as a raw transcript line.
When an image is attached, the image node is connected to the event as supporting multi-modal evidence.
This conversion allows multi-modal observations to support temporal grounding, identity resolution, provenance-aware retrieval, and graph-based reasoning.
Concrete examples of how such records support retrieval and later failure-driven improvement are shown in Figure~\ref{fig:mem_example}.

The same principle also governs what is written and how the graph is maintained over time.
Not every raw observation becomes a permanent memory item: the writer prioritizes information likely to support future embodied or interaction tasks, including identities, object locations, state changes, social commitments, user preferences, abnormal events, temporal facts, and evidence needed for later verification.
After insertion, ABot-AgentOS performs lightweight maintenance to keep the graph compact over long horizons.
Near-duplicate entity or event nodes are merged when they share compatible provenance, temporal context, and identity evidence; frequently observed entities keep compact state summaries with a bounded set of representative evidence nodes.
Stale state facts are superseded by newer observations through temporal edges rather than deleted immediately, allowing retrieval to distinguish current state from historical evidence.
Together, selective writing and post-insertion maintenance reduce storage cost, avoid redundant low-level records, and preserve enough source-grounded evidence for later auditability.

\subsubsection{Retrieval and Grounded Answering}
\label{sec:memory_retrieval}

At runtime, memory retrieval is invoked when the current task requires information beyond the immediate observation or short-term context.
The query is converted into semantic retrieval signals and, when available, lightweight structural cues such as entity mentions, temporal expressions, modality, place references, and expected evidence type.
The retriever first selects candidate seed nodes using semantic embeddings, lexical matching, metadata filters, source constraints, and node-type constraints.
Given a query $q$, this hybrid seed selection can be written as
\begin{equation}
    s(q,v) =
    \lambda_{\mathrm{sem}} s_{\mathrm{sem}}(q,v)
    + \lambda_{\mathrm{lex}} s_{\mathrm{lex}}(q,v)
    + \lambda_{\mathrm{meta}} s_{\mathrm{meta}}(q,v)
    + \lambda_{\mathrm{type}} s_{\mathrm{type}}(q,v),
    \label{eq:hybrid_retrieval_score}
\end{equation}
where $s_{\mathrm{sem}}$ denotes embedding similarity, $s_{\mathrm{lex}}$ denotes lexical overlap, $s_{\mathrm{meta}}$ measures metadata compatibility such as time, source, modality, or place, and $s_{\mathrm{type}}$ favors node types expected by the query.
The top-ranked seeds are then expanded along task-relevant typed edges under a fixed depth and evidence-token budget, producing a local evidence subgraph that contains the semantic, temporal, spatial, identity, and provenance context needed for answering.
This differs from retrieval over independent memory stores, as in dynamic multi-modal retrieval agents such as WorldMM~\cite{worldmm}: ABot-AgentOS retrieves seed nodes and then expands a typed graph neighborhood, so the answerer can reason over relations among evidence items rather than over isolated snippets.
The resulting subgraph is serialized as a compact evidence context, paired with a retrieval trace, and injected into the agent's reasoning context.

% This retrieval process is complementary to dynamic multi-modal retrieval agents such as WorldMM, which adaptively selects among episodic, semantic, and visual memories at different temporal granularities for long-video reasoning \cite{worldmm}.
% Instead of selecting among separate memory stores, ABot-AgentOS performs query-dependent seed retrieval followed by typed graph expansion, allowing the answerer to reason over a local evidence subgraph containing semantic, temporal, spatial, identity, and provenance evidence.

The answerer is instructed to ground its response in retrieved evidence.
This is particularly important for embodied and interaction-memory questions, where the correct answer often depends on local experience, temporal grounding, or visual evidence rather than general world knowledge.
When evidence is insufficient or contradictory, the answerer is expected to report uncertainty instead of hallucinating an unsupported fact; in forced-choice benchmark settings, it still records the evidence limitation in the trace.

Each memory-augmented QA run records a retrieval trace containing the retrieved nodes and edges, source references, evidence summaries, retrieval stages, and ranking signals.
In online deployment, this trace makes the answer inspectable and can help the controller decide whether additional observation or clarification is required.
Offline, it provides the diagnostic signal for failure analysis: an incorrect answer can be attributed to missing memory writing, failed retrieval, evidence misuse, or unresolved temporal, spatial, relation-direction, or entity-grounding operations.

\subsubsection{Failure-Driven Lifelong Self-Evolution}
\label{subsubsec:memory-self-evolution}

% Recent work on self-evolving agents argues that deployed agent systems should not remain static, but should improve their components from interaction data, environmental feedback, and test-time trajectories \cite{selfevolvingsurvey}.
% This view is particularly important for long-term memory: if only the stored contents are updated while the writer, evidence-selection policy, and answerer remain fixed, the memory system can repeatedly make the same extraction, retrieval, grounding, or reasoning mistakes.
% Evo-Memory formulates this issue as a streaming test-time learning problem, where agents must continuously retrieve, integrate, update, and reuse memory across evolving task sequences \cite{evomemory}.
% Recent self-evolving memory systems further show that memory evolution can target different levels of the pipeline, including retrieval configurations, memory architectures, procedural memories, memory skills, and multi-agent memory-cycle coordination \cite{evolvemem,memevolve,reme,memskill,memma}.
% In addition, evidence-gap-driven and evidence-verifiable self-evolution highlights the need to diagnose missing evidence and only learn from source-grounded, auditable signals \cite{evimem,eveagent}.

Long-term memory should evolve not only by accumulating new content, but also by improving the pipeline that writes, retrieves, selects, and uses evidence.
Otherwise, the system may repeatedly make the same extraction, retrieval, temporal-grounding, visual-selection, or answer-composition errors even as the memory graph grows.
This view is aligned with recent work on self-evolving agents, streaming memory evolution, and evidence-verifiable improvement, which emphasizes learning from interaction traces while keeping updates auditable and bounded \cite{selfevolvingsurvey,evomemory,evolvemem,memevolve,reme,memskill,memma,evimem,eveagent}.

Figure~\ref{fig:memory_architecture} shows the online memory components and the offline self-evolution loop at a system level, while Figure~\ref{fig:mem_example} gives concrete examples of how memory failures become targeted evo-assets.
We formalize this process as a split-wise protocol over long-term deployment, or its benchmark approximation, represented by an ordered sequence of disjoint splits
$\mathcal{D}_1,\ldots,\mathcal{D}_T$.
At split $t$, ABot-AgentOS maintains two persistent states: the accumulated memory graph $G_{t-1}$ and the promoted evo-asset set $A_{<t}$.
The graph stores content-level experience, including entities, events, places, visual evidence, temporal context, and provenance, while evo-assets store pipeline-level improvements for memory writing, evidence selection, graph expansion, answer composition, frame selection, temporal normalization, or adapter-level normalization.
During online execution, the agent writes new observations into memory, retrieves evidence for downstream answering, and records retrieval and answer traces under the currently promoted asset set.
Only after split $t$ has been evaluated are failed cases converted into failure traces and processed for possible improvement.
No evo-asset generated from split $t$ is used during inference on split $t$; accepted assets are promoted only for later splits.
This no-leakage constraint turns self-evolution into a cumulative lifelong process rather than a one-shot post-hoc repair.

% Figure~\ref{fig:memory_architecture} shows the online memory components and the offline self-evolution loop at a system level, while Figure~\ref{fig:mem_example} gives concrete examples of how memory failures become targeted evo-assets.
% We formalize the split-wise protocol here without introducing an additional figure.
% We view long-term deployment, or its benchmark approximation, as an ordered sequence of disjoint splits
% $\mathcal{D}_1,\ldots,\mathcal{D}_T$.
% At split $t$, the online memory system maintains two persistent states: the accumulated memory graph $G_{t-1}$ and the promoted evo-asset set $A_{<t}$.
% The memory graph stores content-level experience, including entities, events, places, sessions, visual evidence, temporal context, and provenance.
% The evo-assets store pipeline-level improvements, including writer rules, retrieval-side policies for query normalization, evidence filtering, edge-expansion priority, and evidence serialization within the fixed hybrid retrieval architecture, answerer instructions, frame-selection policies, temporal-normalization policies, and dataset- or application-specific normalization rules.

% During online execution, the agent writes new observations into memory, retrieves evidence for downstream answering, and records retrieval and answer traces.
% After split $t$ has been evaluated, failed cases are converted into failure traces and processed by the offline self-evolution loop.
In benchmark settings, failures are identified using ground-truth answers only after the split is complete; in deployment, analogous feedback can come from environmental signals, task success or failure, human correction, or low-confidence retrieval and answering traces.
Only assets that pass the gate are promoted and used in later splits.
No evo-asset generated from split $t$ is used during inference on split $t$.
This split-wise protocol prevents evo-assets from using supervision from the same split on which they are evaluated, turning self-evolution into a cumulative lifelong process rather than a one-shot post-hoc repair.

Formally, the online run, asset proposal, and asset promotion steps are:
\begin{align}
    \mathcal{T}_t, G_t
    &= \operatorname{Run}(\mathcal{D}_t; G_{t-1}, A_{<t}), \\
    \Delta A_t
    &= \operatorname{Gate}\!\left(
        \operatorname{Compile}\!\left(
        \operatorname{Propose}\!\left(
        \operatorname{Diagnose}(\mathcal{T}_t)
        \right)\right)\right), \\
    A_{\leq t}
    &= A_{<t} \cup \Delta A_t .
    \label{eq:self_evo_update}
\end{align}
Here $\mathcal{T}_t$ denotes the answer traces and failure traces collected on split $t$.
Only $A_{\leq t}$ is available for later splits $\mathcal{D}_{t+1},\ldots,\mathcal{D}_T$.

% Within each split, ABot-AgentOS runs the normal online memory pipeline without changing the base model or the active asset set.
% After the split, the collected traces are used by an offline multi-agent self-evolution loop.
% When supervision is available, a failed QA attempt is converted into a failure trace containing the question, retrieved evidence, prediction, expected answer, retrieval trace, and dataset context.
% A Diagnoser analyzes the failure source, distinguishing among memory-writing errors, evidence-selection misses or ranking errors, answer-composition errors, frame-selection errors, temporal-normalization errors, entity or subject mismatches, relation-direction errors, and adapter-specific errors.
% A Hypothesis Generator proposes candidate fixes.
% A Compiler-Critic converts safe fixes into a constrained JSON DSL asset.

After each split, the collected traces are processed by a constrained failure-to-asset loop.
When supervision is available, a failed QA attempt is represented by the question, retrieved evidence, prediction, expected answer, retrieval trace, and dataset context; in deployment, analogous feedback can come from environmental signals, task outcomes, human correction, or low-confidence traces.
The loop diagnoses the failure source, proposes a candidate repair, compiles the repair into a constrained JSON DSL evo-asset, and evaluates it with both target and regression checks.
This follows the broader principle of safe resource evolution: prompts, tools, memory modules, and other agent resources should be versioned, auditable, and rollbackable rather than updated through unconstrained code generation \cite{autogenesis}.

% Finally, a Gate Analyst evaluates the candidate on the target and regression splits and accepts it only if it improves target behavior without introducing unacceptable regressions.
% This design follows the broader principle of safe resource evolution: prompts, tools, memory modules, and other agent resources should be versioned, auditable, and rollbackable rather than updated through unconstrained code generation \cite{autogenesis}.

% \begin{table}[!t]
%     \centering
%     \small
%     \begin{tabular}{p{0.20\linewidth}p{0.33\linewidth}p{0.35\linewidth}}
%         \toprule
%         \textbf{Agent} & \textbf{Input} & \textbf{Output} \\
%         \midrule
%         Diagnoser &
%         Failure trace, retrieved evidence, expected answer, dataset context &
%         Failure type and root-cause explanation. \\

%         Hypothesis Generator &
%         Diagnosis and memory-pipeline context &
%         Candidate repair hypothesis targeting writer, evidence selection, answerer, frame policy, or adapter behavior. \\

%         Compiler-Critic &
%         Candidate hypothesis and safety constraints &
%         Constrained JSON DSL evo-asset with no generated Python execution. \\

%         Gate Analyst &
%         Candidate asset, target subset, regression subset &
%         Accept/reject decision, validation report, and regression report. \\
%         \bottomrule
%     \end{tabular}
%     \caption{Roles in the offline self-evolution loop.}
%     \label{tab:self_evo_agents}
% \end{table}

The gate evaluates each candidate asset on a target validation subset and a regression subset.
An asset $a$ is accepted only if it improves the target score by at least $\tau_{\mathrm{gain}}$ and does not reduce the regression score by more than $\tau_{\mathrm{reg}}$:
\begin{equation}
    \operatorname{Accept}(a)=
    \mathbb{I}\left[
    \Delta S_{\mathrm{target}}(a) \geq \tau_{\mathrm{gain}}
    \land
    \Delta S_{\mathrm{reg}}(a) \geq -\tau_{\mathrm{reg}}
    \right].
    \label{eq:gate_criterion}
\end{equation}
This criterion makes promotion conservative: an asset must improve the failure pattern it targets while preserving behavior on previously reliable cases.

Accepted evo-assets are lifecycle-managed JSON DSL records and do not execute generated Python code.
Each asset declares its target layer, triggering condition, permitted action, safety constraints, provenance, validation result, and version id.
Assets may target memory writing, evidence selection, answer composition, visual frame selection, temporal normalization, or adapter-level normalization.
Normal QA behavior is unchanged unless accepted assets are explicitly loaded; writer-side and frame-policy assets require fresh graph construction, whereas evidence-selection and answerer assets can be loaded at runtime.
In this way, ABot-AgentOS accumulates two forms of lifelong knowledge: content-level experience in the memory graph and pipeline-level improvements in the evo-asset set.

\subsubsection{Edge-Cloud Collaborative Memory Management}
\label{sec:memory-sync}

Embodied agents continuously acquire heterogeneous memories during interaction with the physical world, including maps, semantic landmarks, obstacles, objects, and human-related observations. 
While public environmental memories, such as roadblocks, construction areas, and navigational landmarks, can improve multi-agent collaboration when shared, privacy-sensitive memories, such as faces, personal belongings, or person-associated objects, must remain local. 
To address this conflict, we design an edge-cloud collaborative memory system that separates private memory on the robot from common memory in the cloud.

\paragraph{Memory Partition}
Each robot maintains a private memory at the edge, which stores all locally perceived memories, including map memory, semantic memory, and multi-modal memory. 
The cloud maintains a common memory that only contains public, low-sensitivity memories useful for collective navigation and task execution. 
This design follows a private-by-default principle: a memory item is never uploaded unless it is explicitly classified as shareable.

\paragraph{Privacy-aware Memory Gating}
For each newly generated memory item, the system performs a privacy judgment before synchronization. 
A memory is regarded as non-shareable if it contains or is associated with personally identifiable information, including persons, faces, names, personal objects, or ownership relations to a specific individual. 
In contrast, public environmental information, such as maps, traffic cones, barriers, road damage, and static landmarks, is classified as shareable.

To evaluate the reliability of the privacy protection mechanism, we constructed a dedicated dataset for privacy classification and upload-decision testing. Experimental results show that the system achieves over 99\% accuracy in identifying whether a memory item is suitable for cloud synchronization. This high-accuracy privacy judgment significantly reduces the risk of uploading sensitive user or environment-specific information, while still allowing non-sensitive public knowledge to be shared across robots. As a result, the proposed memory system balances two key requirements of embodied intelligence: preserving user privacy through edge-side private memory, and enhancing group-level collaboration through cloud-side common memory.

% \subsubsection{Summary}
% \label{sec:memory_summary}

% The memory system provides three advantages for a general robotic AgentOS.
% First, it supports cross-modal binding: entities and evidence records can be linked to dialogue mentions, visual appearances, embeddings, locations, temporal events, image attachments, and source evidence.
% Second, it supports relational recall: the agent can answer questions that require following edges across time, space, identity timelines, participation, provenance, and spatial relations rather than retrieving isolated text chunks.
% Third, it supports lifelong improvement: because each memory-augmented QA run is associated with a retrieval trace, failures can be attributed to specific stages of the memory pipeline and converted into gated runtime assets.
% The split-wise lifelong protocol carries accepted evo-assets from earlier splits to later splits, so ABot-AgentOS accumulates both embodied experience in the memory graph and reusable pipeline improvements in the asset set.
% Thus, memory becomes not only a persistent record of embodied and interaction experience, but also the substrate through which ABot-AgentOS improves across long interaction horizons.
% The memory graph accumulates what the robot has experienced, while the evo-asset set improves how that experience is written, selected, grounded, and used.

% \input{sections/2_4_memory}
\section{EmbodiedWorldBench}  
\label{sec:benchmark}  

Recent embodied AI benchmarks have advanced along two largely independent tracks.
On the indoor side, BEHAVIOR-1K~\cite{li2024behavior1k} defines 1{,}000 everyday household activities,
EmbodiedBench~\cite{yang2025embodiedbench} evaluates multi-modal LLM-driven agents across multiple indoor simulators,
and LongAct~\cite{zhu2026robotchores} targets long-horizon domestic task execution---yet
all are confined to bounded indoor environments and cannot assess large-scale spatial reasoning in open spaces.
On the outdoor side, EmbodiedCity~\cite{gao2024embodiedcity} constructs a city-scale 3D simulation,
CityNav~\cite{lee2025citynav} and OpenFly~\cite{gao2025openfly} provide large-scale aerial VLN datasets,
CityNavAgent~\cite{zhang2025citynavagent} and GeoNav~\cite{xu2025geonav} advance LLM-powered aerial navigation planning,
while GeNIE~\cite{wang2025genie} and ODYSSEY~\cite{wang2025odyssey} establish ground-level outdoor navigation and mobile manipulation benchmarks, respectively---yet
these works predominantly focus on a single task modality (point-to-point navigation or isolated manipulation), lacking social interaction and multi-step reasoning.
TongSIM~\cite{sun2025tongsim}, Wanderland~\cite{liu2026wanderland}, and Embodied Web Agents~\cite{hong2025embodiedweb} attempt to unify indoor and outdoor evaluation
but still lack structured multi-difficulty task generation, strict information isolation, or dynamic NPC-driven events.
Overall, existing benchmarks share three common limitations:
(1)environmental fragmentation---indoor and outdoor scenes are evaluated in isolation;
(2)task homogeneity---only a single capability dimension is tested;
(3)static evaluation---the absence of dynamic events precludes assessment of adaptive replanning.

To address these gaps, we propose a unified embodied evaluation framework spanning 16 executable scenes across indoor, outdoor, and hybrid settings, with tasks organized into four difficulty levels. Unlike existing work, each evaluation case is a complete executable scenario that jointly tests navigation, object search, NPC dialogue, dynamic instruction response, and state tracking. The framework enforces a strict information visibility boundary: the agent receives only a filtered semantic map and must rely on autonomous perception and reasoning to complete tasks. This work is the first to integrate cross-domain continuity, social dynamics, and engineering rigor into a single evaluation system.

\subsection{Overview}
\label{sec:overview}
Our benchmark is designed to evaluate embodied agents in a structured, reproducible, and multi-dimensional manner. Unlike traditional static datasets, every evaluation case is a complete, runnable world—a self-contained scenario defined as:
\begin{equation}
    \text{Scenario} = \langle \mathcal{M}, \mathcal{S}_0, \mathcal{O}, \mathcal{N}, \mathcal{C} \rangle
\end{equation}
where $\mathcal{M}$ is the spatial map, $\mathcal{S}_0$ the initial state, $\mathcal{O}$ the observation rules, $\mathcal{N}$ the NPC behaviors, $\mathcal{C}$ the success criteria. 

\begin{figure}[!t]
    \centering
    \includegraphics[width=1\linewidth]{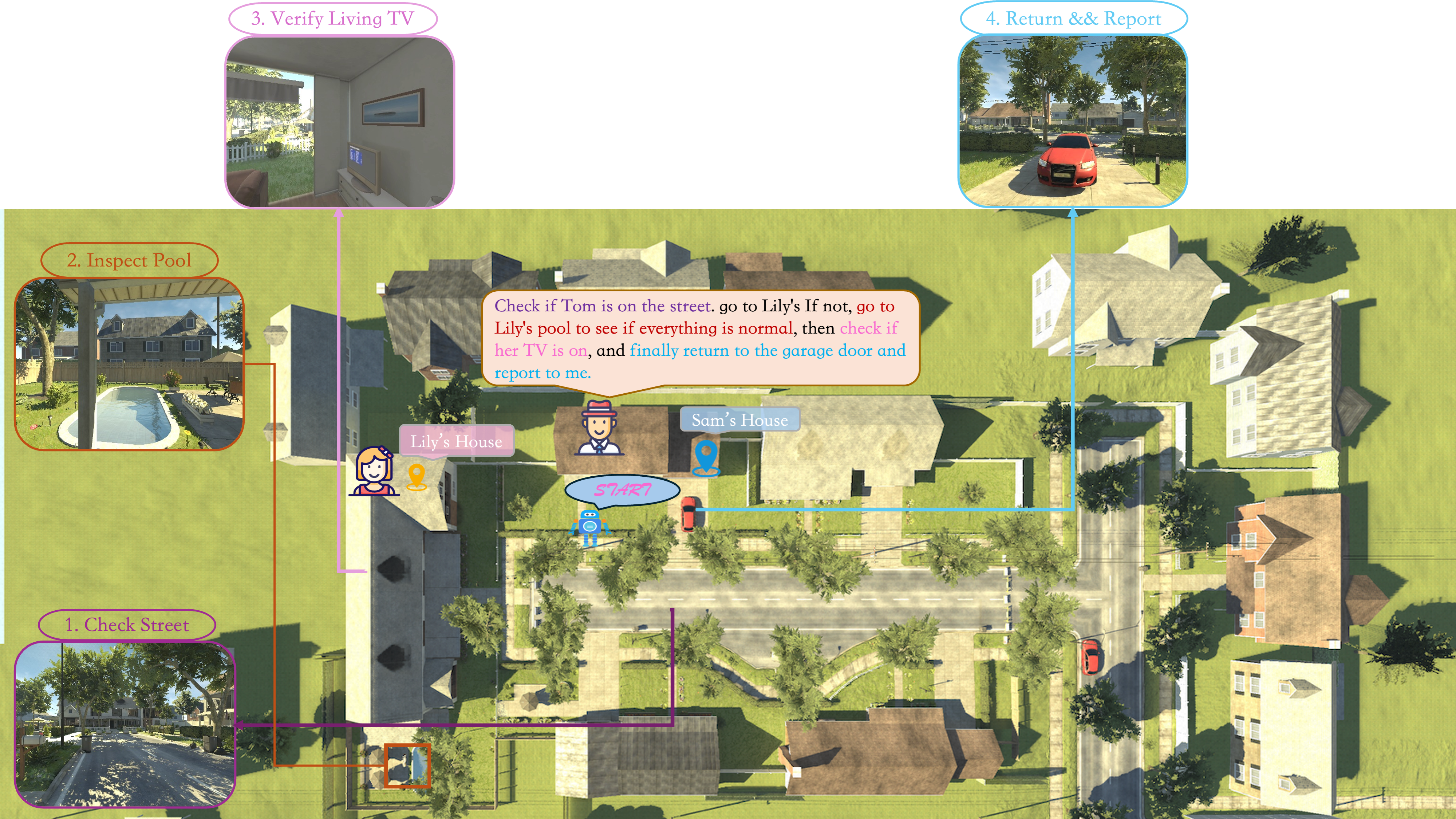}
    \caption{\textbf{Overview of EmbodiedWorldBench.} EmbodiedWorldBench evaluates embodied agents on compound tasks that span indoor and outdoor spaces and require tightly coupled navigation, NPC interaction, and environment perception across diverse scenes. The benchmark covers 16 scenes across four difficulty levels with over 200 tasks, revealing the challenges of achieving cross-scene generalization and adaptive replanning under dynamic events.}
    \label{fig:overview}
\end{figure}

Figure~\ref{fig:overview} illustrates a representative scenario in a
suburban neighborhood. Given a compound instruction, the agent must
sequentially complete four sub-goals---checking a street, inspecting
a backyard pool, verifying an indoor appliance state, and returning
to report---with the task path traversing outdoor streets, a
residential backyard, and an indoor living room within a single
episode. This example demonstrates how each scenario jointly exercises
cross-scene navigation, NPC interaction, environment perception, and
information reporting, rather than isolating a single skill dimension.

The benchmark is guided by four design principles. First, each case must be an executable scenario rather than a standalone query. Second, a single scenario should jointly test multiple embodied capabilities instead of isolating a single skill dimension. Third, all outcomes must be trace-grounded and auditable from the execution trace, thereby avoiding subjective human scoring during evaluation. Fourth, benchmark execution must be reproducible: under the same scenario configuration and model setup, the evaluation trajectory and final score should be stable and repeatable. Taken together, these principles position the benchmark as an end-to-end embodied evaluation suite rather than a prompt-collection benchmark.

\subsection{Benchmark Construction}
\label{sec:construction}

\subsubsection{Environment and Semantic Map Annotation}
\label{sec:env-semantic}

We use UnrealZoo, a collection of photo-realistic 3D environments built on Unreal Engine 5. It provides NavMesh-based navigation, programmable object spawning, and a diverse set of scenes spanning indoor, outdoor, and residential settings.

To bridge the gap between raw engine-level objects and task-level semantics, we launch the UE5 scene and have annotators freely navigate the environment from a first-person perspective. They traverse all reachable areas, manually marking and naming key objects, points of interest (POIs), interaction targets, and typical navigation paths relevant to the evaluation agent, thereby constructing a structured semantic map. This map serves as the foundation for all subsequent task design and query generation.

\subsubsection{Query Generation}
\label{sec:query-gen}

A \textit{query} is the natural-language instruction issued to the agent (e.g., \textit{``find the elderly person in a red down jacket and report her location''}), paired with formal success criteria, NPC configurations. To enable scalable construction of embodied tasks, we propose a task generation framework.

\textbf{Semantic Map Normalization.} Through a normalization pipeline, raw annotated waypoints in the semantic map are structurally transformed into a unified representation comprising three layers: point, room, and polygon. The point layer records spatial coordinates, semantic types, floor and room associations, and object visual attributes. The room layer aggregates points within the same area to form a semantic topology. The polygon layer automatically generates enclosed spatial regions for point membership determination. The framework currently covers 16 executable scenes across indoor, outdoor, and hybrid settings, including hospitals, museums, supermarkets, and suburban neighborhoods, with over 300 annotated waypoints in total.

% \textbf{Multi-difficulty Task Generation.} Based on the normalized semantic map, the framework automatically extracts a scene capability summary, including available regions, NPC spawn points, and key object inventories. This summary is fed into a large language model along with 8 predefined task families (covering navigation, object search, NPC dialogue, dynamic instruction response, state tracking, etc.), which simultaneously generates both task structures and natural language descriptions. To support stratified capability evaluation, generated tasks are organized into four difficulty levels. As difficulty increases, the required scene coverage, NPC interaction complexity, and reasoning depth increase accordingly. Considering differences in scene scale and task characteristics, the benchmark explicitly controls the distribution of task instances across scene and difficulty dimensions to ensure overall coverage and evaluation representativeness. The resulting distribution is summarized in Table~\ref{tab:task_distribution}.

\textbf{Multi-difficulty Task Generation}. Based on the normalized semantic map, the framework automatically extracts a scene capability summary, including available regions, NPC spawn points, and key object inventories. For each scene, annotators first specify a set of feasible task types grounded in the scene's layout, objects, NPCs, and available interactions. This scene-specific task specification is then provided to a large language model, which generates both simple tasks and compound multi-stage tasks together with their natural-language descriptions and formal success criteria. Human annotators then review and refine the generated candidates to ensure executability, semantic clarity, evaluator consistency, and information isolation between agent-visible instructions and hidden scoring fields.

To support stratified capability evaluation, queries are organized into four difficulty levels through human-in-the-loop calibration. Difficulty is treated as a property of the required embodied execution process rather than as a simple function of the number of scenes, NPCs, or objects. The calibration mainly considers the spatial exploration scope, procedural length, interaction complexity, evidence requirements, and need for dynamic adaptation. This design allows the same scene to support both simple localized tasks and compound long-horizon tasks while preserving comparable difficulty semantics across heterogeneous environments.

\textbf{Visibility Isolation and Validation.} Each task defines a strict information visibility boundary: the agent receives only a filtered subset of the semantic map and natural language instructions, while internal fields such as NPC positions, evaluation signals, and expected trajectories are isolated, ensuring that the agent must rely on autonomous perception and reasoning to complete the task. All generated tasks undergo automated validation covering waypoint reference validity, information isolation integrity, and evaluation coverage, guaranteeing task consistency and reproducibility.

\subsection{Evaluation Procedure and Metrics}
\label{sec:evaluation}

\subsubsection{Embodied Evaluation Workflow}
\label{sec:sim-pipeline}

Validated queries are executed in UnrealZoo as closed-loop embodied episodes. At the beginning of each episode, the environment is initialized with the task configuration, while the agent receives only the task instruction and the filtered semantic map. Hidden evaluator states, expected trajectories, and success signals are not exposed to the agent.

The UE environment provides NavMesh-based autonomous navigation, which abstracts low-level path planning and movement control. This design allows the LLM-based agent to focus on high-level embodied decision making, including where to navigate, when to observe, what evidence to collect, and when to terminate, rather than on low-level locomotion control. Visual information is obtained through the agent's VLM-based observation tool, which converts first-person views into textual observation evidence for the LLM controller. Higher-difficulty scenarios may additionally include dynamic events such as NPC relocation or object-state changes, requiring the agent to update its plan during execution.

Each episode produces a structured trajectory comprising navigation commands, visited-region traces, observation records, intermediate outputs, final responses, and optional video recordings. The evaluator uses this trace together with task-defined hidden criteria to perform deterministic checks and evidence-grounded judgment, enabling reproducible evaluation and fine-grained failure analysis.

% ── 3.4 Evaluation Metrics ──
\subsubsection{Evaluation Metrics}\label{sec:evaluation_metrics}

A structured evaluator is employed to automatically assess agent execution results, monitoring key events during task execution, including target location arrival status, NPC dialogue completion status, and information acquisition correctness.

The evaluation metrics include Task Success Rate (TSR) and Goal Completion Rate (GCR). TSR requires all subtask objectives and terminal conditions defined by the task to be satisfied for the trajectory to be considered successful, serving as the most stringent overall measure. GCR is defined as the proportion of satisfied subtask objectives to the total number of subtask objectives, providing a finer-grained reflection of the agent's partial completion across dimensions such as navigation, NPC dialogue interaction, and multi-stage reasoning. Together, the two metrics characterize the agent's task execution capability at both the holistic and granular levels.

Subtask objectives are verified through a hybrid scoring scheme.
Geometric conditions, such as target-location arrival and waypoint
visitation, are checked programmatically by comparing recorded poses to
task-defined reference points within a success radius. Semantic
conditions, such as whether the agent has observed or inspected a target,
are evaluated from run-derived evidence, including the agent's
VLM-generated observation records, intermediate outputs, and final
response. An LLM judge is used only to determine whether this evidence
satisfies the semantic success criteria; it does not observe the
environment directly or access hidden evaluator states beyond the
task-level scoring rubric. Each judgment is emitted as a structured
verdict, keeping semantic scoring auditable and consistent with the
deterministic checks.
\section{Model Training}
\label{sec:training}

\subsection{Overview}
\label{sec:training-overview}

\begin{figure}[!t]
\centering
\includegraphics[width=\linewidth]{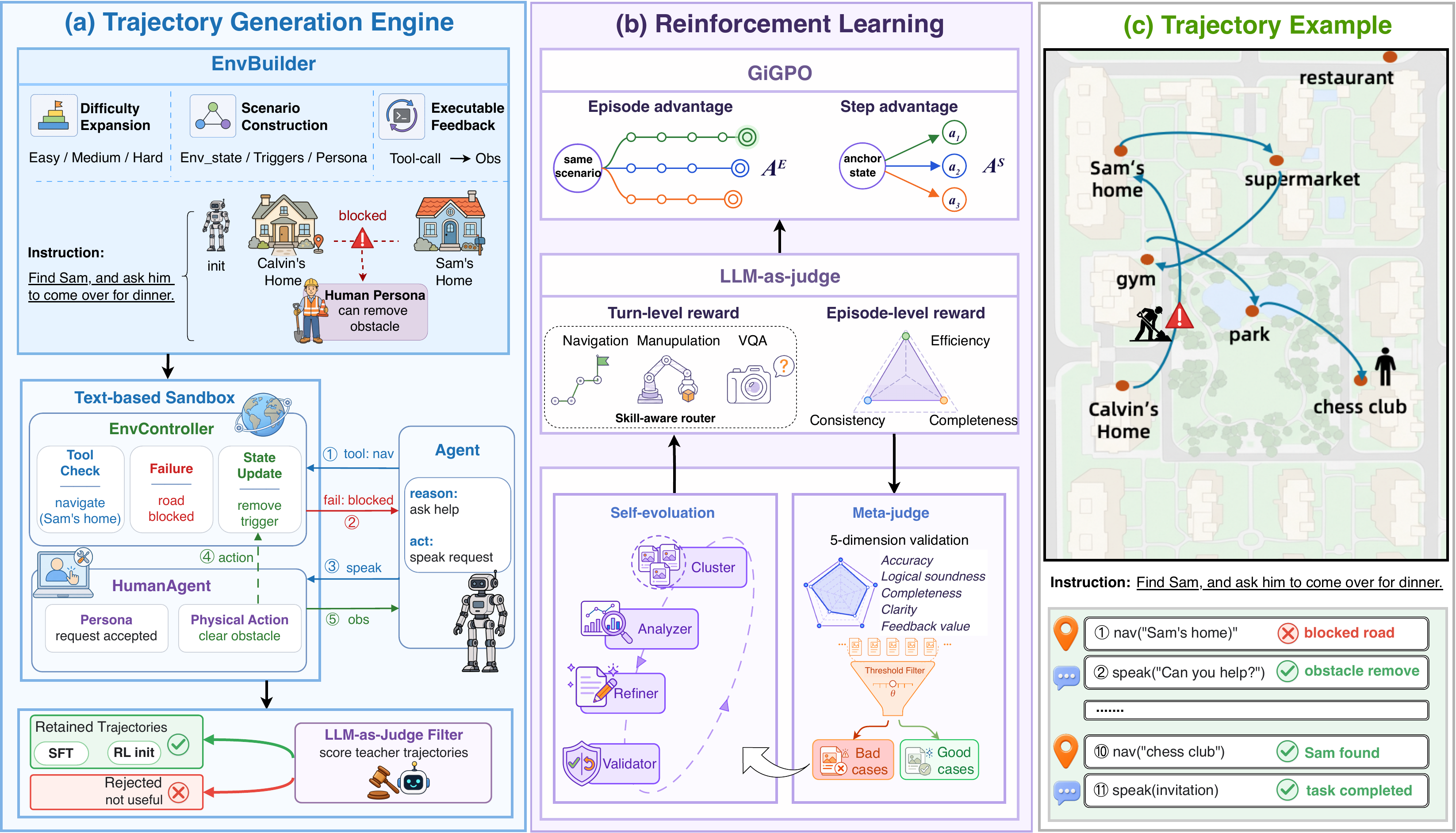}
\vspace{-0.4em}
\caption{Overview of the training pipeline for a deployable ABot-AgentOS student policy. The pipeline constructs controllable text-based environments, distills teacher trajectories for SFT initialization, and improves the policy through online RL with LLM-as-a-Judge rewards and GiGPO advantages.}
\label{fig:trajectory_generation_engine}
\end{figure}

Our goal is to transfer ABot-AgentOS-style long-horizon planning and
tool-use capabilities from large teacher models to a smaller deployable
agent model.
To this end, we design a closed-loop training pipeline that connects
four stages: text-based environment construction, teacher trajectory
distillation, supervised fine-tuning (SFT), and reinforcement learning
(RL). The pipeline first builds controllable semantic sandboxes for
embodied tool use, then uses strong teacher agents to produce
tool-interaction trajectories, and finally trains a smaller student
policy from these trajectories and from online rollouts evaluated by an
LLM-as-a-Judge reward engine, as summarized in Figure~\ref{fig:trajectory_generation_engine}.
This design follows the recent trend that language agents should be
trained in executable or stateful settings rather than only on static
instruction-response pairs. Prior work on ReAct, tool-use data
generation, interactive agent benchmarks, and synthesized
tool-interactive environments shows that realistic agent behavior
requires persistent state, multi-turn feedback, executable actions, and
verifiable outcomes~\cite{yao2022react,qin2023toolllm,zhou2023webarena,jimenez2023swebench,yang2024sweagent,yao2024taubench,song2026envscaler}.
Our pipeline adapts this principle to embodied robot tasks by building a
text-based semantic sandbox around ABot-AgentOS skills, human
assistance, and physical obstacles.

The sandbox is not intended to replace a visual simulator. Instead, it
provides a scalable semantic interface for generating and evaluating
tool-use behavior under controlled state changes. The same trace format
supports both stages of training: teacher rollouts can be filtered and
converted into SFT examples, while student rollouts in the same sandbox
provide trajectory evidence for LLM-as-a-Judge rewards during online RL.

\subsection{Text-Based Environment and Task Construction}
\label{sec:text-env}

As shown in Figure~\ref{fig:trajectory_generation_engine}(a), each natural-language
task instruction is converted into a stateful sandbox instance. The
environment builder prompts a large language model to generate \textsc{Easy}, \textsc{Medium}, and \textsc{Hard} variants, each containing three structured components \texttt{env\_state} ,
which specifies locations, reachability, object locations, and object
states; \texttt{failure\_triggers} , which define physical or semantic obstacles; and \texttt{human\_persona} , which defines the human NPC's role, capabilities,
and possible assistance.

The resulting instance is an interactive task world rather than a static
question. During execution, the agent receives feedback only through
tool calls. The EnvController validates ABot-AgentOS skill calls against
the current state and failure triggers, returning observations or
failures accordingly, while the HumanAgent handles valid help requests
and may perform actions that update \texttt{env\_state}. This design makes
obstacle handling, help seeking, replanning, and state updates part of
the training signal.

% We use difficulty expansion to improve the diversity and curriculum structure of the data. In \textsc{Easy} scenarios, the path is typically unobstructed, so the model mainly learns task decomposition and basic skill selection. In \textsc{Medium} scenarios, one obstacle is introduced, requiring the model to recover from a failed observation through human assistance or replanning. In \textsc{Hard} scenarios, multiple or nested obstacles require multi-turn interaction, persistent goal tracking, and careful recovery from failures. This curriculum exposes the student model to both straightforward execution and complex exception handling while keeping the sandbox controllable enough for scalable data generation and online rollout.

Difficulty expansion provides a simple curriculum. \textsc{Easy}
scenarios emphasize task decomposition and basic skill selection;
\textsc{Medium} scenarios introduce recoverable obstacles; and
\textsc{Hard} scenarios require multi-turn interaction, persistent
goal tracking, and recovery from multiple or nested failures. The
curriculum increases behavioral diversity while keeping the sandbox
controllable for scalable trajectory generation and online rollout.

\subsection{Teacher Trajectory Distillation and SFT}
\label{sec:trajectory-distillation}

After environment construction, a strong teacher model controls an
ABot-AgentOS-style agent in the sandbox, producing ReAct-style
interaction traces that interleave reasoning, tool calls, observations,
human responses when applicable, and final answers~\cite{yao2022react}.
The teacher reuses the ABot-AgentOS system prompt and tool definitions,
while the execution backend is replaced by the text-based sandbox. As a
result, the collected trajectories capture not only successful
executions, but also failed attempts, help-seeking behavior, retries, and
recovery steps, which are essential for training a smaller model to act
robustly under unexpected feedback.

% In implementation, the teacher agent reuses the system prompt and tool definitions of the ABot-AgentOS agent, but the execution backend is replaced by the text-based sandbox described above. Calls to ABot-AgentOS skills are routed to the EnvController, speech actions are routed to the HumanAgent, and human physical actions are fed back to the EnvController to update the world state. Consequently, the collected trajectories include successful executions, failed attempts, help-seeking behavior, retries, and recovery steps. These behaviors are especially important for training smaller models, since robust embodied agents must learn how to react to unexpected observations rather than simply imitate ideal action sequences.

% To ensure data quality, we apply an LLM-as-a-Judge filter to the distilled trajectories. The judge receives the original task, the initial environment, the tool-call sequence, observations, and the final response. It evaluates whether the trajectory completes the task, whether obstacles are handled correctly, whether the agent enters invalid loops, whether tool calls are physically feasible, and whether the behavior respects the robot dog's capability constraints. Only trajectories that are logically consistent and useful for training are retained. The detailed 

To ensure data quality, we apply an LLM-as-a-Judge filter to the
distilled trajectories. Given the task, initial environment, tool-call
sequence, observations, and final response, the judge evaluates task
completion, obstacle handling, loop avoidance, tool feasibility, and
consistency with embodiment-specific capability constraints. Only
logically consistent trajectories are retained. The same judge family is
later used as the reward provider for online RL, with the self-evolving
reward engine described in Section \ref{sec:judge}.

% LLM-as-a-Judge has been widely used for scalable evaluation of open-ended model outputs. G-Eval\cite{liu2023geval} uses an LLM evaluator with explicit rubrics and structured scoring for natural language generation . MT-Bench and Chatbot Arena study the agreement and biases of LLM judges in conversational evaluation \cite{zheng2023judging}. In our setting, the judge is combined with structured environment state: the sandbox produces local observations, while the judge evaluates the global trajectory. This hybrid design keeps the evaluation scalable while reducing the risk that the judge ignores physical feasibility or task-specific constraints.  For detailed implementation of LLM-as-a-Judge, please refer to Sec \ref{sec:judge}.

The retained trajectories are converted into tool-call SFT examples that
preserve the interaction structure instead of flattening each episode
into a single response. This trains the student model as an executable
policy over ABot-AgentOS skills: it learns task decomposition, tool
selection, recovery from failed observations, help seeking, goal
maintenance across multiple turns, and avoidance of unsupported actions.
The resulting SFT model provides the behavioral prior for online RL in
the same sandbox interface.

\subsection{RL: Online Exploration with LLM-as-a-Judge Reward}
\label{sec:rl}

% \subsubsection{Online RL Framework}

% After SFT, the student policy is further trained through online RL in the same text-based sandbox used for trajectory generation.  As shown in Figure~\ref{fig:trajectory_generation_engine}(b), this stage changes the role of the model interacting with the sandbox: instead of a teacher agent producing filtered demonstrations, the SFT-initialized student policy samples actions online, receives EnvController/NPC feedback, and updates from reward-guided policy optimization. This follows recent multi-turn agent RL systems that move beyond static imitation toward interaction-based improvement~\citep{zhou2024archer,wei2025webagentr1,wang2025ragen}.

After SFT, the student policy is further trained through online RL in
the same text-based sandbox used for trajectory generation. Instead of
imitating teacher demonstrations, the SFT-initialized student policy
samples actions online, receives EnvController and NPC feedback, and
updates from reward-guided policy optimization. This follows recent
multi-turn agent RL systems that move beyond static imitation toward
interaction-based improvement~\cite{zhou2024archer,wei2025webagentr1,wang2025ragen}.

% The RL loop has four roles. The policy is the only trainable component. The EnvController executes tool calls against the current sandbox state and returns environment feedback. The NPC module provides human-assistance responses when the policy asks for help. The frozen LLM-as-a-Judge reward provider reads rollout evidence and assigns rewards. This separation keeps online RL focused on improving the small policy, while sandbox dynamics, NPC behavior, and judge rewards remain fixed feedback.

The RL loop has four roles. The policy is the only trainable component.
The EnvController executes tool calls against the current sandbox state
and returns environment feedback. The NPC module provides
human-assistance responses when the policy asks for help. The frozen
LLM-as-a-Judge reward provider reads rollout evidence and assigns
rewards. This separation keeps online RL focused on improving the small
policy, while sandbox dynamics, NPC behavior, and judge rewards remain
fixed feedback. Rewards are supplied by the LLM-as-a-Judge reward engine
described in Section~\ref{sec:judge}, which provides turn-level feedback for local
action quality and episode-level feedback for global task completion.

We use GiGPO~\cite{feng2025gigpo} to convert these rewards into relative
advantages rather than using raw scores directly. For each scenario,
multiple rollouts are sampled from the same initial sandbox state. GiGPO
compares their merged returns to form an episode relative advantage,
which identifies which complete rollout solves the same task better. It
also compares turn-level rewards at aligned or comparable decision steps
to form a step relative advantage, so local recovery choices can be
reinforced when the policy reaches similar situations, such as the same
blocked passage, nearby NPC, or candidate target. The final advantage
combines episode-level and step-level relative comparisons, so policy
updates favor both better complete rollouts and better local actions.

\label{sec:rl}

\iffalse
\subsection{Judge Model}
\label{sec:judge}

The Judge Model serves as the reward oracle during RL training, evaluating both individual actions and overall task completion.

\begin{itemize}
    \item \textbf{Scope:} Turn-level scoring (action quality per step) and episode-level scoring (task completion assessment).
    \item \textbf{Implementation:} Built on \texttt{TurnLevelJudge}, supporting both HTTP service mode (for distributed training) and in-process mode (\texttt{judge\_local}, for debugging).
    \item \textbf{Format enforcement:} Includes configurable format-rule penalties that discourage structurally invalid outputs, complementing the semantic reward signal.
\end{itemize}
\fi

\subsection{Self-Evolving Reward Engine}
\label{sec:judge}
As shown in Figure~\ref{fig:trajectory_generation_engine}(b), this section introduces a self-evolving reward engine for embodied agent evaluation and optimization. It connects three functions in one loop: LLM-as-a-Judge produces rewards for embodied trajectories, Meta-Judge validation checks whether these rewards are reliable, and a multi-agent workflow converts low-quality judge cases into localized prompt updates. The purpose is not only to provide rewards for online RL, but also to keep the reward prompt improving as skills, tasks, and agent behaviors change.

\subsubsection{Reward from LLM-as-a-Judge}
\label{subsec:multi_tier_reward}

The reward engine converts each embodied trajectory into turn-level and episode-level signals. Turn-level rewards provide dense supervision for individual actions, helping the policy distinguish useful, redundant, unsafe, malformed, or context-inconsistent steps. Episode-level rewards evaluate global task completion and capture quality that cannot be judged from isolated turns alone. Together, these signals reduce the credit-assignment difficulty of long-horizon trajectories, where terminal success or failure is too sparse to explain which intermediate action caused the outcome~\citep{wei2025reinforcing}.

At the turn level, reward generation is skill-aware. The engine routes navigation, manipulation, visual question answering, and tool-use actions to different rubrics because their correctness criteria differ. It then combines LLM-based semantic judgment with verifiable rule checks, including invalid tools, illegal commands, malformed arguments, and missing temporal preconditions. We also add action omission penalties for locally plausible actions that skip necessary causal steps, such as reporting a destination before verifying the target or asking for help without first checking the relevant environmental constraint.

At the episode level, the judge summarizes efficiency, consistency, and completeness over the full trajectory~\citep{han2026swetrace}. This episode-level signal is the terminal term used by RL in Section~\ref{sec:rl}. Given an embodied trajectory $\tau$, it is decomposed as:
\begin{equation}
r^{\mathrm{episode}}(\tau)
=
\lambda_{\mathrm{eff}} R_{\mathrm{eff}}(\tau)
+ \lambda_{\mathrm{cons}} R_{\mathrm{cons}}(\tau)
+ \lambda_{\mathrm{comp}} R_{\mathrm{comp}}(\tau),
\label{eq:episode_reward}
\end{equation}
where $R_{\mathrm{eff}}$, $R_{\mathrm{cons}}$, and $R_{\mathrm{comp}}$ denote episode-level efficiency, consistency, and completeness rewards. Combined with the effective turn-level rewards, the merged return is:
\begin{equation}
R(\tau)=
\sum_{t=1}^{T}\hat{r}_{t}
+
r^{\mathrm{episode}}(\tau),
\label{eq:multi_tier_reward}
\end{equation}
where $\hat{r}_{t}$ is the effective turn-level reward produced by skill-specific rubric routing, LLM semantic judgment, verifiable rule checks, and omission penalties. This formulation matches the reward interface used by GiGPO while making explicit how the reward engine produces the terminal episode-level component. It turns LLM-as-a-Judge from a static evaluator into a structured reward generator for embodied agent optimization~\citep{zheng2023judging}.

\subsubsection{Meta-Judge Validation}
\label{subsec:meta_judge_validation}

The second component validates the reliability of the reward signal. LLM-as-a-Judge outputs can contain scoring mistakes, missed evidence, rubric misapplication, or plausible but unsupported rationales~\citep{silva2026metajudging}. These failures are especially risky in embodied settings, where correct judgment depends on physical preconditions, tool constraints, temporal order, capability boundaries, and task-specific preferences.

Meta-Judge does not directly decide whether the embodied agent acted correctly. Instead, it receives the first-order judge output, the highlighted turn, and the full trajectory context as validation evidence, then checks whether the judge's score and rationale are reliable under the relevant rubric. Its quality dimensions are accuracy, logical soundness, completeness, clarity, and feedback value. Accuracy checks whether the score is supported by trajectory evidence; logical soundness checks whether the explanation connects evidence, rubric, and score without unsupported jumps; completeness checks whether key embodied constraints and omitted or redundant actions are covered; clarity checks whether evidence and rule use are explicit; and feedback value checks whether the rationale is localized enough for prompt refinement.

These dimensions are aggregated into an overall quality score:
\begin{equation}
Q(x_t)=\sum_{k=1}^{5} w_k q_k,
\label{eq:meta_quality}
\end{equation}
where $x_t$ denotes the validated judge output, $q_k$ is the score of the $k$-th validation dimension, and $w_k$ is its weight. A sample is marked as a low-quality judge case when:
\begin{equation}
\mathrm{LowQualityJudgeCase}(x_t)=\mathbb{I}[Q(x_t)<\theta].
\label{eq:low_quality_case}
\end{equation}
These cases are more useful than binary error labels because they preserve diagnostic feedback. They indicate whether the reward prompt is underspecified, ambiguous, or misaligned with embodied execution constraints, and they provide the input signal for the next refinement stage~\citep{wang2026outcome,li2025multiagent}.

\subsubsection{Multi-Agent Self-Evolution}
\label{subsec:structured_reward_optimization}

The third component converts low-quality judge cases into localized reward-prompt updates. Generic prompt optimization methods often optimize a monolithic prompt with task-level feedback~\citep{yang2024opro,khattab2023dspy}. In our setting, the reward prompt contains skill-specific rubrics, temporal precondition rules, reasoning examples, and output constraints. A global rewrite may repair one failure while damaging another stable skill, so we treat the prompt as editable structured components, following the broader idea of modular textual parameters and localized textual modification from execution feedback~\citep{he2026learning}.

The self-evolution loop is implemented as a multi-agent workflow. The Cluster stage groups low-quality judge cases by skill and failure type. The Analyzer identifies the defective rubric component or example pattern and produces a revision plan. The Refiner applies localized edits to the relevant rubric, rule, or example. The Validator then evaluates the revised prompt on the full validation set and accepts the update only when it improves overall quality.

If a candidate update hurts full-set performance, per-skill behavior, or Meta-Judge quality, the system rolls back to the previous prompt snapshot. Under the current validation setting, the initial Judge Model achieves roughly 60\% human alignment, while the Meta-Judge-driven self-evolution process improves alignment to above 90\%. This gain reflects the coupled effect of the full loop: LLM-as-a-Judge produces decomposed rewards, Meta-Judge turns unreliable judgments into diagnostic cases, and the multi-agent optimizer converts these cases into validated prompt updates. Overall, the reward engine closes the loop around reward production, validation, and refinement so that the evaluation system can remain reliable as embodied skills, task distributions, and agent behaviors evolve.

% \input{sections/5router}
% =========================================================
% Required packages:
% \usepackage{booktabs}
% \usepackage{multirow}
% \usepackage{tabularx}
% \usepackage{threeparttable}
% =========================================================

\section{Experiments}
\label{sec:experiments}

\subsection{Agent Evaluation}
\label{subsec:exp-agent}

\subsubsection{Experimental Setup}
\label{subsubsec:agent-setup}

We evaluate the proposed embodied agent on a subset of the current benchmark. The subset covers the main scene types considered by the benchmark, including indoor, outdoor, and indoor-outdoor environments, and includes several task forms such as semantic navigation, regional person or object search, object inspection, NPC information query, status reporting, and multi-stage instruction following. The results in this section are intended as an initial system validation rather than a complete benchmark leaderboard.

The agent operates in UnrealZoo through the recorder interface, which executes NavMesh-based navigation commands and records the resulting trajectory and interaction trace. Visual observations are obtained through the agent's VLM-based observation tool rather than exposed as privileged environment state. During evaluation, the agent must act from the task instruction, filtered semantic map, observable feedback, and dialogue events, without access to hidden evaluator states or success signals.

% The agent controls an ego agent through the recorder API for movement, observation, and interaction. During evaluation, the agent must make decisions from the task instruction, observable environment information, and interaction feedback, without access to hidden evaluator states or success signals. The experiment therefore focuses on the agent's ability to understand long-horizon instructions, use scene information, gather evidence, and complete interactions.

We compare three settings: Qwen3.6-Plus, Qwen3.6-Plus (Ours), and DeepSeek-V4-Pro (Ours). Qwen3.6-Plus is the baseline with a single LLM controller, while settings marked with (Ours) use the proposed hierarchical agent architecture, including local skill execution, task memory, and final verification. This allows us to compare the effect of the agent architecture under the same model, as well as the effect of changing the main LLM within our architecture. Across all agent-evaluation settings, visual observation is handled by the same Qwen3-VL-Plus observation tool. The compared models refer to the main LLM controller, which performs task reasoning, planning, tool selection, and termination decisions. 

We use the metrics defined in the benchmark section and mainly report \textbf{TSR} and \textbf{GCR}. TSR
measures complete task success, while GCR measures goal-level completion,  under the benchmark's structured success
criteria and terminal conditions. Since this section
reports results on a small subset, we only discuss the main trends and
avoid fine-grained claims over models or task categories.

\subsubsection{Experimental Results}
\label{subsubsec:agent-main-results}

Table~\ref{tab:stage1-main-results} summarizes the current experimental
results. Under the same Qwen3.6-Plus model, Ours improves over the
baseline on both TSR and GCR. Replacing the main model of Ours with
DeepSeek-V4-Pro further improves both metrics.

\begin{table}[t]
    \centering
    \caption{Agent evaluation results on the EmbodiedWorldBench subset.}
    \label{tab:stage1-main-results}
    \begin{tabular}{llcc}
        \hline
        Agent & Model & TSR & GCR \\
        \hline
        ReAct & Qwen3.6-Plus & 49.97\% & 57.95\% \\
        ABot-AgentOS & Qwen3.6-Plus& 61.96\% & 68.79\% \\
        ABot-AgentOS & DeepSeek-V4-Pro & 68.18\% & 74.62\% \\
        \hline
    \end{tabular}
\end{table}

With the same Qwen3.6-Plus model, Ours improves TSR by 11.99\% and GCR by 10.84\% over the baseline. This suggests that hierarchical execution, task memory, skill-level feedback, and finish-time verification help long-horizon embodied execution under the same base model and tool interface. DeepSeek-V4-Pro (Ours) further improves over Qwen3.6-Plus (Ours) by 6.22\% in TSR and 5.83\% in GCR, suggesting that a stronger main model may improve task interpretation, local recovery, and termination decisions.

\subsubsection{Analysis}
\label{subsubsec:agent-analysis}

Overall, the results support a preliminary conclusion: compared with the
single-controller baseline, Ours is better suited for embodied tasks
that require multi-stage progress, local search, interaction evidence,
and verified termination. The simultaneous improvement in GCR and TSR
suggests that the gain is not limited to partial progress; the
hierarchical structure helps convert more trajectories into complete
task success.

The main architectural benefit appears to come from maintaining
structured task state while delegating sustained local execution to
skill subagents and checking progress through verification. This reduces
common long-horizon failure modes of the single-controller baseline,
including context drift, premature termination, weak evidence gathering,
and local search loops.

The difference between main LLMs reflects the impact of the backbone
model on long-horizon execution quality. DeepSeek-V4-Pro achieves higher
results on this subset, suggesting that stronger instruction
understanding, stage planning, and failure recovery can further improve
the same agent architecture.

Remaining failures are mainly related to perception feedback and scene
understanding. First, the agent still lacks sufficiently fine-grained
active observation and feedback mechanisms, which makes it difficult to
reliably confirm target states or correct subsequent actions after
approaching a plausible location. Second, the VLM observation tool can confuse
people and objects, which affects NPC localization, object search, and
interaction decisions. Third, in scenes with indoor-outdoor connections
or large visible openings, the agent may confuse visible regions with
its own location, for example interpreting an outdoor area seen from
indoors as evidence that it has already moved outdoors. In addition,
uncertainty in real UE/recorder execution makes evaluation more
difficult. Future work should improve fine-grained observation policies,
visual evidence verification, and spatial reasoning over scene regions.

\providecommand{\NA}{--}
\providecommand{\memgroup}[2]{%
\midrule
\multicolumn{#1}{c}{\textit{#2}}\\
\midrule
}

\subsection{Memory Evaluation}
\label{subsec:exp-memory}

This section evaluates the proposed memory module independently from the agent-control component.
All memory experiments use the same base hybrid graph retriever described in Section~\ref{sec:memory_retrieval}; therefore, retrieval itself is not treated as an experimental variable.
The evaluation is organized around five benchmarks that stress complementary memory capabilities: long-term conversational recall, embodied question answering, multi-modal identity and relation grounding, temporal video reasoning, and egocentric daily-life memory.
Specifically, we evaluate on LoCoMo, OpenEQA, Mem-Gallery, NExT-QA, and EgoLife \cite{locomo,openeqa,memgallery,nextqa,egolife}.
These benchmarks differ substantially in modality, temporal scale, supervision format, and evaluation metric, so we report dataset-specific results rather than averaging raw scores across benchmarks.

Tables~\ref{tab:mem-locomo}--\ref{tab:mem-egolife} report the static ABot-AgentOS memory system together with external memory, VQA, and upper-bound baselines.
The lifelong self-evolution results are reported separately in Section~\ref{subsubsec:memory-selfevo}.
Across all ABot-AgentOS variants in this section, the memory schema and base hybrid graph-retrieval backbone are fixed.
Self-evolution may only add constrained runtime policies, such as query-rewriting, temporal-normalization, evidence-selection, answerer-calibration, or frame-selection policies; it does not change the underlying retrieval implementation.

\begin{table*}[!t]
\centering
\caption[LoCoMo memory results.]{
LoCoMo memory results by question type.
Single-hop, Temporal, Multi-hop, Open-domain, and Adversarial denote the LoCoMo question categories.
All scores follow the Mem0 judge-prompt protocol.
ABot-AgentOS Static uses Qwen3.6-Plus as memory writer and answerer.
}
\label{tab:mem-locomo}
\small
\setlength{\tabcolsep}{4.0pt}
\renewcommand{\arraystretch}{0.94}
\begin{tabular}{llcccccc}
\toprule
\multirow{2}{*}{Method}
& \multirow{2}{*}{Setting}
& \multicolumn{6}{c}{LoCoMo} \\
\cmidrule(lr){3-8}
& & Single-hop & Temporal & Multi-hop & Open-domain & Adversarial & Overall \\
\memgroup{8}{QA and full-context references}
SimpleSearch~\cite{wu2025longmemeval}
& Retrieval
& 76.6 & 72.9 & 71.9 & 83.9 & 72.6 & 77.9 \\
Qwen3.6-Plus
& Full context
& 93.9 & 70.4 & 90.1 & 59.4 & 70.6 & 82.7 \\
GPT-5.4
& Full context
& 93.1 & 65.1 & 90.1 & 68.7 & 81.6 & 84.4 \\
\memgroup{8}{Memory-based methods}
A-MEM~\cite{xu2025amem}
& Memory
& 35.9 & 31.8 & 23.1 & 23.9 & 7.4 & 26.4 \\
MIRIX~\cite{wang2025mirix}
& Memory
& 82.6 & 81.3 & 82.9 & 64.6 & 36.3 & 71.2 \\
MemGPT~\cite{packer2023memgpt}
& Memory
& 86.3 & 85.6 & 86.8 & 57.3 & 65.9 & 80.3 \\
MemInsight~\cite{salama2025meminsight}
& Memory
& 82.1 & 76.9 & 80.5 & 64.6 & \textbf{90.3} & 82.0 \\
Mem0~\cite{mem0}
& Memory
& \textbf{94.1} & \textbf{96.6} & 90.8 & 68.8 & 62.3 & 85.6 \\
ABot-AgentOS Static
& Graph memory
& 92.9 & 87.5 & \textbf{90.9} & \textbf{70.8} & 80.9 & \textbf{87.5} \\
\memgroup{8}{Human benchmark}
Human~\cite{locomo}
& Human
& 95.1 & 92.6 & 85.8 & 75.4 & 89.4 & 87.9 \\
\bottomrule
\end{tabular}
\end{table*}

\begin{table}[h]
\centering
\caption[OpenEQA EM-EQA results.]{
OpenEQA EM-EQA results by data source.
ScanNet, HM3D, and Overall report LLM-Match scores.
ABot-AgentOS uses GPT-5.4 as memory writer, answerer, and judge.
For GaussExplorer, the source paper does not report a fixed reconstruction-image budget.
}
\label{tab:mem-openeqa}
\small
\setlength{\tabcolsep}{3.8pt}
\renewcommand{\arraystretch}{0.94}
\begin{tabular}{llcccc}
\toprule
\multirow{2}{*}{Method}
& \multirow{2}{*}{Memory / Input}
& \multirow{2}{*}{Frames}
& \multicolumn{3}{c}{EM-EQA} \\
\cmidrule(lr){4-6}
& & & ScanNet & HM3D & Overall \\
\memgroup{6}{Direct VQA references}
Qwen3.6-Plus
& Direct VQA
& 8
& 72.6 & 55.3 & 66.7 \\
GPT-5.4
& Direct VQA
& 8
& 72.7 & 58.4 & 67.8 \\
Qwen3.6-Plus
& Direct VQA
& 24
& 77.0 & 65.2 & 73.0 \\
GPT-5.4
& Direct VQA
& 24
& 75.9 & 70.5 & 74.1 \\
\memgroup{6}{Memory-based methods}
GPT-4 + ConceptGraphs~\cite{openeqa}
& Scene graph
& 10
& 37.8 & 34.0 & 36.5 \\
GPT-4 + LLaVA-1.5~\cite{openeqa}
& Caption memory
& 10
& 45.4 & 40.0 & 43.6 \\
R-EQA~\cite{ong2025reqa}
& Retrieved captions
& 3
& 49.1 & 42.8 & 46.0 \\
GraphPad~\cite{ali2025graphpad}
& Dynamic scene memory
& 5--20
& \NA & \NA & 55.3 \\
3D-Mem / SnapMem~\cite{yang2025threedmem}
& 3D snapshot memory
& 3.1
& \NA & \NA & 57.2 \\
GaussExplorer~\cite{kim2026gaussexplorer}
& 3DGS memory
& n/a
& \NA & \NA & 57.8 \\
ABot-AgentOS Static
& Graph memory
& 8
& \textbf{62.8} & 52.3 & 59.2 \\
ABot-AgentOS Static
& Graph memory
& 24
& 61.9 & \textbf{55.7} & \textbf{59.9} \\
\memgroup{6}{Human upper bound}
Human~\cite{openeqa}
& Human
& \NA
& 87.7 & 85.1 & 86.8 \\
\bottomrule
\end{tabular}
\end{table}

\subsubsection{Experimental Setup}
\label{subsubsec:memory-setup}

\paragraph{Datasets}
We evaluate the memory module on five datasets that cover complementary aspects of long-term and embodied memory.
LoCoMo evaluates very long-term multi-session conversational memory.
OpenEQA evaluates open-vocabulary embodied question answering over indoor environments.
Mem-Gallery evaluates multi-modal long-term conversational memory with visual-textual dependencies.
NExT-QA evaluates temporal, causal, and descriptive video question answering.
EgoLife evaluates long-context egocentric daily-life question answering.
Because these benchmarks differ in modality, context length, question type, and answer format, we report results separately and avoid aggregating them into a single cross-dataset score.

\paragraph{Base Hybrid Retriever}
All experiments use the same hybrid graph retriever.
Given a query, the retriever first selects candidate memory nodes using semantic embeddings, lexical matching, metadata filters, source constraints, and node-type constraints.
It then expands the selected seed nodes through typed graph edges to collect supporting events, frames, sessions, places, participants, identity timelines, provenance records, and spatial or interaction relations.
The retrieved evidence is serialized as a compact evidence context and provided to the answerer.
This fixed retrieval backbone allows us to isolate the effect of structured memory records and, later, the effect of lifelong self-evolution assets.

\paragraph{Writer and Answerer Models}
In ABot-AgentOS, the memory writer converts raw observations, dialogue turns, image attachments, video frames, timestamps, and interaction traces into structured, source-grounded memory records.
The answerer then produces the final response from the user query and the retrieved evidence context.
We instantiate ABot-AgentOS with benchmark-specific base models according to each dataset's modality and evaluation protocol.
For OpenEQA, GPT-5.4 is used as the memory writer, answerer, and LLM-Match judge.
For LoCoMo, Qwen3.6-Plus is used as both the memory writer and answerer.
For Mem-Gallery, Qwen3.6-Plus is used as the memory writer, answerer, and LLM judge.
For NExT-QA, Qwen3.6-Plus is used to write structured video memories.
For EgoLife, Qwen3.5-Flash is used to write egocentric memories.

\paragraph{Comparison Methods}
For memory-oriented baselines, we compare against representative methods that cover scalable conversational memory, multi-modal long-term memory, dynamic video memory, and embodied scene memory.
For direct VQA baselines, we include blind LLM-only or direct visual/video QA results when the dataset supports them.
External results are used as reference points because protocols can differ in input representation, visual backbone, frame budget, model scale, evaluation subset, and judge model.
The most controlled comparison is therefore between reproduced ABot-AgentOS Static and ABot-AgentOS + Self-evo runs under the same split, metric, and implementation.

\paragraph{Information Boundary}
All non-oracle memory graphs are constructed only from information available to the agent at inference time: dialogue turns, observations, frames, timestamps, model-extracted captions, detected entities, and source metadata.
Gold answers, gold rationales, gold supporting evidence, and benchmark-internal annotations are never used to construct the standard memory graph and are never placed in the answerer's inference context.
When supervision is used for self-evolution, it is used only after the corresponding evaluation split has already been evaluated, and the resulting evo-assets can only affect later splits.

\subsubsection{Experimental Results}
\label{subsubsec:memory-results}

\paragraph{LoCoMo Results}
Table~\ref{tab:mem-locomo} reports results on LoCoMo.
Since LoCoMo is a long-term conversational memory benchmark, direct VQA baselines are not applicable.
For a consistent comparison, we reproduce most baseline and all memory results under our evaluation pipeline, except for retrieval method, and evaluate all methods using the Mem0 judge-prompt protocol.
ABot-AgentOS Static obtains $87.5$ overall, outperforming the strongest reproduced memory baseline Mem0 by $1.9$ points and approaching the human overall score of $87.9$.
This result suggests that graph memory is especially effective when the benchmark requires combining long-horizon conversational facts with relation- and provenance-aware retrieval.

\paragraph{OpenEQA Results}
Table~\ref{tab:mem-openeqa} reports results on OpenEQA EM-EQA.
The static ABot-AgentOS memory system achieves up to $59.9$ with 24 frames, outperforming the listed scene-graph, caption-memory, retrieved-caption, 3D snapshot, and 3DGS memory baselines under the reported reference numbers.
On ScanNet, it reaches $62.8$, exceeding the listed memory baselines and showing that explicit memory retrieval can compensate for limited visual context in embodied QA.
The direct VQA rows are included as non-memory upper-bound references rather than as memory systems.

\paragraph{Mem-Gallery Results}
Table~\ref{tab:mem-memgallery} reports results on Mem-Gallery.
ABot-AgentOS Static obtains $88.6$ overall, which is higher than the listed textual and multi-modal memory baselines while remaining below the full-context upper-bound rows.
The largest advantages of the static graph memory appear in categories that require visual-centric reasoning, conflict detection, and answer refusal, where source-grounded graph records help the answerer preserve provenance and avoid unsupported recall.

\begin{table*}[!t]
\centering
\caption[Mem-Gallery LLM-Judge results]{
Mem-Gallery LLM-Judge results by question type.
FR, VS, TTL, TR, VR, MR, KR, CD, and AR denote factual retrieval, visual-centric search, test-time learning, temporal reasoning, visual-centric reasoning, multi-entity reasoning, knowledge resolution, conflict detection, and answer refusal.
ABot-AgentOS uses Qwen3.6-Plus as memory writer, answerer, and judge.
}
\label{tab:mem-memgallery}
\scriptsize
\setlength{\tabcolsep}{2.5pt}
\renewcommand{\arraystretch}{1.08}
\begin{adjustbox}{max width=\textwidth}
\begin{tabular}{llcccccccccc}
\toprule
\multirow{2}{*}{Method}
& \multirow{2}{*}{Memory / Input}
& \multicolumn{10}{c}{Mem-Gallery} \\
\cmidrule(lr){3-12}
& & FR & VS & TTL & TR & VR & MR & KR & CD & AR & Overall \\
\memgroup{12}{QA and full-context references}
GPT-5.4
& Raw RAG
& 91.7 & 79.2 & 93.3 & 87.8 & 83.6 & 92.9 & 85.2 & 91.3 & 100.0 & 89.4 \\
Qwen3.6-Plus
& Raw RAG
& 90.6 & 80.9 & 93.2 & 91.0 & 83.3 & 93.5 & 86.4 & 100.0 & 100.0 & 90.2 \\
GPT-5.4
& Full context
& 98.2 & 78.4 & 96.1 & 92.7 & 91.7 & 97.1 & 90.7 & 92.6 & 100.0 & 92.6 \\
Qwen3.6-Plus
& Full context
& 98.4 & 80.4 & 95.6 & 90.6 & 94.8 & 93.2 & 86.4 & 97.5 & 100.0 & 92.6 \\
\memgroup{12}{Textual memory}
A-MEM~\cite{xu2025amem}
& Captioned memory
& 84.0 & 84.2 & 90.1 & 79.7 & 56.0 & 85.2 & 75.9 & 51.2 & 92.1 & 81.2 \\
MemoryOS~\cite{kang2025memoryos}
& Captioned memory
& 85.6 & 83.5 & 90.2 & 73.6 & 58.1 & 85.7 & 78.4 & 60.5 & 92.1 & 81.7 \\
MemGPT~\cite{packer2023memgpt}
& Captioned memory
& \textbf{94.8} & \textbf{92.7} & 90.2 & \textbf{89.0} & 78.5 & \textbf{92.7} & \textbf{80.3} & 55.6 & 84.8 & 87.6 \\
\memgroup{12}{Multi-modal memory}
NGM~\cite{fisher2025ngm}
& Graph memory
& 80.8 & 86.1 & 91.8 & 69.9 & 59.2 & 80.6 & 59.9 & 63.0 & 92.1 & 80.3 \\
AUGUSTUS~\cite{jain2025augustus}
& Multi-modal memory
& 79.2 & 87.8 & 91.4 & 78.9 & 57.2 & 78.6 & 66.7 & 56.8 & 92.4 & 80.6 \\
MuRAG~\cite{chen2022murag}
& Multi-modal RAG
& 89.3 & 90.5 & \textbf{92.6} & 78.9 & 62.6 & 87.1 & 77.2 & 53.1 & 91.3 & 84.4 \\
UniversalRAG~\cite{yeo2025universalrag}
& Multi-modal RAG
& 90.6 & 92.0 & 89.8 & 80.9 & 67.8 & 87.4 & 72.2 & 54.3 & 90.8 & 84.7 \\
ABot-AgentOS Static
& Graph memory
& 89.7 & 89.7 & 84.2 & \textbf{89.0} & \textbf{81.0} & 90.0 & 76.5 & \textbf{97.5} & \textbf{100.0} & \textbf{88.6} \\
\bottomrule
\end{tabular}
\end{adjustbox}
\end{table*}

\begin{table}[!htbp]
\centering
\caption[NExT-QA validation accuracy.]{
NExT-QA validation accuracy.
Acc@C, Acc@T, Acc@D, and Acc@All denote causal, temporal, descriptive, and overall accuracy.
}
\label{tab:mem-nextqa}
\small
\setlength{\tabcolsep}{3.8pt}
\renewcommand{\arraystretch}{1}
\begin{tabular}{llcccc}
\toprule
\multirow{2}{*}{Method}
& \multirow{2}{*}{Paradigm}
& \multicolumn{4}{c}{Validation} \\
\cmidrule(lr){3-6}
& & Acc@C & Acc@T & Acc@D & Acc@All \\
\memgroup{6}{Supervised VQA}
VFC~\cite{yang2021justask}
& Supervised
& 49.6 & 51.5 & 63.2 & 52.3 \\
ATP~\cite{buch2022revisiting}
& Supervised
& 53.1 & 50.2 & 66.8 & 54.3 \\
MIST~\cite{MIST}
& Supervised
& 54.6 & 56.6 & 66.9 & 57.2 \\
GF~\cite{bai2023glance}
& Supervised
& 56.9 & 57.1 & 70.5 & 58.8 \\
CoVGT~\cite{covgt}
& Supervised
& 59.7 & 58.0 & 69.9 & 60.7 \\
SeViT~\cite{kim2023semiparametricvideogroundedtextgeneration}
& Supervised
& 54.0 & 54.1 & 71.3 & 56.7 \\
HiTeA~\cite{HiTeA}
& Supervised
& 62.4 & 58.3 & 75.6 & 63.1 \\
\memgroup{6}{Zero-shot VQA}
VFC~\cite{VFC}
& Zero-shot
& 51.6 & 45.4 & 64.1 & 51.5 \\
InternVideo~\cite{wang2022internvideogeneralvideofoundation}
& Zero-shot
& 43.4 & 48.0 & 65.1 & 49.1 \\
AssistGPT~\cite{gao2023assistgptgeneralmultimodalassistant}
& Zero-shot
& 60.0 & 51.4 & 67.3 & 58.4 \\
ViperGPT~\cite{ViperGPT}
& Zero-shot
& \NA & \NA & \NA & 60.0 \\
SeViLA~\cite{kim2023semiparametricvideogroundedtextgeneration}
& Zero-shot
& 61.3 & 61.5 & 75.6 & 63.6 \\
LLoVi~\cite{zhang2023simple}
& Zero-shot
& 69.5 & 61.0 & 75.6 & 67.7 \\
GPT-5.4
& Direct QA
& 53.7 & 52.3 & 42.1 & 51.5 \\
Qwen3.6-Plus
& Direct QA
& 83.2 & 90.5 & 77.2 & 81.9 \\
\memgroup{6}{Memory-based methods}
VideoAgent~\cite{fan2025videoagent}
& Video memory
& 72.7 & 64.5 & 81.1 & 71.3 \\
GraphVideoAgent~\cite{GraphVideoAgent}
& Graph video memory
& 74.6 & 65.2 & \textbf{83.5} & 73.3 \\
ABot-AgentOS Static
& Graph memory
& \textbf{77.9} & \textbf{73.4} & 83.4 & \textbf{76.5} \\
\bottomrule
\end{tabular}
\end{table}

\begin{table*}[!htbp]
\centering
\caption[EgoLifeQA multiple-choice accuracy.]{
Multiple-choice question answering accuracy (\%) on EgoLifeQA.
EL, ER, HI, RM, and TM denote EntityLog, EventRecall, HabitInsight, RelationMap, and TaskMaster.
Under Frames, ``1FPS$\rightarrow$X'' means frames are sampled at 1 FPS and $X$ frames are retrieved for reasoning; ``Full'' denotes access to all frames.
}
\label{tab:mem-egolife}
\small
\setlength{\tabcolsep}{3.3pt}
\renewcommand{\arraystretch}{1}
\begin{tabular}{lllcccccc}
\toprule
\multirow{2}{*}{Method}
& \multirow{2}{*}{Frames}
& \multirow{2}{*}{Modality}
& \multicolumn{6}{c}{Accuracy (\%)} \\
\cmidrule(lr){4-9}
& & & EL & ER & HI & RM & TM & Avg. \\
\memgroup{9}{Conventional VQA}
GPT-4.1~\cite{gpt4_1}
& 1FPS
& T
& 32.0 & 39.7 & 39.3 & 32.8 & 39.7 & 36.0 \\
GPT-5.4 mini~\cite{gpt5_4mini}
& 64
& V+T
& 37.6 & 40.5 & 44.3 & 34.4 & 41.3 & 38.8 \\
Gemini 2.5 Pro~\cite{gemini2_5pro}
& 3000
& V+T
& 42.4 & 39.7 & 45.9 & 39.2 & 46.0 & 41.8 \\
Qwen3.5-Flash~\cite{team2026qwen3}
& 64
& V+T
& 42.4 & 39.7 & 45.9 & 39.2 & 46.0 & 41.8 \\
\memgroup{9}{Agentic memory-based methods}
VideoAgent~\cite{fan2025videoagent}
& 1FPS$\rightarrow$8
& V
& \NA & \NA & \NA & \NA & \NA & 29.2 \\
EgoButler-GPT-4o~\cite{egolife}
& \NA
& T
& 34.4 & 42.1 & 29.5 & 30.4 & 44.4 & 36.2 \\
EgoButler-EgoGPT~\cite{egolife}
& 1FPS
& V+T
& 39.2 & 36.5 & 27.9 & 29.6 & 36.5 & 36.0 \\
EGAgent-Gemini2.5 Pro~\cite{rege2026EGAgent}
& 1FPS$\rightarrow$50
& V+T
& 54.4 & 57.1 & 60.3 & 62.4 & \textbf{74.6} & 57.5 \\
WorldMM-Qwen3.5 Flash~\cite{worldmm}
& Full
& V+T
& 49.6 & 54.8 & 54.1 & 62.4 & 60.3 & 56.0 \\
ABot-AgentOS-Qwen3.5 Flash
& 1FPS$\rightarrow$1
& V+T
& \textbf{63.2} & \textbf{60.3} & \textbf{70.5} & \textbf{66.4} & 66.7 & \textbf{65.4} \\
\bottomrule
\end{tabular}
\end{table*}

\paragraph{NExT-QA Results}
Table~\ref{tab:mem-nextqa} reports results on the NExT-QA validation set.
ABot-AgentOS Static reaches $76.5$ Acc@All, improving over the previous memory-based GraphVideoAgent baseline by $3.2$ points.
The gains are strongest on causal and temporal questions, where the memory graph preserves event order, entity relations, and interactions across video segments.
The Qwen3.6-Plus direct zero-shot row is included as a strong non-memory model reference; it is not a memory-system baseline.

\paragraph{EgoLife Results}
Table~\ref{tab:mem-egolife} reports results on EgoLifeQA.
ABot-AgentOS-Qwen3.5-Flash reaches $65.4\%$ average accuracy while retrieving only one relevant frame from 1FPS-sampled video.
It outperforms the listed conventional VQA and agentic memory-based methods on the average score, and obtains the best scores on EntityLog, EventRecall, HabitInsight, and RelationMap.
The only category where another method is higher is TaskMaster, where EGAgent-Gemini2.5 Pro obtains $74.6\%$ and ABot-AgentOS obtains $66.7\%$.

\FloatBarrier

\subsubsection{Lifelong Self-Evolution Results}
\label{subsubsec:memory-selfevo}

\begin{figure}[!htbp]
    \centering
    \includegraphics[width=1\linewidth]{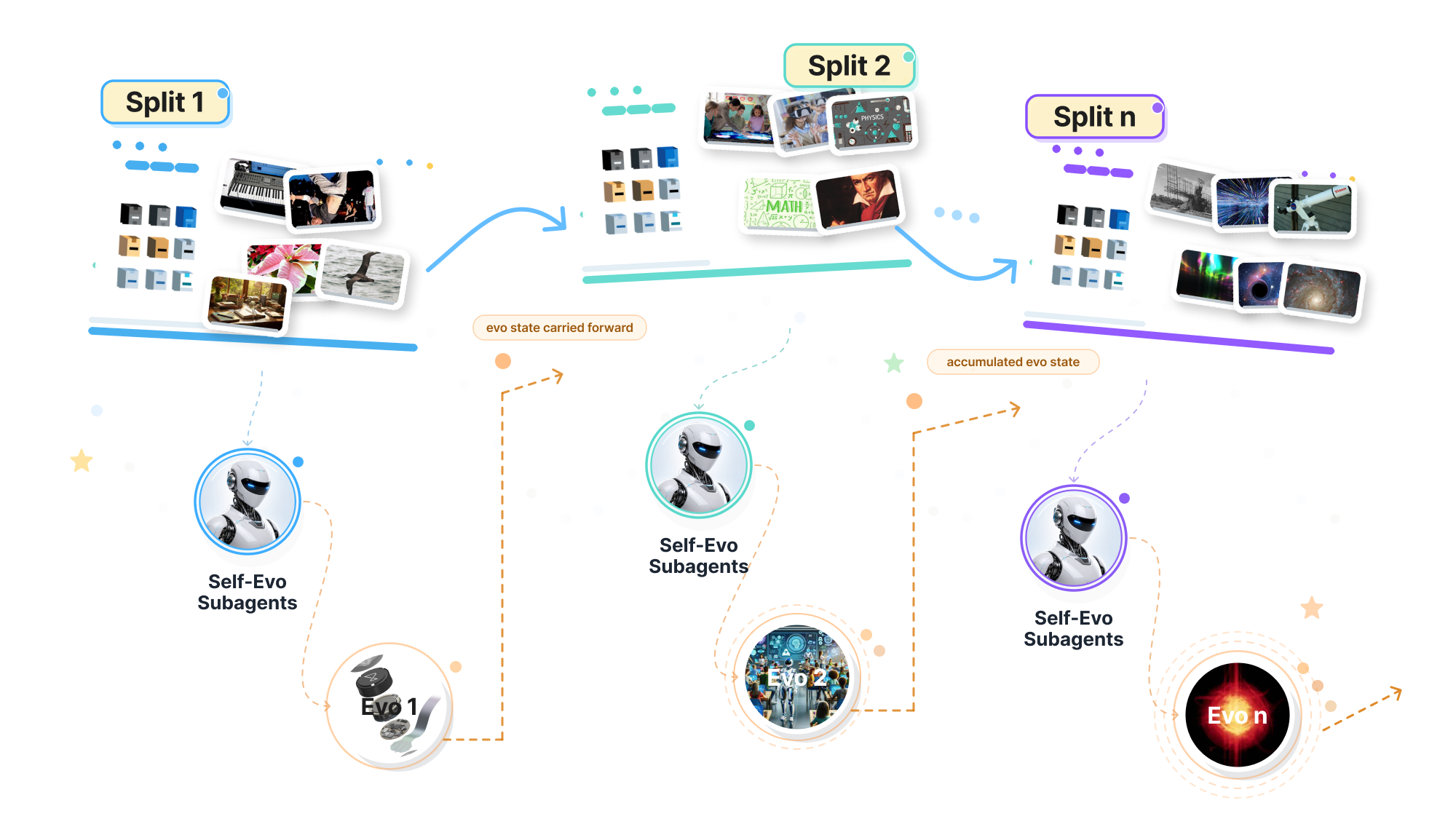}
    \caption{
    Lifelong memory self-evolution across sequential splits.
    Each split uses only evo-assets promoted from previous splits; failures from the current split are diagnosed and gated after evaluation, and accepted assets are used only by later splits.
    }
    \label{fig:lifelong_selfevo_protocol}
\end{figure}

This subsection focuses on the effect of failure-driven lifelong self-evolution, complementing the dataset-specific static memory results.
The self-evolving variant uses the same memory graph schema, the same base hybrid graph retriever, and the same base writer--answerer configuration as the corresponding static run.
The only difference is that later splits can load evo-assets promoted from earlier splits, such as writer rules, evidence-selection preferences, answerer calibration rules, frame-selection policies, or temporal-normalization policies.

\begin{figure*}[!t]
    \centering
    \includegraphics[width=\textwidth]{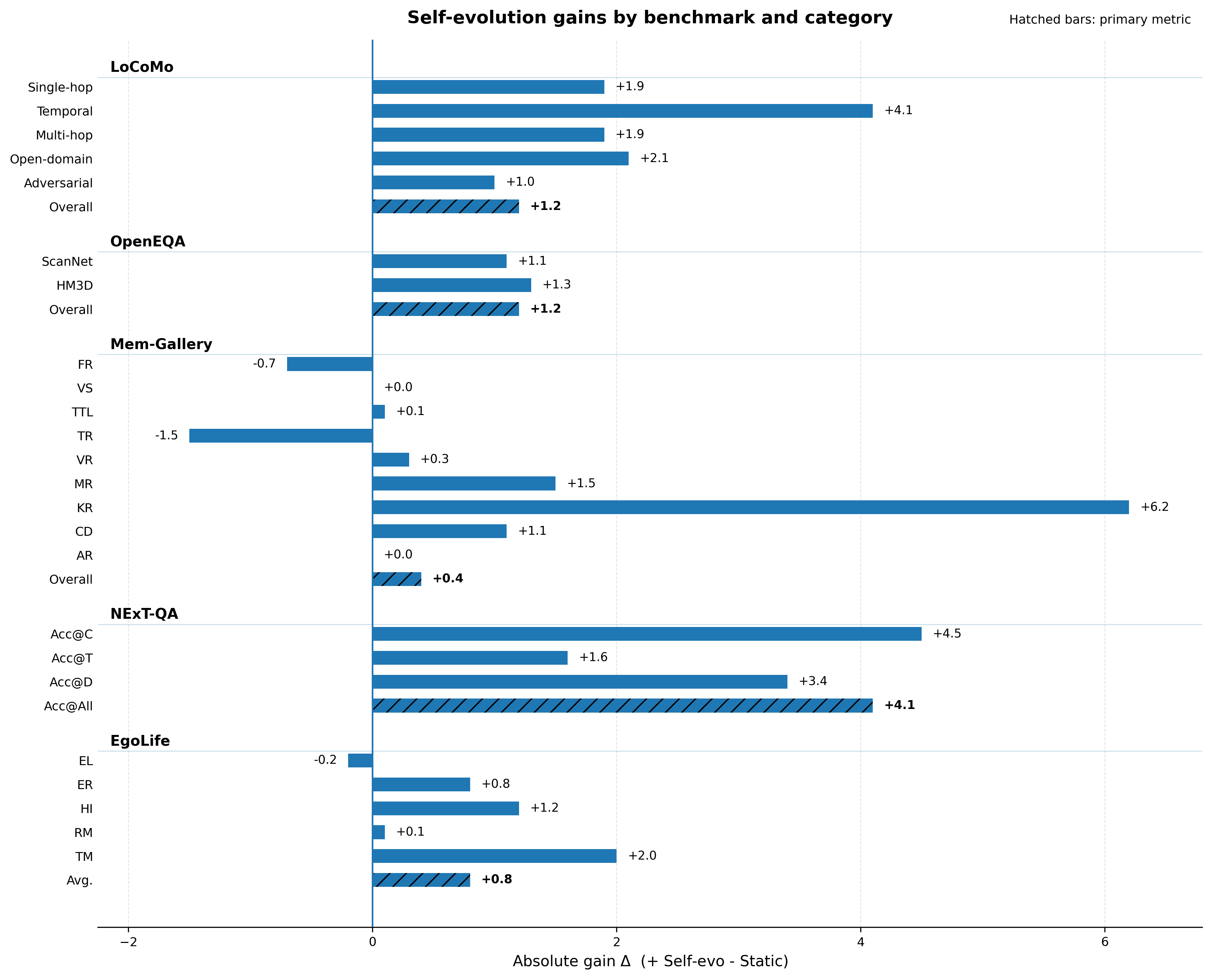}
    \caption{
    Self-evolution gains by benchmark and category.
    Bars show absolute score changes from Static to + Self-evo; hatched bars indicate the primary metric for each benchmark.
    }
    \label{fig:memory-selfevo-gains}
\end{figure*}

Because these experiments are conducted in EQA and QA benchmark settings, the self-evolution loop needs ground-truth answers as a post-hoc correctness signal for diagnosing failures.
This ground-truth signal is used only after a split has already been evaluated.
It is not available during inference, is not written into the memory graph, and does not generate assets for the same split.
This is analogous to how a deployed embodied agent would use interaction feedback in the real world: a user may correct a wrong answer, confirm a right answer, or provide follow-up evidence, and the environment itself may reveal task success or failure through state changes, observations, or execution outcomes.
Thus, the benchmark protocol is a controlled substitute for natural embodied feedback rather than a ground-truth leakage channel.

As illustrated in Figure~\ref{fig:lifelong_selfevo_protocol}, each dataset is partitioned into an ordered sequence of disjoint splits
$\mathcal{D}_1,\ldots,\mathcal{D}_T$.
At evaluation split $t$, the self-evolving variant uses only evo-assets promoted from previous splits, denoted $A_{<t}$.
Failure traces from split $t$ are processed only after evaluation on that split is complete, and the resulting assets can only affect later splits:
\begin{align}
    \mathcal{T}_t, G_t
    &= \operatorname{Run}(\mathcal{D}_t; G_{t-1}, A_{<t}), \\
    \Delta A_t
    &= \operatorname{Gate}\!\left(
        \operatorname{Compile}\!\left(
        \operatorname{Propose}\!\left(
        \operatorname{Diagnose}(\mathcal{T}_t)
        \right)\right)\right), \\
    A_{\leq t}
    &= A_{<t} \cup \Delta A_t .
    \label{eq:memory_eval_self_evo_update}
\end{align}
Here $G_t$ is the updated memory graph and $\mathcal{T}_t$ denotes the recorded answer traces and failure traces.
Only $A_{\leq t}$ is available for later splits.

Figure~\ref{fig:memory-selfevo-gains} visualizes the effect of self-evolution by dataset and category.
Across all five benchmarks, self-evolution improves the primary score over the static memory system.
The largest absolute gain appears on NExT-QA, where Acc@All improves by $4.1$ points.
LoCoMo and OpenEQA both improve by $1.2$ points.
Mem-Gallery shows a smaller overall gain of $0.4$ points, but the category-level gains highlight targeted improvements on knowledge resolution, conflict detection, and multi-entity reasoning.
EgoLife improves from $65.4$ to $66.2$ average accuracy, with the largest category-level gain on TaskMaster.

\FloatBarrier

\subsubsection{Discussion}
\label{subsubsec:memory-discussion}

The per-dataset results show that ABot-AgentOS is strongest when a question requires structured evidence rather than a single visual or textual snippet.
On LoCoMo, the static memory graph reaches near-human overall performance and improves over strong conversational memory baselines, suggesting that source-grounded events, temporal links, and provenance-aware retrieval help long-session recall.
On OpenEQA, the memory system improves over the listed scene-memory baselines under an 8-frame budget, showing that compact graph memories can provide useful spatial and semantic grounding even when visual context is limited.
On Mem-Gallery, ABot-AgentOS Static outperforms the listed memory systems overall and is especially competitive on reasoning, conflict detection, and refusal categories, where provenance and multi-modal relation tracking are important.
On NExT-QA, ABot-AgentOS improves over previous memory-based video agents, indicating that typed event and relation memories are useful for causal and temporal video QA.
On EgoLife, the system achieves the strongest average score while using only one retrieved frame, suggesting that the memory retriever can compress long egocentric streams into highly selective evidence.

The self-evolution results in Section~\ref{subsubsec:memory-selfevo} further show that the memory pipeline can improve without changing the base retriever or leaking current-split answers.
The most consistent gains appear in categories that expose systematic pipeline errors: temporal normalization in LoCoMo, embodied scene disambiguation in OpenEQA, relation and conflict handling in Mem-Gallery, and causal-temporal reasoning in NExT-QA.
Because assets are promoted only after gated validation and are used only on later splits, positive gains indicate reusable memory-pipeline improvements rather than post-hoc repair of the current evaluation split.
The same mechanism is directly compatible with real embodied deployment: instead of benchmark ground truth, the agent can use environmental outcomes and human interaction feedback as the correctness signal for future self-evolution.
% \input{sections/sec6_experiment}

% \\
% \\
% \section{Limitations \& Conclusion}
\section{Conclusion}
We presented ABot-AgentOS, a general-purpose robotic agent operating
system that connects foundation VLM/VLA models with physical robot
execution through a deliberative agent layer. Its modular Agent Harness separates
reasoning, context management, skill execution, and verification, while
its multi-modal memory system stores embodied experience as a typed,
source-grounded graph for traceable retrieval. We also introduced
EmbodiedWorldBench, an executable benchmark for long-horizon embodied
tasks across indoor, outdoor, and hybrid scenarios with dynamic events,
NPC interaction, and deterministic scoring.

The results indicate that ABot-AgentOS is useful both as an online
reasoning-execution layer and as a persistent learning substrate. The
graph memory performs strongly across long-term conversational recall,
embodied QA, multi-modal memory, video reasoning, and egocentric
daily-life QA, while failure-driven self-evolution improves later splits
through gated, trace-diagnosed evo-assets. Initial agent evaluations on
an EmbodiedWorldBench subset further suggest that hierarchical control,
task memory, skill feedback, and finish-time verification improve
long-horizon execution. We also outline a deployment-oriented path for
transferring ABot-AgentOS-style tool use to smaller agent models through
sandboxed trajectory distillation, SFT/RL, and judge-guided
optimization.

Several limitations remain. Large-scale real-world validation is still
needed under noisy perception, imperfect actuation, network latency,
safety constraints, and heterogeneous robot embodiments. Although
EmbodiedWorldBench covers diverse scenario types, its scene diversity,
social interaction depth, and current agent evaluation coverage remain
limited; future work will report the complete benchmark evaluation and
release EmbodiedWorldBench for open research use. The memory and
self-evolution pipeline also depends on structured traces and post-hoc
feedback, and privacy-aware edge-cloud memory sharing requires stronger
auditing, user control, access management, and failure analysis. Finally,
the small-model training pipeline should move beyond text-state semantic
sandboxes by incorporating visual observations, multi-modal feedback, and
additional simulation platforms.

Overall, ABot-AgentOS provides a unified foundation for physically
grounded, memory-augmented, and continuously improving embodied agents
that can perceive, remember, reason, act, verify, and adapt over
long-term interaction with the physical world.
\label{sec:conclusion}

\clearpage
\section{Contributions}
\label{sec:contributions}
% \textit{Authors are listed in contribution order within each role.}

\setlength{\parskip}{0pt} % 让段落之间没有额外空隙
\setlength{\itemsep}{0pt} % 如果用itemize
\setlength{\parsep}{0pt}  % 控制段落间距
\begin{multicols}{2}
\raggedcolumns

\subsubsection*{Research}
\begin{itemize}
\item Jiayi Tian
\item Shiao Liu
\item Yuting Xu
\item Jia Lu
\item Yifei Qian
\item Tianlin Zhang
\item Fei Wang
\item Xiuxian Li
\item Wei Zhang
\item Shihui Su
\item Xiaolong Wu
\item Jiaheng Liu
\item Zhaoxiang Zhang
\end{itemize}

\subsubsection*{Engineering}
\begin{itemize}
\item Zihao Guan
\item Honglin Han
\item Di Yang
\item Minqi Gu
\item Yanqing Zhu
\item Zeqian Ye
\item Menglin Yang
\end{itemize}

\columnbreak

\subsubsection*{Benchmark Construction}
\begin{itemize}
\item Jiayi Tian
\item Fei Wang
\item Tianlin Zhang
\item Yuting Xu
\item Xu Hu
\item Xiuxian Li
\item Wei Zhang
\item Jia Lu
\item Yiyan Ji
\item Jingbo Wang
\item Ziteng Feng

\end{itemize}

\subsubsection*{Project Lead}
\begin{itemize}
\item Mingyang Yin$^{\dagger}$
\item Zedong Chu$^{\dagger}$
\end{itemize}

\subsubsection*{Advisor}
\begin{itemize}
\item Mu Xu
\item Wenbin Tang
\end{itemize}

\end{multicols}

\subsubsection*{Acknowledgments}
We extend our sincere gratitude to the broader team for their support, particularly Wenbin Tang, Shihui Su, Zixiao Tang, Zhining Gu, Yang Cai, Linbo Zheng, and Jingjing Ma.

\vfill
\noindent $^{\dagger}$ Corresponding authors: yinmingyang1@163.com, chuzedong.czd@alibaba-inc.com
\clearpage

\bibliographystyle{plainnat}
\bibliography{main}

\begin{thebibliography}{118}
\providecommand{\natexlab}[1]{#1}
\providecommand{\url}[1]{\texttt{#1}}
\expandafter\ifx\csname urlstyle\endcsname\relax
  \providecommand{\doi}[1]{doi: #1}\else
  \providecommand{\doi}{doi: \begingroup \urlstyle{rm}\Url}\fi

\bibitem[Ahn et~al.(2022)Ahn, Brohan, Brown, Chebotar, Cortes, David, Finn, Fu, Gopalakrishnan, Hausman, Herzog, Ho, Hsu, Ibarz, Ichter, Irpan, Jang, Ruano, Jeffrey, Jesmonth, Joshi, Julian, Kalashnikov, Kuang, Lee, Levine, Lu, Luu, Parada, Pastor, Quiambao, Rao, Rettinghouse, Reyes, Sermanet, Sievers, Tan, Toshev, Vanhoucke, Xia, Xiao, Xu, Xu, Yan, and Zeng]{ahn2022saycan}
Michael Ahn, Anthony Brohan, Noah Brown, Yevgen Chebotar, Omar Cortes, Byron David, Chelsea Finn, Chuyuan Fu, Keerthana Gopalakrishnan, Karol Hausman, Alex Herzog, Daniel Ho, Jasmine Hsu, Julian Ibarz, Brian Ichter, Alex Irpan, Eric Jang, Rosario~Jauregui Ruano, Kyle Jeffrey, Sally Jesmonth, Nikhil~J. Joshi, Ryan Julian, Dmitry Kalashnikov, Yuheng Kuang, Kuang-Huei Lee, Sergey Levine, Yao Lu, Linda Luu, Carolina Parada, Peter Pastor, Jornell Quiambao, Kanishka Rao, Jarek Rettinghouse, Diego Reyes, Pierre Sermanet, Nicolas Sievers, Clayton Tan, Alexander Toshev, Vincent Vanhoucke, Fei Xia, Ted Xiao, Peng Xu, Sichun Xu, Mengyuan Yan, and Andy Zeng.
\newblock Do as i can, not as i say: Grounding language in robotic affordances.
\newblock \emph{arXiv preprint arXiv:2204.01691}, 2022.

\bibitem[Ali et~al.(2025)Ali, Nair, Wong, Cui, and Chen]{ali2025graphpad}
Muhammad~Qasim Ali, Saeejith Nair, Alexander Wong, Yuchen Cui, and Yuhao Chen.
\newblock Graphpad: Inference-time 3d scene graph updates for embodied question answering.
\newblock \emph{arXiv preprint arXiv:2506.01174}, 2025.
\newblock URL \url{https://arxiv.org/abs/2506.01174}.

\bibitem[{Anthropic}(2024)]{anthropic2024computeruse}
{Anthropic}.
\newblock Introducing computer use, a new {Claude} 3.5 {Sonnet}, and {Claude} 3.5 {Haiku}.
\newblock \url{https://www.anthropic.com/news/3-5-models-and-computer-use}, 2024.
\newblock Accessed: 2026-06-26.

\bibitem[{Anthropic}(2026)]{anthropic2026advisortool}
{Anthropic}.
\newblock Advisor tool.
\newblock \url{https://platform.claude.com/docs/en/agents-and-tools/tool-use/advisor-tool}, 2026.
\newblock Accessed: 2026-06-26.

\bibitem[Arai and Ichikawa(2026)]{eveagent}
Yamato Arai and Yuma Ichikawa.
\newblock Eve-agent: Evidence-verifiable self-evolving agents.
\newblock \emph{arXiv preprint arXiv:2605.22905}, 2026.

\bibitem[Bai et~al.(2023)Bai, Wang, and CHEN]{bai2023glance}
Ziyi Bai, Ruiping Wang, and Xilin CHEN.
\newblock Glance and focus: Memory prompting for multi-event video question answering.
\newblock In \emph{Thirty-seventh Conference on Neural Information Processing Systems}, 2023.
\newblock URL \url{https://openreview.net/forum?id=J6Niv3yrMq}.

\bibitem[Bei et~al.(2026)Bei, Wei, Ning, Zhao, Liu, Lin, Zhu, Hamann, He, and Tong]{memgallery}
Yuanchen Bei, Tianxin Wei, Xuying Ning, Yanjun Zhao, Zhining Liu, Xiao Lin, Yada Zhu, Hendrik Hamann, Jingrui He, and Hanghang Tong.
\newblock Mem-gallery: Benchmarking multimodal long-term conversational memory for mllm agents.
\newblock \emph{arXiv preprint arXiv:2601.03515}, 2026.

\bibitem[Buch et~al.(2022)Buch, Eyzaguirre, Gaidon, Wu, Fei-Fei, and Niebles]{buch2022revisiting}
Shyamal Buch, Cristobal Eyzaguirre, Adrien Gaidon, Jiajun Wu, Li~Fei-Fei, and Juan~Carlos Niebles.
\newblock {Revisiting the ``Video'' in Video-Language Understanding}.
\newblock In \emph{Proceedings of the IEEE/CVF Conference on Computer Vision and Pattern Recognition (CVPR)}, 2022.

\bibitem[Cao et~al.(2025)Cao, Deng, Yu, Zhou, Liu, Ding, and Zhao]{reme}
Zouying Cao, Jiaji Deng, Li~Yu, Weikang Zhou, Zhaoyang Liu, Bolin Ding, and Hai Zhao.
\newblock Remember me, refine me: A dynamic procedural memory framework for experience-driven agent evolution.
\newblock \emph{arXiv preprint arXiv:2512.10696}, 2025.

\bibitem[Chen et~al.(2025)Chen, Hu, Bai, Luo, Xue, Ren, Bai, Xie, Chen, Liu, et~al.]{chen2026astranavworldworldmodelforesight}
Jintao Chen, Junjun Hu, Haochen Bai, Minghua Luo, Xinda Xue, Botao Ren, Chengyu Bai, Shichao Xie, Ziyi Chen, Fei Liu, et~al.
\newblock Astranav-world: World model for foresight control and consistency, 2025.

\bibitem[Chen et~al.(2024)Chen, Zaharia, and Zou]{chen2023frugalgpt}
Lingjiao Chen, Matei Zaharia, and James Zou.
\newblock {FrugalGPT}: How to use large language models while reducing cost and improving performance.
\newblock \emph{Transactions on Machine Learning Research}, 2024.
\newblock URL \url{https://openreview.net/forum?id=cSimKw5p6R}.

\bibitem[Chen et~al.(2022)Chen, Hu, Chen, Verga, and Cohen]{chen2022murag}
Wenhu Chen, Hexiang Hu, Xi~Chen, Pat Verga, and William~W. Cohen.
\newblock Murag: Multimodal retrieval-augmented generator for open question answering over images and text.
\newblock In \emph{Proceedings of the 2022 Conference on Empirical Methods in Natural Language Processing}, pages 5558--5570. Association for Computational Linguistics, 2022.
\newblock URL \url{https://aclanthology.org/2022.emnlp-main.375}.

\bibitem[Chen et~al.(2026{\natexlab{a}})Chen, Xie, Gu, Jia, Luo, Liu, Chu, Shen, Wu, and Xu]{chen2026explorelikehumansautonomous}
Xu~Chen, Shichao Xie, Zhining Gu, Lu~Jia, Minghua Luo, Fei Liu, Zedong Chu, Yanfen Shen, Xiaolong Wu, and Mu~Xu.
\newblock Explore like humans: Autonomous exploration with online sg-memo construction for embodied agents, 2026{\natexlab{a}}.
\newblock URL \url{https://arxiv.org/abs/2604.19034}.

\bibitem[Chen et~al.(2026{\natexlab{b}})Chen, Guo, Chu, Luo, Shen, Sun, Hu, Xie, Kuan, Shi, et~al.]{Chen_2026_CVPR}
Ziyi Chen, Yingnan Guo, Zedong Chu, Minghua Luo, Yanfen Shen, Mingchao Sun, Junjun Hu, Shichao Xie, Yang Kuan, Pei Shi, et~al.
\newblock Socialnav: Training human-inspired foundation model for socially-aware embodied navigation.
\newblock pages 28796--28806, 2026{\natexlab{b}}.

\bibitem[Chhikara et~al.(2025)Chhikara, Khant, Aryan, Singh, and Yadav]{mem0}
Prateek Chhikara, Dev Khant, Saket Aryan, Taranjeet Singh, and Deshraj Yadav.
\newblock Mem0: Building production-ready ai agents with scalable long-term memory.
\newblock \emph{arXiv preprint arXiv:2504.19413}, 2025.

\bibitem[Chu et~al.(2025)Chu, Li, and Chua]{GraphVideoAgent}
Meng Chu, Yicong Li, and Tat-Seng Chua.
\newblock Graphvideoagent: Enhancing long-form video understanding with entity relation graphs.
\newblock In \emph{Proceedings of the 33rd ACM International Conference on Multimedia}, MM '25, page 4639–4648, New York, NY, USA, 2025. Association for Computing Machinery.
\newblock ISBN 9798400720352.
\newblock \doi{10.1145/3746027.3755537}.
\newblock URL \url{https://doi.org/10.1145/3746027.3755537}.

\bibitem[Chu et~al.(2026{\natexlab{a}})Chu, Xie, Wu, Shen, Luo, Wang, Liu, Leng, Hu, Yin, Lu, Guo, Yang, Han, Chen, Zhu, Zhao, Liu, Yang, He, Wang, Cai, Zhang, Gao, Liu, Sun, Jiang, Wang, Liu, Pan, Han, Gu, Yang, Zhang, Jing, Guan, Guo, Liu, Yang, Yang, Yang, Xing, Li, and Xu]{chu2026abotn0technicalreportvla}
Zedong Chu, Shichao Xie, Xiaolong Wu, Yanfen Shen, Minghua Luo, Zhengbo Wang, Fei Liu, Xiaoxu Leng, Junjun Hu, Mingyang Yin, Jia Lu, Yingnan Guo, Kai Yang, Jiawei Han, Xu~Chen, Yanqing Zhu, Yuxiang Zhao, Xin Liu, Yirong Yang, Ye~He, Jiahang Wang, Yang Cai, Tianlin Zhang, Li~Gao, Liu Liu, Mingchao Sun, Fan Jiang, Chiyu Wang, Zhicheng Liu, Hongyu Pan, Honglin Han, Zhining Gu, Kuan Yang, Jianfang Zhang, Di~Jing, Zihao Guan, Wei Guo, Guoqing Liu, Di~Yang, Xiangpo Yang, Menglin Yang, Hongguang Xing, Weiguo Li, and Mu~Xu.
\newblock Abot-n0: Technical report on the vla foundation model for versatile embodied navigation, 2026{\natexlab{a}}.
\newblock URL \url{https://arxiv.org/abs/2602.11598}.

\bibitem[Chu et~al.(2026{\natexlab{b}})Chu, Xie, Wu, Shen, Luo, Wang, Liu, Leng, Hu, Yin, et~al.]{chu2026abot}
Zedong Chu, Shichao Xie, Xiaolong Wu, Yanfen Shen, Minghua Luo, Zhengbo Wang, Fei Liu, Xiaoxu Leng, Junjun Hu, Mingyang Yin, et~al.
\newblock Abot-n0: Technical report on the vla foundation model for versatile embodied navigation.
\newblock \emph{arXiv preprint arXiv:2602.11598}, 2026{\natexlab{b}}.

\bibitem[Fan et~al.(2025)Fan, Ma, Wu, Du, Li, Gao, and Li]{fan2025videoagent}
Yue Fan, Xiaojian Ma, Rujie Wu, Yuntao Du, Jiaqi Li, Zhi Gao, and Qing Li.
\newblock Videoagent: A memory-augmented multimodal agent for video understanding.
\newblock In \emph{European Conference on Computer Vision}, pages 75--92. Springer, 2025.

\bibitem[Feng et~al.(2025)Feng, Xue, Liu, and An]{feng2025gigpo}
Lang Feng, Zhenghai Xue, Tingcong Liu, and Bo~An.
\newblock Group-in-group policy optimization for llm agent training.
\newblock \emph{arXiv preprint arXiv:2505.10978}, 2025.
\newblock URL \url{https://arxiv.org/abs/2505.10978}.

\bibitem[Fisher(2025)]{fisher2025ngm}
Matthew Fisher.
\newblock Neural graph memory: A structured approach to long-term memory in multimodal agents.
\newblock Zenodo preprint, 2025.
\newblock URL \url{https://zenodo.org/records/16809331}.

\bibitem[Gao et~al.(2024)Gao, Zhao, Zhang, Mao, Zhang, Zheng, Man, Fang, Zhou, Cui, Chen, and Li]{gao2024embodiedcity}
Chen Gao, Baining Zhao, Weichen Zhang, Jinzhu Mao, Jun Zhang, Zhiheng Zheng, Fanhang Man, Jianjie Fang, Zile Zhou, Jinqiang Cui, Xinlei Chen, and Yong Li.
\newblock {EmbodiedCity}: A benchmark platform for embodied agent in real-world city environment.
\newblock \emph{arXiv preprint arXiv:2410.09604}, 2024.

\bibitem[Gao et~al.(2023{\natexlab{a}})Gao, Ji, Zhou, Lin, Chen, Fan, and Shou]{gao2023assistgptgeneralmultimodalassistant}
Difei Gao, Lei Ji, Luowei Zhou, Kevin~Qinghong Lin, Joya Chen, Zihan Fan, and Mike~Zheng Shou.
\newblock Assistgpt: A general multi-modal assistant that can plan, execute, inspect, and learn, 2023{\natexlab{a}}.
\newblock URL \url{https://arxiv.org/abs/2306.08640}.

\bibitem[Gao et~al.(2023{\natexlab{b}})Gao, Zhou, Ji, Zhu, Yang, and Shou]{MIST}
Difei Gao, Luowei Zhou, Lei Ji, Linchao Zhu, Yi~Yang, and Mike~Zheng Shou.
\newblock Mist : Multi-modal iterative spatial-temporal transformer for long-form video question answering.
\newblock In \emph{2023 IEEE/CVF Conference on Computer Vision and Pattern Recognition (CVPR)}, pages 14773--14783, 2023{\natexlab{b}}.
\newblock \doi{10.1109/CVPR52729.2023.01419}.

\bibitem[Gao et~al.(2025{\natexlab{a}})Gao, Geng, Hua, Hu, Juan, Liu, Liu, Qiu, Qi, Wu, et~al.]{selfevolvingsurvey}
Huan-ang Gao, Jiayi Geng, Wenyue Hua, Mengkang Hu, Xinzhe Juan, Hongzhang Liu, Shilong Liu, Jiahao Qiu, Xuan Qi, Yiran Wu, et~al.
\newblock A survey of self-evolving agents: What, when, how, and where to evolve on the path to artificial super intelligence.
\newblock \emph{arXiv preprint arXiv:2507.21046}, 2025{\natexlab{a}}.

\bibitem[Gao et~al.(2025{\natexlab{b}})Gao, Li, You, Liu, Li, Chen, Chen, Tang, Wang, Yang, et~al.]{gao2025openfly}
Yunpeng Gao, Chenhui Li, Zhongrui You, Junli Liu, Zhen Li, Pengan Chen, Qizhi Chen, Zhonghan Tang, Liansheng Wang, Penghui Yang, et~al.
\newblock Openfly: A comprehensive platform for aerial vision-language navigation.
\newblock \emph{arXiv preprint arXiv:2502.18041}, 2025{\natexlab{b}}.

\bibitem[Gong et~al.(2026)Gong, Zhang, Zhao, Sun, Shen, Chu, Gu, Guo, Cheng, Li, Niu, Zhu, Wu, Li, and Xu]{gong2026poinavbenchmarkingenhancingfinalmeters}
Ruiyan Gong, Meisheng Zhang, Yuxiang Zhao, Mingchao Sun, Yanfen Shen, Zedong Chu, Zhining Gu, Wei Guo, Xiaolong Cheng, Qiming Li, Kangning Niu, Yanqing Zhu, Xiaolong Wu, Tianlun Li, and Mu~Xu.
\newblock Poinav: Benchmarking and enhancing final-meters arrival in real-world vision-language navigation, 2026.
\newblock URL \url{https://arxiv.org/abs/2605.28237}.

\bibitem[Google(2025)]{gemini2_5pro}
Google.
\newblock Google gemini 2.5 pro.
\newblock \url{https://docs.cloud.google.com/gemini-enterprise-agent-platform/models/gemini/2-5-pro?hl=zh-cn}, 2025.
\newblock Gemini Enterprise Agent Platform.

\bibitem[{Google DeepMind}(2025)]{deepmind2025geminirobotics}
{Google DeepMind}.
\newblock {Gemini Robotics} brings {AI} into the physical world.
\newblock \url{https://deepmind.google/blog/gemini-robotics-brings-ai-into-the-physical-world/}, 2025.
\newblock Accessed: 2026-06-26.

\bibitem[Han et~al.(2026)Han, Xie, Ma, et~al.]{han2026swetrace}
Hao Han, Jin Xie, Xuehao Ma, et~al.
\newblock Swe-trace: Optimizing long-horizon swe agents through rubric process reward models and heuristic test-time scaling.
\newblock \emph{arXiv preprint arXiv:2604.14820}, 2026.

\bibitem[He et~al.(2026)He, Wang, Du, Bai, Cao, Cheng, and Zheng]{he2026learning}
Shan He, Runze Wang, Zhuoyun Du, Huiyu Bai, Zouying Cao, Yu~Cheng, and Bo~Zheng.
\newblock Learning to evolve: A self-improving framework for multi-agent systems via textual parameter graph optimization.
\newblock \emph{arXiv preprint arXiv:2604.20714}, 2026.

\bibitem[Hong et~al.(2025)Hong, Sun, Li, Yao, Wu, Chien, Yin, Wu, Wang, and Chang]{hong2025embodiedweb}
Yining Hong, Rui Sun, Bingxuan Li, Xingcheng Yao, Maxine Wu, Alexander Chien, Da~Yin, Ying~Nian Wu, Zhecan~James Wang, and Kai-Wei Chang.
\newblock Embodied web agents: Bridging physical-digital realms for integrated agent intelligence.
\newblock \emph{arXiv preprint arXiv:2506.15677}, 2025.

\bibitem[Huang et~al.(2022)Huang, Xia, Xiao, Chan, Liang, Florence, Zeng, Tompson, Mordatch, Chebotar, Sermanet, Brown, Jackson, Luu, Levine, Hausman, and Ichter]{huang2022innermonologue}
Wenlong Huang, Fei Xia, Ted Xiao, Harris Chan, Jacky Liang, Pete Florence, Andy Zeng, Jonathan Tompson, Igor Mordatch, Yevgen Chebotar, Pierre Sermanet, Noah Brown, Tomas Jackson, Linda Luu, Sergey Levine, Karol Hausman, and Brian Ichter.
\newblock Inner monologue: Embodied reasoning through planning with language models.
\newblock \emph{arXiv preprint arXiv:2207.05608}, 2022.

\bibitem[Jain et~al.(2025)Jain, Maheshwari, Yu, Hwu, and Shi]{jain2025augustus}
Jitesh Jain, Shubham Maheshwari, Ning Yu, Wen-mei Hwu, and Humphrey Shi.
\newblock Augustus: An llm-driven multimodal agent system with contextualized user memory.
\newblock \emph{arXiv preprint arXiv:2510.15261}, 2025.
\newblock URL \url{https://arxiv.org/abs/2510.15261}.

\bibitem[Ji et~al.(2025)Ji, Tan, Shi, Hao, Zhang, Zhang, Wang, Zhao, Mu, An, et~al.]{ji2025robobrain}
Yuheng Ji, Huajie Tan, Jiayu Shi, Xiaoshuai Hao, Yuan Zhang, Hengyuan Zhang, Pengwei Wang, Mengdi Zhao, Yao Mu, Pengju An, et~al.
\newblock Robobrain: A unified brain model for robotic manipulation from abstract to concrete.
\newblock In \emph{Proceedings of the IEEE/CVF Conference on Computer Vision and Pattern Recognition}, pages 1724--1734, 2025.

\bibitem[Jiang et~al.(2025)Jiang, Yuan, Liu, Lu, Cui, Liu, Cheng, Gao, Xu, and Zhao]{jiang2025galaxea}
Tao Jiang, Tianyuan Yuan, Yicheng Liu, Chenhao Lu, Jianning Cui, Xiao Liu, Shuiqi Cheng, Jiyang Gao, Huazhe Xu, and Hang Zhao.
\newblock Galaxea open-world dataset and g0 dual-system vla model.
\newblock \emph{arXiv preprint arXiv:2509.00576}, 2025.

\bibitem[Jimenez et~al.(2023)Jimenez, Yang, Wettig, Yao, Pei, Press, and Narasimhan]{jimenez2023swebench}
Carlos~E. Jimenez, John Yang, Alexander Wettig, Shunyu Yao, Kexin Pei, Ofir Press, and Karthik Narasimhan.
\newblock Swe-bench: Can language models resolve real-world github issues?
\newblock \emph{arXiv preprint arXiv:2310.06770}, 2023.

\bibitem[Kang et~al.(2025)Kang, Ji, Zhao, and Bai]{kang2025memoryos}
Jiazheng Kang, Mingming Ji, Zhe Zhao, and Ting Bai.
\newblock Memory os of ai agent.
\newblock In \emph{Proceedings of the 2025 Conference on Empirical Methods in Natural Language Processing}, pages 25961--25970. Association for Computational Linguistics, 2025.
\newblock URL \url{https://aclanthology.org/2025.emnlp-main.1318}.

\bibitem[Khattab et~al.(2023)Khattab, Singhvi, Maheshwari, Zhang, Santhanam, Vardhamanan, Haq, Sharma, Joshi, Moazam, et~al.]{khattab2023dspy}
Omar Khattab, Arnav Singhvi, Paridhi Maheshwari, Zhiyuan Zhang, Keshav Santhanam, Sri Vardhamanan, Saiful Haq, Ashutosh Sharma, Thomas~T. Joshi, Hanna Moazam, et~al.
\newblock Dspy: Compiling declarative language model calls into self-improving pipelines.
\newblock \emph{arXiv preprint arXiv:2310.03714}, 2023.

\bibitem[Kim et~al.(2023)Kim, Kim, Lee, and Seo]{kim2023semiparametricvideogroundedtextgeneration}
Sungdong Kim, Jin-Hwa Kim, Jiyoung Lee, and Minjoon Seo.
\newblock Semi-parametric video-grounded text generation, 2023.
\newblock URL \url{https://arxiv.org/abs/2301.11507}.

\bibitem[Kim et~al.(2026)Kim, Lee, Kim, Kim, Nam, Kwon, Wang, Choe, and Oh]{kim2026gaussexplorer}
Yu-Ji Kim, Dahye Lee, Jun-Seong Kim, GeonU Kim, Hyeon-Woo Nam, Yongjin Kwon, Yu-Chiang~Frank Wang, Jaesung Choe, and Tae-Hyun Oh.
\newblock Gaussexplorer: 3d gaussian splatting for embodied exploration and reasoning.
\newblock \emph{arXiv preprint arXiv:2601.13132}, 2026.
\newblock URL \url{https://arxiv.org/abs/2601.13132}.

\bibitem[Lee et~al.(2025)Lee, Miyanishi, Kurita, Sakamoto, Azuma, Matsuo, and Inoue]{lee2025citynav}
Jungdae Lee, Taiki Miyanishi, Shuhei Kurita, Koya Sakamoto, Daichi Azuma, Yutaka Matsuo, and Nakamasa Inoue.
\newblock Citynav: A large-scale dataset for real-world aerial navigation.
\newblock In \emph{Proceedings of the IEEE/CVF International Conference on Computer Vision}, pages 5912--5922, 2025.

\bibitem[Lei et~al.(2025)Lei, Cai, Yang, Wu, Ren, Cui, Tan, Hong, Hu, Zhu, et~al.]{lei2025robomemory}
Mingcong Lei, Honghao Cai, Yuyuan Yang, Yimou Wu, Jinke Ren, Zezhou Cui, Liangchen Tan, Junkun Hong, Gehan Hu, Shuangyu Zhu, et~al.
\newblock Robomemory: A brain-inspired multi-memory agentic framework for interactive environmental learning in physical embodied systems.
\newblock \emph{arXiv preprint arXiv:2508.01415}, 2025.

\bibitem[Li et~al.(2023)Li, Zhang, Wong, Gokmen, Srivastava, Mart{\'\i}n-Mart{\'\i}n, Wang, Levine, Lingelbach, Sun, et~al.]{li2024behavior1k}
Chengshu Li, Ruohan Zhang, Josiah Wong, Cem Gokmen, Sanjana Srivastava, Roberto Mart{\'\i}n-Mart{\'\i}n, Chen Wang, Gabrael Levine, Michael Lingelbach, Jiankai Sun, et~al.
\newblock Behavior-1k: A benchmark for embodied ai with 1,000 everyday activities and realistic simulation.
\newblock pages 80--93, 2023.

\bibitem[Li et~al.(2025)Li, Mohamud, Sun, Wu, and Boulet]{li2025multiagent}
Yuran Li, Jama~Hussein Mohamud, Chongren Sun, Di~Wu, and Benoit Boulet.
\newblock Leveraging llms as meta-judges: A multi-agent framework for evaluating llm judgments.
\newblock \emph{arXiv preprint arXiv:2504.17087}, 2025.

\bibitem[Li et~al.(2026)Li, He, Zhang, and Gong]{evimem}
Yuyang Li, Yime He, Zeyu Zhang, and Dong Gong.
\newblock Evimem: Evidence-gap-driven iterative retrieval for long-term conversational memory.
\newblock \emph{arXiv preprint arXiv:2604.27695}, 2026.

\bibitem[Lin et~al.(2026)Lin, Zhang, Lu, Liu, Tang, He, Zhang, and Wang]{memma}
Minhua Lin, Zhiwei Zhang, Hanqing Lu, Hui Liu, Xianfeng Tang, Qi~He, Xiang Zhang, and Suhang Wang.
\newblock Memma: Coordinating the memory cycle through multi-agent reasoning and in-situ self-evolution.
\newblock \emph{arXiv preprint arXiv:2603.18718}, 2026.

\bibitem[Liu et~al.(2026{\natexlab{a}})Liu, Xie, Luo, Chu, Hu, Wu, and Xu]{Liu_2026_CVPR}
Fei Liu, Shichao Xie, Minghua Luo, Zedong Chu, Junjun Hu, Xiaolong Wu, and Mu~Xu.
\newblock Navforesee: A unified vision-language world model for hierarchical planning and dual-horizon navigation prediction.
\newblock pages 32431--32440, 2026{\natexlab{a}}.

\bibitem[Liu et~al.(2026{\natexlab{b}})Liu, Su, Xia, Han, Zheng, Xie, Ding, and Yao]{simplemem}
Jiaqi Liu, Yaofeng Su, Peng Xia, Siwei Han, Zeyu Zheng, Cihang Xie, Mingyu Ding, and Huaxiu Yao.
\newblock Simplemem: Efficient lifelong memory for llm agents.
\newblock \emph{arXiv preprint arXiv:2601.02553}, 2026{\natexlab{b}}.

\bibitem[Liu et~al.(2026{\natexlab{c}})Liu, Ye, Xia, Zheng, Xie, Ding, and Yao]{evolvemem}
Jiaqi Liu, Xinyu Ye, Peng Xia, Zeyu Zheng, Cihang Xie, Mingyu Ding, and Huaxiu Yao.
\newblock Evolvemem: Self-evolving memory architecture via autoresearch for llm agents.
\newblock \emph{arXiv preprint arXiv:2605.13941}, 2026{\natexlab{c}}.

\bibitem[Liu et~al.(2026{\natexlab{d}})Liu, Li, Deng, Chen, Zhang, Ma, Guo, Li, Zhang, and Feng]{liu2026wanderland}
Xinhao Liu, Jiaqi Li, Youming Deng, Ruxin Chen, Yingjia Zhang, Yifei Ma, Li~Guo, Yiming Li, Jing Zhang, and Chen Feng.
\newblock Wanderland: Geometrically grounded simulation for open-world embodied ai.
\newblock In \emph{Proceedings of the IEEE/CVF Conference on Computer Vision and Pattern Recognition}, pages 1041--1052, 2026{\natexlab{d}}.

\bibitem[Maharana et~al.(2024)Maharana, Lee, Tulyakov, Bansal, Barbieri, and Fang]{locomo}
Adyasha Maharana, Dong-Ho Lee, Sergey Tulyakov, Mohit Bansal, Francesco Barbieri, and Yuwei Fang.
\newblock Evaluating very long-term conversational memory of llm agents.
\newblock \emph{arXiv preprint arXiv:2402.17753}, 2024.

\bibitem[Majumdar et~al.(2024)Majumdar, Ajay, Zhang, Putta, Yenamandra, Henaff, Silwal, McVay, Maksymets, Arnaud, et~al.]{openeqa}
Arjun Majumdar, Anurag Ajay, Xiaohan Zhang, Pranav Putta, Sriram Yenamandra, Mikael Henaff, Sneha Silwal, Paul McVay, Oleksandr Maksymets, Sergio Arnaud, et~al.
\newblock Openeqa: Embodied question answering in the era of foundation models.
\newblock In \emph{Proceedings of the IEEE/CVF Conference on Computer Vision and Pattern Recognition}, 2024.

\bibitem[Momeni et~al.(2023)Momeni, Caron, Nagrani, Zisserman, and Schmid]{VFC}
Liliane Momeni, Mathilde Caron, Arsha Nagrani, Andrew Zisserman, and Cordelia Schmid.
\newblock Verbs in action: Improving verb understanding in video-language models.
\newblock In \emph{2023 IEEE/CVF International Conference on Computer Vision (ICCV)}, pages 15533--15545, 2023.
\newblock \doi{10.1109/ICCV51070.2023.01428}.

\bibitem[Ong and Jang(2025)]{ong2025reqa}
Hyobin Ong and Minsu Jang.
\newblock R-eqa: Retrieval-augmented generation for embodied question answering, 2025.
\newblock URL \url{https://embodied-ai.org/papers/2025/6_R_EQA_Retrieval_Augmented_Ge.pdf}.
\newblock Embodied AI Workshop.

\bibitem[Ong et~al.(2024)Ong, Almahairi, Wu, Chiang, Wu, Gonzalez, Kadous, and Stoica]{ong2024routellm}
Isaac Ong, Amjad Almahairi, Vincent Wu, Wei-Lin Chiang, Tianhao Wu, Joseph~E. Gonzalez, M.~Waleed Kadous, and Ion Stoica.
\newblock {RouteLLM}: Learning to route {LLMs} with preference data.
\newblock \emph{arXiv preprint arXiv:2406.18665}, 2024.

\bibitem[OpenAI(2025)]{gpt4_1}
OpenAI.
\newblock Openai gpt-4.1.
\newblock \url{https://developers.openai.com/api/docs/models/gpt-4.1}, 2025.
\newblock OpenAI Developers.

\bibitem[{OpenAI}(2025{\natexlab{a}})]{openai2025chatgptagent}
{OpenAI}.
\newblock Introducing {ChatGPT} agent: Bridging research and action.
\newblock \url{https://openai.com/index/introducing-chatgpt-agent/}, 2025{\natexlab{a}}.
\newblock Accessed: 2026-06-26.

\bibitem[{OpenAI}(2025{\natexlab{b}})]{openai2025operator}
{OpenAI}.
\newblock Introducing {Operator}.
\newblock \url{https://openai.com/index/introducing-operator/}, 2025{\natexlab{b}}.
\newblock Accessed: 2026-06-26.

\bibitem[OpenAI(2026)]{gpt5_4mini}
OpenAI.
\newblock Openai gpt-5.4 mini.
\newblock \url{https://developers.openai.com/api/docs/models/gpt-5.4-mini}, 2026.
\newblock OpenAI Developers.

\bibitem[Packer et~al.(2023)Packer, Wooders, Lin, Fang, Patil, Stoica, and Gonzalez]{packer2023memgpt}
Charles Packer, Sarah Wooders, Kevin Lin, Vivian Fang, Shishir~G. Patil, Ion Stoica, and Joseph~E. Gonzalez.
\newblock Memgpt: Towards llms as operating systems.
\newblock \emph{arXiv preprint arXiv:2310.08560}, 2023.

\bibitem[Park et~al.(2023)Park, O'Brien, Cai, Morris, Liang, and Bernstein]{generativeagents}
Joon~Sung Park, Joseph~C. O'Brien, Carrie~J. Cai, Meredith~Ringel Morris, Percy Liang, and Michael~S. Bernstein.
\newblock Generative agents: Interactive simulacra of human behavior.
\newblock In \emph{Proceedings of the 36th Annual ACM Symposium on User Interface Software and Technology}, 2023.

\bibitem[Qin et~al.(2023)Qin, Liang, Ye, Zhu, Yan, Lu, Lin, Cong, Tang, Qian, Zhao, Hong, Tian, Xie, Zhou, Gerstein, Li, Liu, and Sun]{qin2023toolllm}
Yujia Qin, Shihao Liang, Yining Ye, Kunlun Zhu, Lan Yan, Yaxi Lu, Yankai Lin, Xin Cong, Xiangru Tang, Bill Qian, Sihan Zhao, Lauren Hong, Runchu Tian, Ruobing Xie, Jie Zhou, Mark Gerstein, Dahai Li, Zhiyuan Liu, and Maosong Sun.
\newblock Toolllm: Facilitating large language models to master 16000+ real-world apis.
\newblock \emph{arXiv preprint arXiv:2307.16789}, 2023.

\bibitem[Rege et~al.(2026)Rege, Sadhu, Li, Li, Vinayak, Chai, Lee, and Kim]{rege2026EGAgent}
Aniket Rege, Arka Sadhu, Yuliang Li, Kejie Li, Ramya~Korlakai Vinayak, Yuning Chai, Yong~Jae Lee, and Hyo~Jin Kim.
\newblock Agentic very long video understanding.
\newblock \emph{arXiv preprint arXiv:2601.18157}, 2026.

\bibitem[Salama et~al.(2025)Salama, Cai, Yuan, Currey, Sunkara, Zhang, and Benajiba]{salama2025meminsight}
Rana Salama, Jason Cai, Michelle Yuan, Anna Currey, Monica Sunkara, Yi~Zhang, and Yassine Benajiba.
\newblock Meminsight: Autonomous memory augmentation for llm agents.
\newblock \emph{arXiv preprint arXiv:2503.21760}, 2025.
\newblock URL \url{https://arxiv.org/abs/2503.21760}.

\bibitem[Schick et~al.(2023)Schick, Dwivedi-Yu, Dess{\`i}, Raileanu, Lomeli, Hambro, Zettlemoyer, Cancedda, and Scialom]{schick2023toolformer}
Timo Schick, Jane Dwivedi-Yu, Roberto Dess{\`i}, Roberta Raileanu, Maria Lomeli, Eric Hambro, Luke Zettlemoyer, Nicola Cancedda, and Thomas Scialom.
\newblock {Toolformer}: Language models can teach themselves to use tools.
\newblock In \emph{Advances in Neural Information Processing Systems}, 2023.
\newblock URL \url{https://proceedings.neurips.cc/paper_files/paper/2023/hash/d842425e4bf79ba039352da0f658a906-Abstract-Conference.html}.

\bibitem[Shi et~al.(2025)Shi, Ichter, Equi, Ke, Pertsch, Vuong, Tanner, Walling, Wang, Fusai, et~al.]{shi2025hi}
Lucy~Xiaoyang Shi, Brian Ichter, Michael Equi, Liyiming Ke, Karl Pertsch, Quan Vuong, James Tanner, Anna Walling, Haohuan Wang, Niccolo Fusai, et~al.
\newblock Hi robot: Open-ended instruction following with hierarchical vision-language-action models.
\newblock \emph{arXiv preprint arXiv:2502.19417}, 2025.

\bibitem[Silva et~al.(2026)Silva, Mendes, and Oliveira]{silva2026metajudging}
Hugo Silva, Mateus Mendes, and Hugo~Gonçalo Oliveira.
\newblock Meta-judging with large language models: Concepts, methods, and challenges.
\newblock \emph{arXiv preprint arXiv:2601.17312}, 2026.

\bibitem[Song et~al.(2026)Song, Chang, Dong, Zhu, Dou, and Wen]{song2026envscaler}
Xiaoshuai Song, Haofei Chang, Guanting Dong, Yutao Zhu, Zhicheng Dou, and Ji-Rong Wen.
\newblock Envscaler: Scaling tool-interactive environments for llm agent via programmatic synthesis.
\newblock \emph{arXiv preprint arXiv:2601.05808}, 2026.

\bibitem[Sun et~al.(2025)Sun, Wu, Fu, Song, Shi, et~al.]{sun2025tongsim}
Zhe Sun, Kunlun Wu, Chuanjian Fu, Zeming Song, Langyong Shi, et~al.
\newblock {TongSIM}: A general platform for simulating intelligent machines.
\newblock \emph{arXiv preprint arXiv:2512.20206}, 2025.

\bibitem[Surís et~al.(2023)Surís, Menon, and Vondrick]{ViperGPT}
Dídac Surís, Sachit Menon, and Carl Vondrick.
\newblock Vipergpt: Visual inference via python execution for reasoning.
\newblock In \emph{2023 IEEE/CVF International Conference on Computer Vision (ICCV)}, pages 11854--11864, 2023.
\newblock \doi{10.1109/ICCV51070.2023.01092}.

\bibitem[Team et~al.(2025{\natexlab{a}})Team, Cao, Tan, Ji, Chen, Lin, Li, Cao, Wang, Zhou, et~al.]{team2025robobrain}
BAAI~RoboBrain Team, Mingyu Cao, Huajie Tan, Yuheng Ji, Xiansheng Chen, Minglan Lin, Zhiyu Li, Zhou Cao, Pengwei Wang, Enshen Zhou, et~al.
\newblock Robobrain 2.0 technical report.
\newblock \emph{arXiv preprint arXiv:2507.02029}, 2025{\natexlab{a}}.

\bibitem[Team et~al.(2025{\natexlab{b}})Team, Abdolmaleki, Abeyruwan, Ainslie, Alayrac, Arenas, Balakrishna, Batchelor, Bewley, Bingham, et~al.]{team2025gemini}
Gemini~Robotics Team, Abbas Abdolmaleki, Saminda Abeyruwan, Joshua Ainslie, Jean-Baptiste Alayrac, Montserrat~Gonzalez Arenas, Ashwin Balakrishna, Nathan Batchelor, Alex Bewley, Jeff Bingham, et~al.
\newblock Gemini robotics 1.5: Pushing the frontier of generalist robots with advanced embodied reasoning, thinking, and motion transfer.
\newblock \emph{arXiv preprint arXiv:2510.03342}, 2025{\natexlab{b}}.

\bibitem[Team(2026)]{team2026qwen3}
Qwen Team.
\newblock Qwen3.5-omni technical report.
\newblock \emph{arXiv preprint arXiv:2604.15804}, 2026.

\bibitem[Wang et~al.(2026)Wang, Liu, Liu, Tang, Wang, Gao, Zheng, Zhang, Yu, Liu, et~al.]{wang2026outcome}
Binghai Wang, Yantao Liu, Yuxuan Liu, Tianyi Tang, Shenzhi Wang, Chang Gao, Chujie Zheng, Yichang Zhang, Le~Yu, Shixuan Liu, et~al.
\newblock Outcome accuracy is not enough: Aligning the reasoning process of reward models.
\newblock \emph{arXiv preprint arXiv:2602.04649}, 2026.

\bibitem[Wang et~al.(2023)Wang, Xie, Jiang, Mandlekar, Xiao, Zhu, Fan, and Anandkumar]{wang2023voyager}
Guanzhi Wang, Yuqi Xie, Yunfan Jiang, Ajay Mandlekar, Chaowei Xiao, Yuke Zhu, Linxi Fan, and Anima Anandkumar.
\newblock {Voyager}: An open-ended embodied agent with large language models.
\newblock \emph{arXiv preprint arXiv:2305.16291}, 2023.

\bibitem[Wang et~al.(2025{\natexlab{a}})Wang, Liu, Chen, Da, Qian, Tram, and Soh]{wang2025genie}
Jiaming Wang, Diwen Liu, Jizhuo Chen, Jiaxuan Da, Nuowen Qian, Minh~Man Tram, and Harold Soh.
\newblock Genie: A generalizable navigation system for in-the-wild environments.
\newblock \emph{IEEE Robotics and Automation Letters}, 2025{\natexlab{a}}.

\bibitem[Wang et~al.(2025{\natexlab{b}})Wang, Lu, Liu, Jiang, Li, Zhang, Zheng, Yu, Chen, and Shen]{wang2025odyssey}
Kaijun Wang, Liqin Lu, Mingyu Liu, Jianuo Jiang, Zeju Li, Bolin Zhang, Wancai Zheng, Xinyi Yu, Hao Chen, and Chunhua Shen.
\newblock {ODYSSEY}: Open-world quadrupeds exploration and manipulation for long-horizon tasks.
\newblock \emph{arXiv preprint arXiv:2508.08240}, 2025{\natexlab{b}}.

\bibitem[Wang et~al.(2022)Wang, Li, Li, He, Huang, Zhao, Zhang, Xu, Liu, Wang, Xing, Chen, Pan, Yu, Wang, Wang, and Qiao]{wang2022internvideogeneralvideofoundation}
Yi~Wang, Kunchang Li, Yizhuo Li, Yinan He, Bingkun Huang, Zhiyu Zhao, Hongjie Zhang, Jilan Xu, Yi~Liu, Zun Wang, Sen Xing, Guo Chen, Junting Pan, Jiashuo Yu, Yali Wang, Limin Wang, and Yu~Qiao.
\newblock Internvideo: General video foundation models via generative and discriminative learning, 2022.
\newblock URL \url{https://arxiv.org/abs/2212.03191}.

\bibitem[Wang and Chen(2025)]{wang2025mirix}
Yu~Wang and Xi~Chen.
\newblock Mirix: Multi-agent memory system for llm-based agents.
\newblock \emph{arXiv preprint arXiv:2507.07957}, 2025.
\newblock URL \url{https://arxiv.org/abs/2507.07957}.

\bibitem[Wang et~al.(2025{\natexlab{c}})Wang, Wang, Wang, Zhang, Li, Yang, Yu, Nguyen, Liu, Gottlieb, Lam, Lu, Cho, Wu, Fei-Fei, Wang, Choi, and Li]{wang2025ragen}
Zihan Wang, Kangrui Wang, Qineng Wang, Pingyue Zhang, Linjie Li, Zhengyuan Yang, Kefan Yu, Minh~Nhat Nguyen, Licheng Liu, Eli Gottlieb, Monica Lam, Yiping Lu, Kyunghyun Cho, Jiajun Wu, Li~Fei-Fei, Lijuan Wang, Yejin Choi, and Manling Li.
\newblock Ragen: Understanding self-evolution in llm agents via multi-turn reinforcement learning.
\newblock \emph{arXiv preprint arXiv:2504.20073}, 2025{\natexlab{c}}.
\newblock URL \url{https://arxiv.org/abs/2504.20073}.

\bibitem[Wei et~al.(2025{\natexlab{a}})Wei, Zeng, Li, et~al.]{wei2025reinforcing}
Quan Wei, Siliang Zeng, Chenliang Li, et~al.
\newblock Reinforcing multi-turn reasoning in llm agents via turn-level reward design.
\newblock \emph{arXiv preprint arXiv:2505.11821}, 2025{\natexlab{a}}.

\bibitem[Wei et~al.(2025{\natexlab{b}})Wei, Sachdeva, Coleman, He, Bei, Ning, Ai, Li, He, Chi, et~al.]{evomemory}
Tianxin Wei, Noveen Sachdeva, Benjamin Coleman, Zhankui He, Yuanchen Bei, Xuying Ning, Mengting Ai, Yunzhe Li, Jingrui He, Ed~H. Chi, et~al.
\newblock Evo-memory: Benchmarking llm agent test-time learning with self-evolving memory.
\newblock \emph{arXiv preprint arXiv:2511.20857}, 2025{\natexlab{b}}.

\bibitem[Wei et~al.(2025{\natexlab{c}})Wei, Yao, Liu, Zhang, Lu, Qiu, Yu, Xu, Zhang, Yin, Yun, and Li]{wei2025webagentr1}
Zhepei Wei, Wenlin Yao, Yao Liu, Weizhi Zhang, Qin Lu, Liang Qiu, Changlong Yu, Puyang Xu, Chao Zhang, Bing Yin, Hyokun Yun, and Lihong Li.
\newblock Webagent-r1: Training web agents via end-to-end multi-turn reinforcement learning.
\newblock \emph{arXiv preprint arXiv:2505.16421}, 2025{\natexlab{c}}.
\newblock URL \url{https://arxiv.org/abs/2505.16421}.

\bibitem[Wu et~al.(2025)Wu, Wang, Yu, Zhang, Chang, and Yu]{wu2025longmemeval}
Di~Wu, Hongwei Wang, Wenhao Yu, Yuwei Zhang, Kai-Wei Chang, and Dong Yu.
\newblock Longmemeval: Benchmarking chat assistants on long-term interactive memory.
\newblock In \emph{International Conference on Learning Representations}, 2025.
\newblock URL \url{https://arxiv.org/abs/2410.10813}.

\bibitem[Wu et~al.(2023)Wu, Bansal, Zhang, Wu, Li, Zhu, Jiang, Zhang, Zhang, Liu, Awadallah, White, Burger, and Wang]{wu2023autogen}
Qingyun Wu, Gagan Bansal, Jieyu Zhang, Yiran Wu, Beibin Li, Erkang Zhu, Li~Jiang, Xiaoyun Zhang, Shaokun Zhang, Jiale Liu, Ahmed~Hassan Awadallah, Ryen~W. White, Doug Burger, and Chi Wang.
\newblock {AutoGen}: Enabling next-gen {LLM} applications via multi-agent conversation.
\newblock \emph{arXiv preprint arXiv:2308.08155}, 2023.

\bibitem[Xiang et~al.(2025)Xiang, Zhang, Yang, Chu, Chu, Xie, Yuan, Sun, Gu, Wang, et~al.]{xiang2025navr2dualrelationreasoninggeneralizable}
Wentao Xiang, Haokang Zhang, Tianhang Yang, Zedong Chu, Ruihang Chu, Shichao Xie, Yujian Yuan, Jian Sun, Zhining Gu, Junjie Wang, et~al.
\newblock Nav-$r^2$ dual-relation reasoning for generalizable open-vocabulary object-goal navigation, 2025.

\bibitem[Xiao et~al.(2021)Xiao, Shang, Yao, and Chua]{nextqa}
Junbin Xiao, Xindi Shang, Angela Yao, and Tat-Seng Chua.
\newblock Next-qa: Next phase of question-answering to explaining temporal actions.
\newblock In \emph{Proceedings of the IEEE/CVF Conference on Computer Vision and Pattern Recognition}, 2021.

\bibitem[Xiao et~al.(2023)Xiao, Zhou, Yao, Li, Hong, Yan, and Chua]{covgt}
Junbin Xiao, Pan Zhou, Angela Yao, Yicong Li, Richang Hong, Shuicheng Yan, and Tat-Seng Chua.
\newblock Contrastive video question answering via video graph transformer.
\newblock \emph{IEEE Transactions on Pattern Analysis and Machine Intelligence}, 45\penalty0 (11):\penalty0 13265--13280, 2023.
\newblock \doi{10.1109/TPAMI.2023.3292266}.

\bibitem[Xie et~al.(2024)Xie, Zhang, Chen, Li, Zhao, Cao, Hua, Cheng, Shin, Lei, Liu, Xu, Zhou, Savarese, Xiong, Zhong, and Yu]{xie2024osworld}
Tianbao Xie, Danyang Zhang, Jixuan Chen, Xiaochuan Li, Siheng Zhao, Ruisheng Cao, Toh~Jing Hua, Zhoujun Cheng, Dongchan Shin, Fangyu Lei, Yitao Liu, Yiheng Xu, Shuyan Zhou, Silvio Savarese, Caiming Xiong, Victor Zhong, and Tao Yu.
\newblock {OSWorld}: Benchmarking multimodal agents for open-ended tasks in real computer environments.
\newblock \emph{arXiv preprint arXiv:2404.07972}, 2024.

\bibitem[Xu et~al.(2025{\natexlab{a}})Xu, Hu, Gao, Zhu, Zhao, Li, and Yin]{xu2025geonav}
Haotian Xu, Yue Hu, Chen Gao, Zhengqiu Zhu, Yong Zhao, Yong Li, and Quanjun Yin.
\newblock {GeoNav}: Empowering {MLLMs} with dual-scale geospatial reasoning for language-goal aerial navigation.
\newblock \emph{arXiv preprint arXiv:2504.09587}, 2025{\natexlab{a}}.
\newblock URL \url{https://arxiv.org/abs/2504.09587}.

\bibitem[Xu et~al.(2025{\natexlab{b}})Xu, Liang, Mei, Gao, Tan, and Zhang]{xu2025amem}
Wujiang Xu, Zujie Liang, Kai Mei, Hang Gao, Juntao Tan, and Yongfeng Zhang.
\newblock A-mem: Agentic memory for llm agents.
\newblock \emph{arXiv preprint arXiv:2502.12110}, 2025{\natexlab{b}}.
\newblock URL \url{https://arxiv.org/abs/2502.12110}.

\bibitem[Xue et~al.(2025)Xue, Hu, Luo, Xie, Chen, Xie, Quan, Guo, Xu, and Chu]{xue2026omninavunifiedframeworkprospective}
Xinda Xue, Junjun Hu, Minghua Luo, Shichao Xie, Jintao Chen, Zixun Xie, Kuichen Quan, Wei Guo, Mu~Xu, and Zedong Chu.
\newblock Omninav: A unified framework for prospective exploration and visual-language navigation, 2025.
\newblock URL \url{https://arxiv.org/abs/2509.25687}.

\bibitem[Yang et~al.(2021)Yang, Miech, Sivic, Laptev, and Schmid]{yang2021justask}
Antoine Yang, Antoine Miech, Josef Sivic, Ivan Laptev, and Cordelia Schmid.
\newblock Just ask: Learning to answer questions from millions of narrated videos.
\newblock In \emph{ICCV}, 2021.

\bibitem[Yang et~al.(2024{\natexlab{a}})Yang, Wang, Lu, Liu, Le, Zhou, and Chen]{yang2024opro}
Chengrun Yang, Xuezhi Wang, Yifeng Lu, Hanxiao Liu, Quoc~V. Le, Denny Zhou, and Xinyun Chen.
\newblock Large language models as optimizers.
\newblock In \emph{International Conference on Learning Representations}, 2024{\natexlab{a}}.

\bibitem[Yang et~al.(2025{\natexlab{a}})Yang, Liu, Guo, Dong, Zhang, Zhang, Wang, Zhou, Xie, Wang, et~al.]{egolife}
Jingkang Yang, Shuai Liu, Hongming Guo, Yuhao Dong, Xiamengwei Zhang, Sicheng Zhang, Pengyun Wang, Zitang Zhou, Binzhu Xie, Ziyue Wang, et~al.
\newblock Egolife: Towards egocentric life assistant.
\newblock \emph{arXiv preprint arXiv:2503.03803}, 2025{\natexlab{a}}.

\bibitem[Yang et~al.(2024{\natexlab{b}})Yang, Jimenez, Wettig, Lieret, Yao, Narasimhan, and Press]{yang2024sweagent}
John Yang, Carlos~E. Jimenez, Alexander Wettig, Kilian Lieret, Shunyu Yao, Karthik Narasimhan, and Ofir Press.
\newblock Swe-agent: Agent-computer interfaces enable automated software engineering.
\newblock \emph{arXiv preprint arXiv:2405.15793}, 2024{\natexlab{b}}.

\bibitem[Yang et~al.(2025{\natexlab{b}})Yang, Zhang, Wang, Chu, Wu, Cai, and Xu]{yang2025cenavflowguidedreinforcementrefinement}
Kai Yang, Tianlin Zhang, Zhengbo Wang, Zedong Chu, Xiaolong Wu, Yang Cai, and Mu~Xu.
\newblock Ce-nav: Flow-guided reinforcement refinement for cross-embodiment local navigation, 2025{\natexlab{b}}.
\newblock URL \url{https://arxiv.org/abs/2509.23203}.

\bibitem[Yang et~al.(2026{\natexlab{a}})Yang, Chu, Guo, Wang, Xie, Shen, Wu, Li, and Xu]{yang2026asyncshieldplugandplayedgeadapter}
Kai Yang, Zedong Chu, Yingnan Guo, Zhengbo Wang, Shichao Xie, Yanfen Shen, Xiaolong Wu, Xing Li, and Mu~Xu.
\newblock Asyncshield: A plug-and-play edge adapter for asynchronous cloud-based vla navigation, 2026{\natexlab{a}}.
\newblock URL \url{https://arxiv.org/abs/2604.24086}.

\bibitem[Yang et~al.(2025{\natexlab{c}})]{yang2025embodiedbench}
Rui Yang et~al.
\newblock {EmbodiedBench}: Comprehensive benchmarking multi-modal large language models for vision-driven embodied agents.
\newblock \emph{arXiv preprint arXiv:2502.09560}, 2025{\natexlab{c}}.

\bibitem[Yang et~al.(2026{\natexlab{b}})Yang, Zeng, Lin, Chang, Qi, Xiao, Liu, Chen, Chen, Huo, et~al.]{yang2026abot}
Yandan Yang, Shuang Zeng, Tong Lin, Xinyuan Chang, Dekang Qi, Junjin Xiao, Haoyun Liu, Ronghan Chen, Yuzhi Chen, Dongjie Huo, et~al.
\newblock Abot-m0: Vla foundation model for robotic manipulation with action manifold learning.
\newblock \emph{arXiv preprint arXiv:2602.11236}, 2026{\natexlab{b}}.

\bibitem[Yang et~al.(2025{\natexlab{d}})Yang, Yang, Zhou, Chen, Zhang, Du, and Gan]{yang2025threedmem}
Yuncong Yang, Han Yang, Jiachen Zhou, Peihao Chen, Hongxin Zhang, Yilun Du, and Chuang Gan.
\newblock 3d-mem: 3d scene memory for embodied exploration and reasoning.
\newblock In \emph{Proceedings of the IEEE/CVF Conference on Computer Vision and Pattern Recognition}, pages 17294--17303, 2025{\natexlab{d}}.
\newblock URL \url{https://arxiv.org/abs/2411.17735}.

\bibitem[Yang et~al.(2025{\natexlab{e}})Yang, Chen, Zhou, Yan, Song, Liu, Li, Zhang, Zhou, Chen, et~al.]{yang2025agentic}
Zhejian Yang, Yongchao Chen, Xueyang Zhou, Jiangyue Yan, Dingjie Song, Yinuo Liu, Yuting Li, Yu~Zhang, Pan Zhou, Hechang Chen, et~al.
\newblock Agentic robot: A brain-inspired framework for vision-language-action models in embodied agents.
\newblock \emph{arXiv preprint arXiv:2505.23450}, 2025{\natexlab{e}}.

\bibitem[Yao et~al.(2022)Yao, Zhao, Yu, Du, Shafran, Narasimhan, and Cao]{yao2022react}
Shunyu Yao, Jeffrey Zhao, Dian Yu, Nan Du, Izhak Shafran, Karthik Narasimhan, and Yuan Cao.
\newblock React: Synergizing reasoning and acting in language models.
\newblock \emph{arXiv preprint arXiv:2210.03629}, 2022.

\bibitem[Yao et~al.(2024)Yao, Shinn, Razavi, and Narasimhan]{yao2024taubench}
Shunyu Yao, Noah Shinn, Pedram Razavi, and Karthik Narasimhan.
\newblock $\tau$-bench: A benchmark for tool-agent-user interaction in real-world domains.
\newblock \emph{arXiv preprint arXiv:2406.12045}, 2024.

\bibitem[Ye et~al.(2023)Ye, Xu, Yan, Xu, Qian, Zhang, and Huang]{HiTeA}
Qinghao Ye, Guohai Xu, Ming Yan, Haiyang Xu, Qi~Qian, Ji~Zhang, and Fei Huang.
\newblock Hitea: Hierarchical temporal-aware video-language pre-training.
\newblock In \emph{2023 IEEE/CVF International Conference on Computer Vision (ICCV)}, pages 15359--15370, 2023.
\newblock \doi{10.1109/ICCV51070.2023.01413}.

\bibitem[Yeo et~al.(2025{\natexlab{a}})Yeo, Kim, Jeong, Baek, and Hwang]{yeo2025universalrag}
Woongyeong Yeo, Kangsan Kim, Soyeong Jeong, Jinheon Baek, and Sung~Ju Hwang.
\newblock Universalrag: Retrieval-augmented generation over corpora of diverse modalities and granularities.
\newblock \emph{arXiv preprint arXiv:2504.20734}, 2025{\natexlab{a}}.
\newblock URL \url{https://arxiv.org/abs/2504.20734}.

\bibitem[Yeo et~al.(2025{\natexlab{b}})Yeo, Kim, Yoon, and Hwang]{worldmm}
Woongyeong Yeo, Kangsan Kim, Jaehong Yoon, and Sung~Ju Hwang.
\newblock Worldmm: Dynamic multimodal memory agent for long video reasoning.
\newblock \emph{arXiv preprint arXiv:2512.02425}, 2025{\natexlab{b}}.

\bibitem[Zhang et~al.(2023)Zhang, Lu, Islam, Wang, Yu, Bansal, and Bertasius]{zhang2023simple}
Ce~Zhang, Taixi Lu, Md~Mohaiminul Islam, Ziyang Wang, Shoubin Yu, Mohit Bansal, and Gedas Bertasius.
\newblock A simple llm framework for long-range video question-answering, 2023.

\bibitem[Zhang et~al.(2025{\natexlab{a}})Zhang, Ren, Zhan, Zhou, Wang, Zhu, Zhou, and Yan]{memevolve}
Guibin Zhang, Haotian Ren, Chong Zhan, Zhenhong Zhou, Junhao Wang, He~Zhu, Wangchunshu Zhou, and Shuicheng Yan.
\newblock Memevolve: Meta-evolution of agent memory systems.
\newblock \emph{arXiv preprint arXiv:2512.18746}, 2025{\natexlab{a}}.

\bibitem[Zhang et~al.(2026)Zhang, Long, Bao, Feng, Zhang, Yue, and Wang]{memskill}
Haozhen Zhang, Quanyu Long, Jianzhu Bao, Tao Feng, Weizhi Zhang, Haodong Yue, and Wenya Wang.
\newblock Memskill: Learning and evolving memory skills for self-evolving agents.
\newblock \emph{arXiv preprint arXiv:2602.02474}, 2026.

\bibitem[Zhang et~al.(2025{\natexlab{b}})Zhang, Gao, Yu, Peng, Zhao, Zhang, Cui, Chen, and Li]{zhang2025citynavagent}
Weichen Zhang, Chen Gao, Shiquan Yu, Ruiying Peng, Baining Zhao, Qian Zhang, Jinqiang Cui, Xinlei Chen, and Yong Li.
\newblock Citynavagent: Aerial vision-and-language navigation with hierarchical semantic planning and global memory.
\newblock In \emph{Proceedings of the 63rd Annual Meeting of the Association for Computational Linguistics (Volume 1: Long Papers)}, pages 31292--31309, 2025{\natexlab{b}}.

\bibitem[Zhang(2026)]{autogenesis}
Wentao Zhang.
\newblock Autogenesis: A self-evolving agent protocol.
\newblock \emph{arXiv preprint arXiv:2604.15034}, 2026.

\bibitem[Zheng et~al.(2023)Zheng, Chiang, Sheng, Zhuang, Wu, Zhuang, Lin, Li, Li, Xing, et~al.]{zheng2023judging}
Lianmin Zheng, Wei-Lin Chiang, Ying Sheng, Siyuan Zhuang, Zhanghao Wu, Yonghao Zhuang, Zi~Lin, Zhuohan Li, Dacheng Li, Eric~P. Xing, et~al.
\newblock Judging llm-as-a-judge with mt-bench and chatbot arena.
\newblock In \emph{Advances in Neural Information Processing Systems}, 2023.

\bibitem[Zhong et~al.(2023)Zhong, Guo, Gao, Ye, and Wang]{memorybank}
Wanjun Zhong, Lianghong Guo, Qiqi Gao, He~Ye, and Yanlin Wang.
\newblock Memorybank: Enhancing large language models with long-term memory.
\newblock \emph{arXiv preprint arXiv:2305.10250}, 2023.

\bibitem[Zhou et~al.(2023)Zhou, Xu, Zhu, Zhou, Lo, Sridhar, Cheng, Ou, Bisk, Fried, Alon, and Neubig]{zhou2023webarena}
Shuyan Zhou, Frank~F. Xu, Hao Zhu, Xuhui Zhou, Robert Lo, Abishek Sridhar, Xianyi Cheng, Tianyue Ou, Yonatan Bisk, Daniel Fried, Uri Alon, and Graham Neubig.
\newblock Webarena: A realistic web environment for building autonomous agents.
\newblock \emph{arXiv preprint arXiv:2307.13854}, 2023.

\bibitem[Zhou et~al.(2024)Zhou, Zanette, Pan, Levine, and Kumar]{zhou2024archer}
Yifei Zhou, Andrea Zanette, Jiayi Pan, Sergey Levine, and Aviral Kumar.
\newblock Archer: Training language model agents via hierarchical multi-turn rl.
\newblock \emph{arXiv preprint arXiv:2402.19446}, 2024.
\newblock URL \url{https://arxiv.org/abs/2402.19446}.

\bibitem[Zhu et~al.(2026)Zhu, Guo, Mei, Pang, Zhang, He, Ji, and Liu]{zhu2026robotchores}
Zilin Zhu, Longteng Guo, Yanghong Mei, Bowen Pang, Zongxun Zhang, Xingjian He, Ruyi Ji, and Jing Liu.
\newblock When robots do the chores: A benchmark and agent for long-horizon household task execution.
\newblock \emph{arXiv preprint arXiv:2605.14504}, 2026.
\newblock URL \url{https://arxiv.org/abs/2605.14504}.

\end{thebibliography}

\clearpage
% \input{sections/contribution}
% \clearpage
\beginappendix

\section{Memory Benchmark Judge Prompts}
\label{app:memory_judge_prompts}

This appendix reports the judge prompts used for the three LLM-judged memory benchmarks.
Each benchmark uses a separate prompt matched to its answer format and scoring protocol.

\subsection{OpenEQA Judge Prompt}
\label{app:judge_openeqa}

\promptheading{OpenEQA LLM-Match Judge}

\begin{lstlisting}[style=promptstyle]
You are an evaluator for OpenEQA question answering.

Given a question, a reference answer, and a predicted answer, score whether the prediction answers the question correctly.

Use a 5-point score:
- 5: fully correct and complete.
- 4: mostly correct; minor omission or wording difference.
- 3: partially correct; captures the main idea but misses important details.
- 2: weakly related; some relevant information but mostly incomplete or vague.
- 1: incorrect, unsupported, contradictory, or missing.

Guidelines:
- Accept semantically equivalent answers.
- For location questions, accept equivalent room/place descriptions if they identify the same target.
- For object-state questions, polarity must match.
- For count questions, the number must match unless the reference itself is approximate.
- Do not reward hallucinated details that contradict the reference.

Return only valid JSON:
{
  "score": 1,
  "llm_match": false,
  "reason": "brief explanation"
}
\end{lstlisting}

\subsection{Mem-Gallery Judge Prompt}
\label{app:judge_memgallery}

\promptheading{Mem-Gallery LLM-Judge}

\begin{lstlisting}[style=promptstyle]
You are an impartial judge evaluating the memory capabilities of an AI assistant with a question-answering task.

Compare the assistant answer against the ground truth for the given question.

Allowed scores are exactly 0, 0.25, 0.5, 0.75, or 1.
The strict parser preserves these five levels. The optional official-style parser buckets them to the public evaluator scale: below 0.25 -> 0, below 0.75 -> 0.5, otherwise -> 1.

Scoring rubric:
- 0: incorrect or missing; contradicts the ground truth or gives unrelated information.
- 0.25: weakly related but mostly incorrect.
- 0.5: partially correct but missing important information.
- 0.75: mostly correct with minor omission or wording issue.
- 1: correct or semantically equivalent.

Return only valid JSON:
{
  "score": 1,
  "reasoning": "short explanation"
}

Question:
{question}

Ground truth:
{ground_truth}

Assistant answer:
{prediction}
\end{lstlisting}

\subsection{LoCoMo Judge Prompt}
\label{app:judge_locomo}

\promptheading{LoCoMo Memory-Recall Judge}

\begin{lstlisting}[style=promptstyle]
System:
You are evaluating conversational AI memory recall. Return JSON only with the format requested.

User:
Label the generated answer as CORRECT or WRONG.

Rules:

1. PARTIAL CREDIT:
If the generated answer includes at least one correct item from the gold answer's list, mark CORRECT. Getting 1 out of 2, 2 out of 4, etc. is acceptable. Only mark WRONG if none of the gold answer items appear.

2. PARAPHRASES COUNT:
Same concept in different words is CORRECT. Emotions and sentiments in the same positive/negative family count as paraphrases. Judge semantic meaning, not exact wording.

3. EXTRA DETAIL IS FINE:
A longer answer that includes the gold answer's key facts plus additional information is CORRECT. Do not penalize for being more detailed or specific.

4. DATE TOLERANCE:
Dates within 14 days of each other are CORRECT. Durations within 50% are CORRECT, e.g. "5 months" matches "six months" and "19 days" matches "two weeks". Relative dates such as "few days before November" match specific dates in the same window. A specific date that is consistent with a vague reference is CORRECT. Converting "last year" to the actual year relative to the conversation year is CORRECT.

5. SEMANTIC OVERLAP:
Judge whether the generated answer addresses the same topic and captures the core idea of the gold answer. For emotions and feelings questions, answers expressing sentiments in the same valence about the same event are CORRECT.

6. SAME REFERENT:
If the generated answer mentions or references the same named entity, character, person, or concept as the gold answer, mark CORRECT even if the generated answer provides a different physical description or additional detail.

7. FOCUS ON KNOWLEDGE, NOT WORDING:
The goal is to assess whether the system recalled the right fact. Minor differences in specificity, phrasing, or scope should not result in WRONG.

Optional evidence variant:
If actual conversation evidence is supplied and corroborates the generated answer, mark CORRECT even when the generated answer diverges from the gold answer. Use evidence only to accept answers, never to reject them more strictly.

Only mark WRONG if:
- The generated answer contains zero correct items from the gold answer.
- The answer addresses a completely different topic.

Return JSON with:
{
  "reasoning": "one sentence",
  "label": "CORRECT | WRONG"
}
\end{lstlisting}

\section{Lifelong Self-Evolution Process Example}
\label{app:self_evo_process_example}

This appendix example summarizes one OpenEQA lifelong self-evolution run.
The run used fixed data splits and did not modify the dataset schema.
Self-evolution produced auditable JSON DSL runtime assets rather than question-specific answer patches.

\paragraph{Procedure}
For each split, the system executes the following loop.
\begin{enumerate}
    \item \textbf{Incumbent evaluation.} Run the current incumbent memory system on the split.
    \item \textbf{Diagnosis.} The Diagnoser inspects failed QA rows, retrieval traces, and graph evidence, and clusters failures by memory root cause.
    \item \textbf{Hypothesis generation.} The HypothesisGenerator proposes generic JSON DSL candidate assets for writer, retriever, answerer, or frame-policy layers. Candidate metadata can record target failures for audit, but runtime behavior is controlled by generic activation conditions and evidence requirements.
    \item \textbf{Compilation and safety review.} The CompilerCritic narrows candidates into safe runtime policies, rejects executable code, schema migration, direct answer resolvers, fixed answer tables, and qid/gold-answer rules.
    \item \textbf{Gate analysis.} The GateAnalyst adds protected-category constraints, activation guards, cost/lifecycle constraints, and regression risks.
    \item \textbf{Candidate evaluation.} Each materialized asset is evaluated against the current incumbent. Candidate acceptance requires target improvement, no global regression, no low-score-count increase, and no protected-category regression.
    \item \textbf{Stack confirmation.} All accepted assets for the round are evaluated together. Only assets whose combined stack beats the incumbent are carried forward; otherwise newly accepted assets from that round are deprecated.
\end{enumerate}

\paragraph{Representative split example}
The clearest accepted example is \texttt{split\_00}.
The Diagnoser identified a \textit{derived-fact missing / adapter-to-graph materialization} failure:
\begin{itemize}
    \item upstream scene evidence contained an object observation, such as a throw blanket on a bed;
    \item the graph did not expose a corresponding typed object node or directed object-room/object-object relation;
    \item retrieval returned unrelated object records instead of the target object evidence;
    \item the answerer therefore produced an unsupported or missing answer.
\end{itemize}
This diagnosis points to a memory-system failure, not a direct answer-format patch.

\paragraph{Proposed candidates}
The HypothesisGenerator proposed three generic runtime assets, shown in Table~\ref{tab:split00_self_evo_candidates}.

\begin{table}[t]
\centering
\small
\begin{tabular}{p{0.25\linewidth} p{0.12\linewidth} p{0.54\linewidth}}
\hline
\textbf{Candidate} & \textbf{Layer} & \textbf{Proposed evolution} \\
\hline
Writer materialization
& Writer
& Promote adapter keyframe object mentions into typed object observations with aliases, attributes, observed state facts, room/place linkage, last-seen evidence, source reference, frame evidence, confidence, and provenance. \\
Retriever room-anchor/last-seen focus
& Retriever
& Prefer observed object memories using room/place anchors, last-seen evidence, object identity, and directed spatial grounding rather than broad scene-level lexical similarity. \\
Writer directed support relations
& Writer
& Write directed spatial support relations only when both endpoints and relation direction are explicitly grounded by observation evidence. \\
\hline
\end{tabular}
\caption{Candidate assets proposed for the representative OpenEQA self-evolution split.}
\label{tab:split00_self_evo_candidates}
\end{table}

\paragraph{Safety constraints}
CompilerCritic and GateAnalyst narrowed the candidates with generic guards:
\begin{itemize}
    \item do not write objects from question text, room priors, or scene summaries alone;
    \item do not infer object state, affordance, or relation unless visually or observationally grounded;
    \item do not reverse spatial relation direction;
    \item do not suppress exact object identity matches from another room if they are the only grounded evidence;
    \item preserve source reference, frame path when available, confidence, and last-seen evidence for audit;
    \item protect object localization, object state recognition, and last-seen categories from regression.
\end{itemize}

\paragraph{Gate outcomes}
The gate outcomes for \texttt{split\_00} are shown in Table~\ref{tab:split00_self_evo_gate}.

\begin{table}[t]
\centering
\small
\begin{tabular}{p{0.27\linewidth} p{0.16\linewidth} p{0.49\linewidth}}
\hline
\textbf{Candidate} & \textbf{Gate result} & \textbf{Main reason} \\
\hline
Writer materialization
& Rejected
& Although target and global score improved, protected object-state recognition regressed slightly beyond the configured tolerance. \\
Retriever room-anchor/last-seen focus
& Accepted
& Target delta was $+0.800$, global delta was $+0.044$, and low-score count decreased by $10$. \\
Writer directed support relations
& Rejected
& Target delta was negative and protected object-state recognition regressed. \\
\hline
\end{tabular}
\caption{Candidate-level gate outcomes for \texttt{split\_00}.}
\label{tab:split00_self_evo_gate}
\end{table}

\paragraph{Stack confirmation}
The stack containing the accepted retriever asset passed confirmation:
\begin{itemize}
    \item normalized baseline mean: $0.6053$;
    \item normalized stack mean: $0.6545$;
    \item stack delta: $+0.0492$;
    \item low-score count changed from $91$ to $77$;
    \item protected object-state recognition did not regress;
    \item protected object-localization improved by approximately $+0.073$.
\end{itemize}
The accepted asset was then written to the global evolution log as a provisional retriever asset.

\paragraph{Resulting runtime policy.}
The accepted retriever policy can be summarized as follows:
\begin{quote}
\small
\ttfamily
When an OpenEQA retrieval query contains an object anchor, room/place cue, or spatial phrase, rank observation-grounded object memories ahead of broad scene summaries. Prefer records with matching room/place, explicit last\_seen, source\_ref/frame evidence, object identity, and directed spatial relation evidence. Do not infer room priors, reverse relation direction, or use ungrounded scene summaries as object evidence.
\end{quote}
This is a generic retrieval policy. It does not contain gold answers, manual answer rules, or question-specific shortcuts.

\paragraph{Eight-split evolution trace.}

Table~\ref{tab:eight_split_self_evo_trace} summarizes what each split proposed and whether the changes survived stack confirmation.

\begin{table*}[t]
\centering
\scriptsize
\begin{tabular}{p{0.08\textwidth} p{0.36\textwidth} p{0.25\textwidth} p{0.22\textwidth}}
\hline
\textbf{Split} & \textbf{Main proposed evolution} & \textbf{Candidate gate outcome} & \textbf{Stack outcome} \\
\hline
\texttt{split\_00}
& Writer materialization of observed adapter objects; retriever room-anchor/last-seen focus; writer directed support relations.
& Retriever room-anchor/last-seen policy accepted; two writer policies rejected.
& Accepted and carried forward. \\
\texttt{split\_01}
& Retriever anchor-transition trace ranking; writer grounded room-transition facts.
& Both rejected due target/global regression and protected-category risk.
& No new asset promoted. \\
\texttt{split\_02}
& Retriever last-seen/state conflict tracing; retriever directed spatial relation guard; writer observation/state alias completion.
& All rejected due insufficient target gain or global/protected regression.
& No new asset promoted. \\
\texttt{split\_03}
& Retriever grounded spatial/recency/object ranking; writer last-seen completeness; writer identity/alias preservation.
& Retriever candidate passed candidate gate; writer candidates rejected.
& Stack failed; newly accepted retriever asset deprecated. \\
\texttt{split\_04}
& Retriever recency/directional evidence reranking; writer object-observation provenance completion.
& Both rejected due target/global regression.
& No new asset promoted. \\
\texttt{split\_05}
& Writer keyframe object promotion; retriever last-seen/directed spatial ranking.
& Writer candidate accepted by global-rescue gate; retriever rejected.
& Stack failed; newly accepted writer asset deprecated. \\
\texttt{split\_06}
& Retriever anchor-directed same-room binding; writer typed object observation exposure.
& Both rejected due insufficient target gain or global regression.
& No new asset promoted. \\
\texttt{split\_07}
& Retriever directed spatial relation first; writer object/state/last-seen consolidation.
& Retriever candidate passed candidate gate; writer rejected.
& Stack failed; newly accepted retriever asset deprecated. \\
\hline
\end{tabular}
\caption{Eight-split trace of the OpenEQA lifelong self-evolution run. Candidate-level acceptance is not sufficient for lifelong acceptance; new assets are retained only if the full stack passes confirmation.}
\label{tab:eight_split_self_evo_trace}
\end{table*}

\paragraph{Takeaway}
The successful evolution was not an answer patch.
It changed the retriever policy so that evidence already present in graph memory is ranked and exposed more reliably.
Most proposed writer policies were rejected because they risked object-state or localization regressions, and candidate-level wins were not enough unless the full evolved stack also passed confirmation.

\end{document}